\documentclass{article}

\usepackage[main, final]{neurips_2026}

\usepackage[utf8]{inputenc}
\usepackage{caption}   
\usepackage{needspace} 
\usepackage[T1]{fontenc}
\usepackage{hyperref}
\usepackage{url}
\usepackage{booktabs}
\usepackage{wrapfig}
\usepackage{threeparttable}
\usepackage{amsfonts}
\usepackage{wrapfig}
\usepackage{nicefrac}
\usepackage{microtype}
\usepackage[table]{xcolor}

\usepackage{graphicx}
\usepackage{subcaption}
\usepackage{bm}
\usepackage{amsmath}
\usepackage{amssymb}
\usepackage{mathtools}
\usepackage{amsthm}
\usepackage[capitalize,noabbrev]{cleveref}

\usepackage{makecell}
\usepackage{tabularx}
\usepackage{multirow}
\usepackage{adjustbox}
\usepackage{algorithm}
\usepackage{algorithmic}
\usepackage{threeparttable}
\usepackage{tikz}
\usepackage{pifont}
\usepackage[disable,textsize=tiny]{todonotes}

\usepackage{soul}      
\usepackage{ulem}      
\renewcommand{\algorithmicrequire}{\textbf{Input:}}
\renewcommand{\algorithmicensure}{\textbf{Output:}}

\newcommand{\hbarwide}[4]{%
\begin{tikzpicture}[x=1.0cm,y=0.18cm,baseline=-0.45ex]
  \draw[line width=2.2pt,gray!70,line cap=round] (0,0) -- (1,0);
  \draw[line width=3.0pt,black,line cap=round] (#1,0) -- (#3,0);
\end{tikzpicture}%
}

\theoremstyle{plain}

\newtheorem{definition}{Definition}

\theoremstyle{remark}

\title{Foundation Pareto Flow Policy for Multi-Objective Reinforcement Learning}

\author{%
  Zhanjiang Yang$^{1,2}$\thanks{Equal contribution.}
  \qquad
  Lijun Sun$^{1*}$ 
  \qquad
  Yueming Li$^{1}$ 
  \qquad
  Meng Li$^1$ 
  \quad
  Yijun Yang$^{3}$\thanks{Corresponding author.} \\
  $^{1}$Shenzhen Technology University
  \quad
  $^{2}$Shenzhen University 
  \quad
  $^{3}$Tsinghua University\\
  \texttt{2410263030@mails.szu.edu.cn}
  \quad
  \texttt{sunlijun@sztu.edu.cn}\\
  \texttt{202301510025@stumail.sztu.edu.cn}
  \quad
  \texttt{limeng2@sztu.edu.cn}\\
  \texttt{yijun.steven.yang@gmail.com}
}

\begin{document}

\maketitle

\begin{abstract}
The overarching goal of reinforcement learning (RL) is to efficiently optimize the agent's policy for various environments, tasks, dynamics, and even user preferences, which often consist of multiple and potentially conflicting rewards/objectives.
This necessitates the learning of Pareto-optimal policies that can navigate different trade-offs among objectives.
However, existing multi-objective RL (MORL) methods, whether trained online or offline, are typically limited to task- or objective-specific policy learning and do not address generalization of learned trade-offs to unseen tasks/objectives.
In this work, we propose the \textbf{F}oundation \textbf{P}areto \textbf{F}low \textbf{P}olicy (\textbf{FP2}) for offline MORL in continuous robotic control, inspired by the pre-mid-post-training paradigm underlying recent foundation models.
Specifically, we instantiate FP2 as a multi-modal diffusion transformer equipped with a likelihood-free flow matching loss, enabling unified, scalable, and instructable learning of diverse trade-off trajectories.
We further develop a three-stage training pipeline to gradually optimize the policy: (1) pre-training on a large-scale MORL dataset, (2) mid-training on a carefully curated high-quality dataset, and (3) post-training on few collected trajectory data from new tasks and trade-offs.
Empirical results demonstrate that FP2 outperforms previous state-of-the-art methods on most tasks of the D4MORL benchmark, achieving superior zero-shot interpolation and extrapolation on complex Pareto manifolds as well as efficient few-shot adaptation of the learned trade-off prior to new continuous-control tasks.
Code is available at \url{https://github.com/FibonacciRabbit/FP2}.
\end{abstract}

\section{Introduction}

\begin{figure}
\centering
\includegraphics[width=\linewidth]{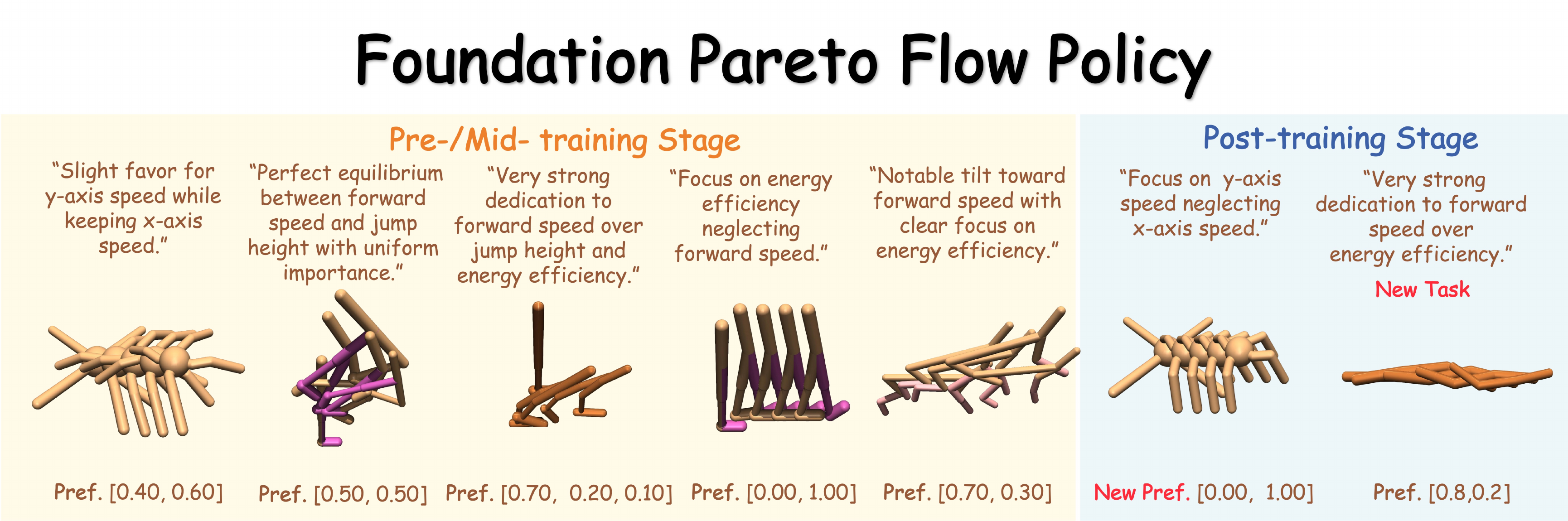}
\caption{\textbf{F}oundation \textbf{P}areto \textbf{F}low \textbf{P}olicy (\textbf{FP2}) for MORL. FP2 is a rectified flow-based policy conditioned on numerical and textual user preferences. Through principled pre- and mid-training on diverse datasets (yellow region), FP2 learns a reusable multi-objective trade-off prior that generalizes to out-of-distribution preferences and can be efficiently adapted to new locomotion tasks (blue region) via few-shot post-training.
}

\label{fig:teaser}
\vspace{-1pt}
\end{figure}

Multi-objective reinforcement learning (MORL)~\cite{roijers2013survey,roijers2017multi} solves real-world decision-making that involves multiple, often conflicting objectives~\cite{vamplew2022scalar} through learning Pareto-optimal policies capturing diverse trade-offs among objectives.
The preferred trade-off of a decision-making task, such as maximizing antibacterial activity while minimizing toxicity and production costs in drug design~\cite{stokes2020deep} or balancing competing objectives in complex continuous robot control~\cite{xu2020prediction,zhu2023scaling}, is typically unpredictable and can evolve over time.
For example, faster robotic movement may be preferred when energy is abundant, whereas careful, energy-efficient strategy is required when energy is limited in locomotion tasks.
Hence, MORL needs to learn Pareto policies that can handle different tasks, scenarios, and preferences, as illustrated by the proof-of-concept result in Fig.~\ref{fig:teaser}.

\raggedbottom
\setlength{\textfloatsep}{6pt plus 1pt minus 2pt}
\setlength{\floatsep}{6pt plus 1pt minus 2pt}
\setlength{\intextsep}{6pt plus 1pt minus 2pt}
\setlength{\abovecaptionskip}{2pt}

\begin{figure}[t]
\centering
\includegraphics[width=0.9\linewidth]{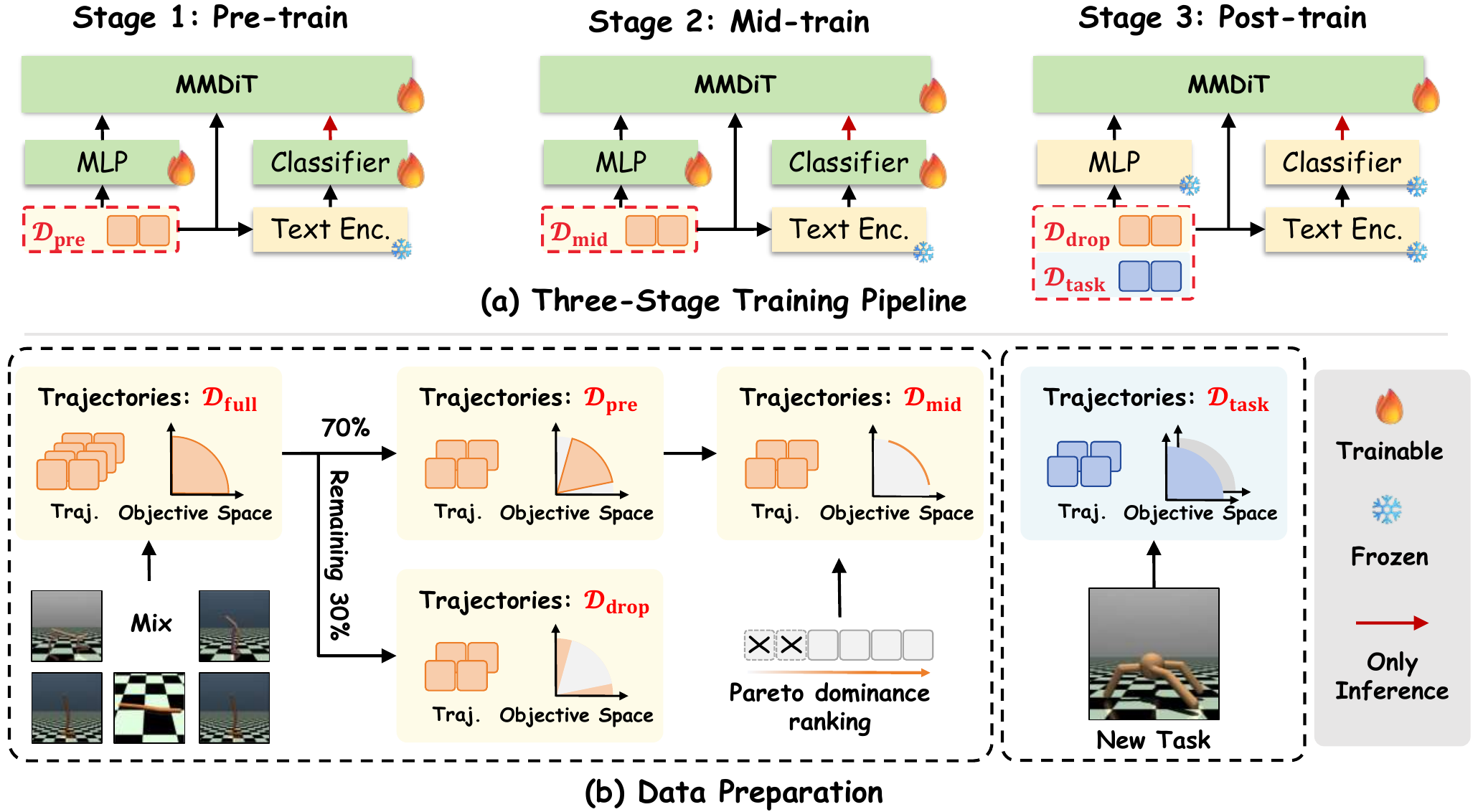}
\caption{Three-stage training pipeline of  \textbf{F}oundation \textbf{P}areto \textbf{F}low \textbf{P}olicy (\textbf{FP2}).
In Stage~1, FP2 is pre-trained to learn diverse long-horizon trade-off behaviors based on large-scale trajectories $\textcolor{red}{\bm{\mathcal{D}}_{\text{pre}}}$.
In Stage~2, FP2 is continuously mid-trained to align policy behavior with the target preference using curated high-quality Pareto non-dominated trajectories $\textcolor{red}{\bm{\mathcal{D}}_{\text{mid}}}$.
In Stage~3, FP2 can be further post-trained to efficiently adapt to unseen preference configurations or new tasks with the remaining trajectories $\textcolor{red}{\bm{\mathcal{D}}_{\text{drop}}}$ or a very small amount of new task data $\textcolor{red}{\bm{\mathcal{D}}_{\text{task}}}$.
(See Sec.~\ref{Sec:3_stage_train} for more details.)}
\label{fig:training_stages}
\vspace{-1pt}
\end{figure}

To this end, we introduce a generalizable offline MORL policy, named \textbf{F}oundation \textbf{P}areto \textbf{F}low \textbf{P}olicy (\textbf{FP2}), which is built upon the multi-modal diffusion transformer (MMDiT)~\cite{peebles2023scalable} equipped with a rectified flow matching loss~\cite{esser2024scaling,liu2023instaflow,ma2025janusflow,black2024pi_0}, enabling unified and instructable sequence modeling of diverse Pareto behavior trajectories.
Furthermore, inspired by the pre-training--post-training paradigm underlying recent foundation models, we develop a three-stage training pipeline to gradually optimize the policy FP2 from rich yet suboptimal trajectory data to scarce but high-quality data.
As shown in Fig.~\ref{fig:training_stages}, we pre-train FP2 in Stage 1 on a large-scale, diverse, but noisy dataset including trajectory data accessible from several offline MORL datasets.
In Stage~2, we design a novel mid-training stage that continuously fine-tunes the pre-trained policy (i.e., FP2-Pre) using a curated set of high-quality Pareto-optimal trajectories, sharpening the policy's action distribution for Pareto-front behaviors while improving policy--preference alignment. 
This stage mitigates the scarcity of expert data by amplifying their supervisory effect for learning fine-grained representations, leading to improved performance.
In Stage~3, we post-train the FP2-Mid policy with a very small amount of task-specific data to efficiently adapt it to unseen preference configurations or environment dynamics. {Moreover,} to boost the policy's robustness to noisy and suboptimal data that exists in commonly-used MORL datasets, FP2 introduces a contrastive alignment regularization to effectively overcome the potential misalignment between 
{numerical/textual} preferences and practical behavior trajectories. 
In experiments, we conduct extensive comparison with  strong MORL baselines on the D4MORL benchmark~\cite{zhu2023scaling}, demonstrating the superiority of FP2 in terms of zero-shot interpolation and extrapolation on complex Pareto manifolds, as well as few-shot adaptation of a shared multi-objective trade-off prior across held-out environments.
\looseness-1

Our contributions are summarized below:
\begin{itemize}
    \item A pre-trained offline MORL policy that provides a reusable multi-objective trade-off prior for continuous control, generalizing zero-shot to unseen preferences and few-shot transferring of this prior to new control tasks.
    \item A principled pipeline including pre-, mid-, and post-training that takes the same spirit as mainstream training recipes adopted by advanced foundation models.
    \item To the best of our knowledge, FP2 is the first offline MORL method that exhibits excellent interpolation/extrapolation abilities on complex Pareto manifolds 
    and transfer of learned multi-objective trade-off structure across locomotion tasks.
\end{itemize}

\section{Related work}

\textbf{Multi-objective optimization (MOO).}
MOO seeks Pareto-optimal solutions balancing multiple, often conflicting objectives.
These solutions form the Pareto set of undominated decisions in the decision space and induce the Pareto front in the objective space~\cite{felten2024multi, hayes2022practical, chen2025gradient}.
Existing MOO methods can be broadly categorized 
by whether preferences are specified \emph{a priori}.
When preferences are known, a single Pareto-optimal solution can be obtained by optimizing the corresponding scalarized objective, obviating the need to recover the entire Pareto set~\cite{roijers2013survey, van2013scalarized, roijers2018multi}.
When preferences are unknown, two classes of approaches enable preference-specific solutions at test time: finite Pareto-set methods and continuous Pareto-set methods.
Finite Pareto-set methods learn a collection of policies that span diverse objective preferences, yielding a discrete approximation of the Pareto set~\cite{xu2020prediction,yang2021pareto, janmohamed2024preference,liu2024c}.
Continuous Pareto-set methods model a preference–solution mapping that captures the underlying continuous Pareto structure through a unified preference-conditioned policy~\cite{yuan2024moduli, du2025controllable}.

MORL builds upon the MOO framework by incorporating sequential decision-making.
In such settings, meaningful behaviors arise only through cumulative trade-offs over an entire rollout, inducing complex temporal correlations and making Pareto-optimal policy learning fundamentally more difficult~\cite{annadani2025preference}.

\textbf{Generative models for offline MORL.}
MORL is generally framed as a sequential decision-making problem involving state transitions.
Traditional RL algorithms typically learn policies using value-based or policy-based methods~\cite{sutton2018reinforcement,lin2005min,mossalam2016multi,yang2019generalized,xu2020prediction,liu2025pareto,li2026cola,li2026reinforcement}. 
Recently, conditional generative models have emerged as an alternative paradigm for offline decision-making by formulating policy learning as return-conditioned sequence modeling.
These approaches can be broadly categorized into two classes.
The first class consists of autoregressive sequence modeling methods, such as Decision Transformer (DT)~\cite{chen2021decision} and Prompt-DT~\cite{xu2022prompting}, which model the joint distribution over states, actions, and returns and generate actions autoregressively conditioned on the return.
The second class comprises non-autoregressive trajectory generation methods, typically based on diffusion models, such as Diffuser~\cite{janner2022planning} and Decision Diffuser (DD)~\cite{ajay2023conditional}.
They jointly model state dynamics and action distributions through a unified sampling-and-planning process, offering a flexible framework for incorporating environment- and task-level constraints beyond reward signals.
However, most generative approaches have focused on single-objective RL settings. 

In the context of MORL, PEDA~\cite{zhu2023scaling} extends single-objective return-conditioned models to multi-objective settings and introduces preference--return--conditioned generative methods, including MODT(P) and MORvS(P).
MODULI~\cite{yuan2024moduli} further improves generalization to out-of-distribution (OOD) preferences.
Despite these advances, generalization to unseen preference configurations remains limited, and learned multi-objective trade-off behaviors do not readily transfer across heterogeneous environment dynamics.
In this direction, and in line with recent efforts to bring foundation models to embodied decision-making~\cite{yang2024embodied,li2024muep}, our work advances diffusion-based generative policies toward a foundation-level Pareto policy, enabling a transferable multi-objective trade-off prior that improves generalization to OOD preferences and few-shot adaptation to unseen continuous-control tasks.

\section{Preliminaries}


\textbf{Multi-objective reinforcement learning (MORL).}
In MORL, the underlying multi-objective sequential decision problem is typically modeled as a multi-objective Markov decision process (MOMDP)
$\langle \mathcal{S}, \mathcal{A}, \mathcal{P}, \mathcal{R}, \gamma, \Omega \rangle$, where 
$\mathcal{S}=\{\bm{s}_t \}$ is the state space; 
$\mathcal{A}=\{\bm{a}_t \}$ is the action space; 
$\mathcal{P}(\bm{s}_{t+1} \mid \bm{s}_t, \bm{a}_t)$ is a transition function mapping the current state-action pair $(\bm{s}_t, \bm{a}_t)$ at time $t$ to a distribution over the next state $\bm{s}_{t+1}$ at time $t+1$; 
$\bm{r}_t=\mathcal{R}(\bm{s}_t, \bm{a}_t) \in \mathbb{R}^{D_o}$ is the vector-valued reward function and $D_o$ is the number of objectives; 
$\gamma \in [0,1)$ is a discount factor that controls the relative importance of future rewards;
$\Omega=\{\, \bm{\omega}=[\omega_i]_{i=1}^{D_o} \in \mathbb{R}^{D_o} \mid \omega_i \in [0,1],\ \sum_{i=1}^{D_o} \omega_i = 1 \,\}$ is the normalized preference space. \looseness-1

The goal of preference-conditioned MORL is to learn a policy $\pi(\bm{a}_t \mid \bm{s}_t, \bm{\omega})$ that maximizes the expected return vector
$\bm{G}^{\pi}=\mathbb{E}_{\bm{a}_t \sim \pi}[\sum_{t=0}^{T}\gamma^t \bm{r}_t]$, where $T$ denotes the episode termination time.
A policy $\pi_p$ is \emph{Pareto-optimal (undominated)} if there exists no policy $\pi_q$ such that
$\bm{G}^{\pi_q}_i \ge \bm{G}^{\pi_p}_i$ for all $i$ and $\bm{G}^{\pi_q}_j > \bm{G}^{\pi_p}_j$ for at least one $j$.
The set of all policies defines the \emph{decision space}, and the set of all achievable return vectors defines the \emph{objective space}.
The \emph{Pareto set (PS)} is the set of Pareto-optimal policies, and the corresponding set of objective vectors forms the \emph{Pareto front} ($\mathcal{F}$).

\textbf{Rectified flow} (RF) is a kind of \emph{flow matching}~\cite{liu2022flow,albergo2023building,lipman2023flow} method. 
Given a dataset $\mathcal{D}$ consisting of continuous data points $\bm{x}$ drawn from an unknown target data distribution $p_1$, RF aims to learn a transformation from a simple distribution $p_0$ to $p_1$.
This is achieved by modeling the data distribution evolution through an \emph{ordinary differential equation (ODE)} defined over the time domain $t~\sim P(t) \in [0,1]$:
\begin{equation}
    \frac{\mathrm{d}\bm{z}_t}{\mathrm{d}t}=v_\theta(\bm{z}_t,t), \quad \bm{z}_0 \sim p_0,
\end{equation}
where $\theta$ denotes the parameters of the velocity neural network $v_{\theta}(\cdot)$, which is used to characterize the data distribution evolution. In practice, $p_0$ is typically chosen as a standard Gaussian distribution $\mathcal{N}(0,\boldsymbol{I})$ due to its analytical tractability and ease of sampling.

In particular, the training objective of $v_{\theta}(\cdot)$ for the interpolated state $\bm{z}_t = t\bm{x} + (1-t)\bm{z}_0$ is: 
\begin{equation}
\label{eq:optimize_rf}
\min_{\theta} 
\mathbb{E}_{\substack{\bm{z}_0 \sim p_0,\, \bm{x} \sim p_1,\\ t \sim P(t)}}
\left[
\left\| v_\theta(\bm{z}_t, t) - (\bm{x} - \bm{z}_0) \right\|_2^2
\right]
\end{equation}

\textbf{Guidance-based conditional generation}.
Compared to generating arbitrary data from the distribution $p_1$, conditional generation aims to produce samples consistent with a specified property $\bm{c}$ and maximize the conditional probability $p(\bm{x}\mid \bm{c})$ for each data-condition pair $(\bm{x},\bm{c})$.
Two widely used approaches are the \emph{classifier guidance (CG)}~\cite{dhariwal2021diffusion} and the \emph{classifier-free guidance (CFG)}~\cite{ho2021classifier}.
CG trains an auxiliary classifier model $C_\phi(\bm{z}_t,t,\bm{c}) := \log p_\phi(\bm{c}\mid \bm{z}_t,t)$ parameterized by $\phi$ and updates the vector field with the guidance control weight $w^G$ at inference time:
\begin{equation}
\label{eq:cg}
\tilde{v}_{\theta}(\bm{z}_t,t\mid \bm{c}) = v_{\theta}(\bm{z}_t,t) + w^G \nabla_{\bm{z}_t} C_\phi(\bm{z}_t,t,\bm{c})
\end{equation}
CFG typically learns a single neural network to parameterize both the unconditional model $v_{\theta}(\bm{z}_t, t\mid\varnothing)$ and the conditional model $v_{\theta}(\bm{z}_t, t\mid \bm{c})$ by randomly setting $\bm{c} = \varnothing$ with probability $\eta$, where $\varnothing$ denotes the null condition.
The training objective is given by:
\begin{equation}
\mathcal{L}_{RF} = \mathbb{E} [ \| v_{\theta}(\bm{z}_t,t \mid \bm{c}) - (\bm{x} - \bm{z}_0) \|^2 ]
\end{equation}
At inference time, the guided vector field is obtained by:
\begin{equation}
\tilde{v}_{\theta}(\bm{z}_t,t\mid \bm{c}) = (1-w^G)v_{\theta}(\bm{z}_t,t\mid\varnothing) + w^G v_{\theta}(\bm{z}_t,t\mid \bm{c})
\end{equation}

\section{Method}
In this section, we introduce the problem setup (Sec.~\ref{sec:setup}), network architecture (Sec.~\ref{sec:archi}, Fig.~\ref{fig:network}), training pipeline (Sec.~\ref{Sec:3_stage_train}, Fig.~\ref{fig:training_stages}), and training objective (Sec.~\ref{Sec:training_objective}) of the \textbf{F}oundation \textbf{P}areto \textbf{F}low \textbf{P}olicy (\textbf{FP2}).
Together, these components address the generalization challenges of offline MORL, such as insufficient data coverage in exponentially growing multi-objective space, prevalence of suboptimal or noisy trajectories, and scarce high-quality Pareto-front data. 

\subsection{Problem setup}\label{sec:setup}
We consider an offline MORL setting with a static trajectory dataset
$\mathcal{D}=\{\tau^{i}\}$, where
$\tau^{i}=\{\bm{\omega}^{i}, \bm{s}^{i}_t, \bm{a}^{i}_t, \bm{r}^{i}_t\}_{t=0}^{T}$.
The goal is to learn a policy $\pi(\bm{a}_t \mid \bm{s}_t, \bm{\omega}, \bm{G}_t)$ conditioned on a preference vector $\bm{\omega}$ and the vector-valued return-to-go (RTG)
$\bm{G}_t=\sum_{k=t}^{T}\gamma^{k-t}\bm{r}_k$.
The policy can also be viewed as a conditional generative model over the joint distribution of state--action sequences, guided by $\bm{\omega}$ and the RTG condition.

\subsection{Unified framework overview} \label{sec:archi}

\begin{wrapfigure}[18]{r}{0.49\linewidth}
\vspace{-1.2em}
\centering
\includegraphics[width=\linewidth]{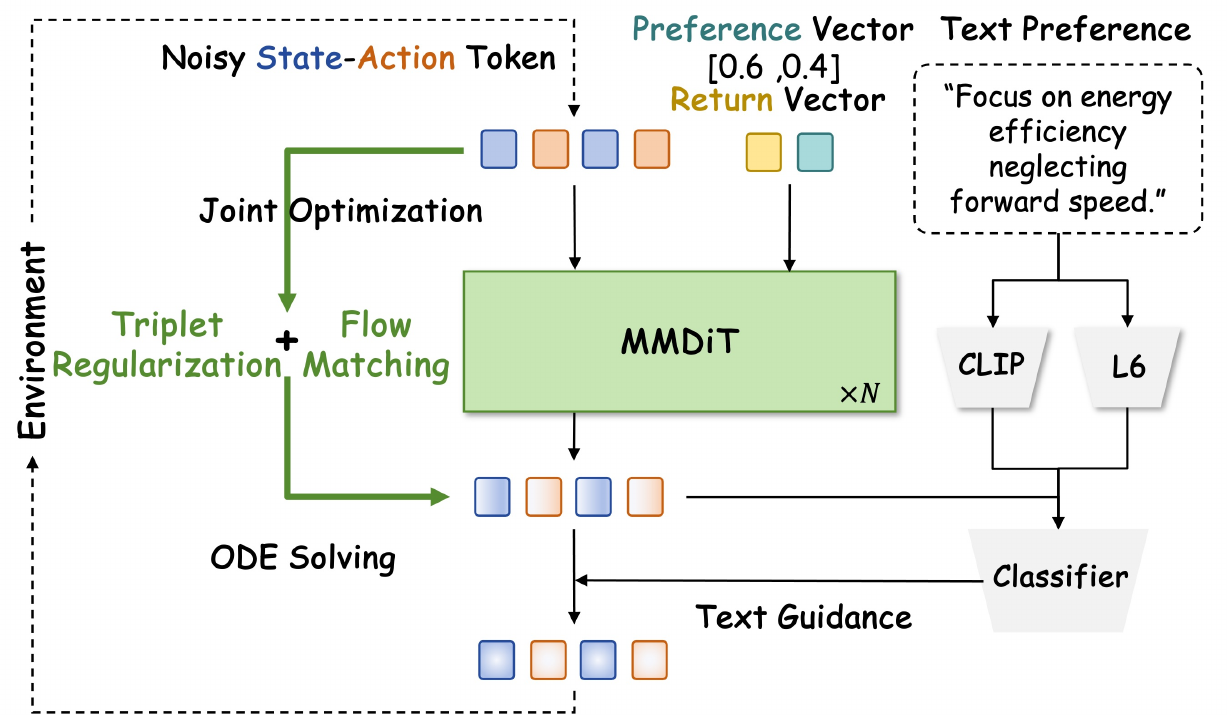}
\caption{Network architecture of \textbf{F}oundation \textbf{P}areto \textbf{F}low \textbf{P}olicy (\textbf{FP2}) for MORL.
FP2 is a rectified flow-based model with a multi-modal diffusion transformer (MMDiT) backbone that denoises state-action chunks and is guided by multi-modal conditions including numerical and textual preferences and multi-objective returns.
FP2 is trained offline and deployed as an online policy.}
\label{fig:network}
\vspace{-0.8em}
\end{wrapfigure}

\textbf{Generative backbone.}
Fig.~\ref{fig:network} illustrates the overall network architecture of FP2.
It adopts a rectified flow-based generative model parameterized by a multi-modal diffusion transformer (MMDiT)~\cite{peebles2023scalable},which provides a scalable sequence-modeling backbone for continuous trajectory generation.
MMDiT is conditioned on three signals: numerical and textual preferences, together with a task-specific return vector.

\textbf{Dual encoders for numerical and textual preferences.} We encode the numerical preference $\bm{\omega}$ and return vector $\bm{G}_t$ using a lightweight MLP, and incorporate the resulting tokens via classifier-free guidance (CFG).
To enable fuzzy control using textual preferences (e.g., ``run aggressively'' versus ``run economically''), we adopt dual text encoders following prior work~\cite{esser2024scaling,polyak2024movie}.
Specifically, we combine all-MiniLM-L6-v2 (L6)~\cite{reimers2019sentence} and CLIP~\cite{clip}: L6 captures fine-grained semantic representations, while CLIP emphasizes high-level, action-centered concepts.
Here, numerical and textual preferences play different roles: numerical inputs enable precise and efficient control through CFG, while textual preferences are encoded as semantic embeddings and incorporated through CG at inference time. 
This design avoids re-training the MMDiT backbone for novel or personalized textual instructions, allowing FP2 to leverage both exact numerical preferences and robust semantic instructions.
\raggedbottom
\setlength{\textfloatsep}{6pt plus 1pt minus 2pt}
\setlength{\floatsep}{6pt plus 1pt minus 2pt}
\setlength{\intextsep}{6pt plus 1pt minus 2pt}
\setlength{\abovecaptionskip}{2pt}
\setlength{\belowcaptionskip}{8pt}

\textbf{Return vector.}
Following prior work~\cite{chen2021decision}, we use the discounted return-to-go (RTG) vector as an additional condition to stabilize training.
At inference time, prior works adopt different RTG conditioning strategies, such as predicting returns from preferences~\cite{zhu2023scaling}.
Here, we set the RTG condition to the maximum normalized return vector $\bm{G}=\boldsymbol{1}^{D_o}$ using the multi-objective preference-aware return normalization in~\cite{yuan2024moduli}.

\subsection{Three-stage training pipeline} \label{Sec:3_stage_train}
Fig.~\ref{fig:training_stages} provides an overview of the proposed three-stage training pipeline. 
While multi-stage training is common in the era of foundation models, our goal is to adapt this recipe to offline MORL scenarios, which present some unique challenges.
First, data sampling in multi-objective spaces grows exponentially with the number of objectives, leading to insufficient coverage in practice and poor generalization.
For example, sampling 100 points along each objective dimension in an $n$-dimensional space yields $100^{n}$ samples, making full-space coverage impractical.
Second, offline MORL datasets' effective supervision for preference alignment remains sparse: actions in most trajectories are sub-optimal or noisy, while high-quality Pareto-front samples are scarce both in absolute number and relative proportion.
These motivate a staged design that first learns broad multi-objective control priors and then enhances fine-grained preference alignment.

\textbf{Data preparation.}
As shown in Fig.~\ref{fig:training_stages}, we derive all datasets from D4MORL~\cite{zhu2023scaling}, which contains six multi-objective robotic control environments.
We denote the full dataset as $\mathcal{D}_{\text{full}}$ and associate each trajectory with both a numerical preference vector and a textual preference description.
Since D4MORL does not provide language annotations, text prompts are deterministically generated from $\bm{\omega}$ using fixed templates (see Appendix~\ref{sec:appendix_common_setup}).
Following~\cite{yuan2024moduli}, we split $\mathcal{D}_{\text{full}}$ into $\mathcal{D}_{\text{pre}}$ by retaining $70\%$ of the data, with the remaining $30\%$ forming $\mathcal{D}_{\text{drop}}$.
From $\mathcal{D}_{\text{pre}}$, we further construct the alignment dataset $\mathcal{D}_{\text{mid}}$ by retaining trajectories from the top two fronts identified via non-dominated sorting.
These fronts are dataset-relative and do not imply ground-truth optimality, which may be unavailable in practice.
For cross-task experiments, we train on five environments and evaluate on the held-out environment, denoted as $\mathcal{D}_{\text{task}}$.

\textbf{Stage 1: Pre-training.}
FP2 is pre-trained on the large-scale dataset $\mathcal{D}_{\text{pre}}$ containing both Pareto-optimal and suboptimal trajectories.
This stage focuses on learning fundamental multi-objective trade-off behaviors, facilitating faster policy learning in later stages. 

\textbf{Stage 2: Mid-training.}
While pre-training yields a strong multi-objective representation prior, sub-optimal and noisy trajectories in Stage~1 can weaken preference--behavior alignment, motivating a tailored mid-training stage to enhance fine-grained alignment. More importantly, Stage~2 does not require ground-truth Pareto-optimal trajectories; instead, it uses dataset-relative high-quality trajectories to fine-tune FP2 and reduces the potential preference-behavior inconsistency introduced in Stage~1.
Formally, we define preference alignment as follows.

\begin{definition}[Preference Alignment]
\label{def:alignment}
Let $\bm{\omega}_{-}, \bm{\omega}, \bm{\omega}_{+} \in \mathbb{R}^{D_o}$ be preference vectors such that
$\bm{\omega}_{-} \preceq_{\omega} \bm{\omega} \preceq_{\omega} \bm{\omega}_{+}$,
where $\preceq_{\omega}$ denotes a component-wise partial order in preference space.
Let $\bm{y}^*(\bm{\omega}_{-}), \bm{y}^*(\bm{\omega}), \bm{y}^*(\bm{\omega}_{+}) \in \mathcal{F}$ denote  objective outcomes on the Pareto frontier $\mathcal{F}$.
A predicted frontier point $\hat{\bm{y}}$ for $\bm{y}^*(\bm{\omega})$ is said to be \textbf{aligned} with preference $\bm{\omega}$ if
$\bm{y}^*(\bm{\omega}_{-}) \preceq \hat{\bm{y}} \preceq \bm{y}^*(\bm{\omega}_{+})$,
where $\preceq$ denotes a partial order (e.g., Pareto dominance) over objective space.
\end{definition}

Accordingly, Stage~2 is trained on the dataset $\mathcal{D}_{\text{mid}}$.
This design retains local structure along the Pareto manifold, encouraging consistent and monotonic preference--behavior mappings.

\textbf{Stage 3: Post-training.}
In the final stage, FP2 is adapted to unseen preferences or new environments using a small number of trajectories drawn from either the new preference dataset $\mathcal{D}_{\text{drop}}$ or the new task dataset $\mathcal{D}_{\text{task}}$.
To avoid catastrophic forgetting, we adopt a parameter-efficient post-training method that only finetunes the MMDiT backbone while freezing all remaining modules.

\subsection{Training objective}
\label{Sec:training_objective}
The three stages in our training pipeline use a unified objective comprising the rectified flow matching loss, a triplet regularization term, and a post-training penalty:
\begin{equation}
\mathcal{L} = \mathcal{L}_{\text{RF}} + \lambda_{\text{tri}} \mathcal{L}_{\text{triplet}} + \boldsymbol{1}_{\text{post}} \mathcal{L}_{\text{weight}}
\end{equation}
where $\lambda_{\text{tri}}$ controls the regularization strength, and $\boldsymbol{1}_{\text{post}}=1$ during post-training and 0 otherwise. 

\textbf{Rectified flow matching loss.}
Given a pair of samples $(\bm{z}_0, \bm{x}) \in \mathcal{D}$ and an interpolation time $t \sim \mathcal{U}(0,1)$, rectified flow training minimizes the discrepancy between the predicted velocity and the target displacement $\bm{x} - \bm{z}_0$:
\begin{equation}
\label{eq:loss_rf}
\mathcal{L}_{\text{RF}} =
\mathbb{E}_{\bm{z}_0, \bm{x}, t}
\left[
\left\lVert
v_\theta(\bm{z}_t, t \mid \bm{c}) - (\bm{x} - \bm{z}_0)
\right\rVert_2^2
\right],
\end{equation}
where $\bm{z}_t = (1-t)\bm{z}_0 + t \bm{x}$, and $\bm{c}=[\bm{\omega}, \bm{G}_t]$.

\textbf{Triplet velocity regularization.} Preferences lie in a continuous space, where nearby preferences should induce similar yet distinguishable behaviors.
However, these fine-grained behavioral relations can be easily misaligned or incorrectly ordered during training. To address this, we introduce a velocity-space regularization that constructs hard negatives from nearest neighbors in preference space, yielding a triplet formulation that enforces both similarity and distinguishability.
For a mini-batch sample $b\sim \mathcal{B}$, we define the positive distance as 
$d_{\text{pos}}^{(b)} = \frac{1}{D}\|\bm{v}_{\text{pred}}^{(b)} - \bm{v}_{\text{tgt}}^{(b)}\|_2^2$ with $\bm{v}_{\text{tgt}}=\bm{x}-\bm{z}_0$ as in Eq. (\ref{eq:loss_rf}), 
and select the nearest neighbor in preference space 
$b' = \arg\min_{j \neq b} \|\bm{\omega}^{(j)} - \bm{\omega}^{(b)}\|_2$, 
yielding the negative distance 
$d_{\text{neg}}^{(b)} = \frac{1}{D}\|\bm{v}_{\text{pred}}^{(b)} - \bm{v}_{\text{pred}}^{(b')}\|_2^2$. 
Finally, the triplet loss is defined as:
\begin{equation}
\mathcal{L}_{\text{triplet}} =
\mathbb{E}_{b \sim \mathcal{B}}
\left[
\max\left(0,\;
d_{\text{pos}}^{(b)} - \lambda\, d_{\text{neg}}^{(b)}
\right)
\right],
\end{equation}
where $\lambda>0$ controls the margin, and the loss focuses on hard samples when $d_{\text{neg}}^{(b)} < d_{\text{pos}}^{(b)} / \lambda$.

\textbf{Post-training penalty.}
During post-training, we apply an $\ell_2$ regularization term to mitigate catastrophic forgetting:
\begin{equation}
\label{eq:loss_weight}
\mathcal{L}_{\text{weight}}=\|\theta-\theta_{\text{mid}}\|_2^2 .
\end{equation}
Here, $\theta$ denotes the post-trained model weights, and $\theta_{\text{mid}}$ denotes the weights after mid-training.

\section{Experiments}
In this section, we first evaluate the effectiveness of FP2 and its three-stage training pipeline on task-specific locomotion tasks (Sec.~\ref{sec:exp_task_spec_pre_mid} and Sec.~\ref{sec:exp_task_spec_post_train}). We then examine its foundational learning and transfer capabilities in cross-task settings (Sec.~\ref{sec:exp_cross_task}), followed by ablation studies (Sec.~\ref{sec:exp_ablation}).

\textbf{Dataset.}
We evaluate our proposed method on the D4MORL dataset~\cite{zhu2023scaling}, a benchmark specifically designed for offline MORL, as summarized in Table~\ref{tab:D4MORL}.
It comprises 1.8 million annotated demonstrations across six multi-objective (MO) MuJoCo task environments, each featuring two to three objectives. 
For each environment, D4MORL provides two datasets, \texttt{Expert} and \texttt{Amateur}, with

\begin{wraptable}[10]{hr}{0.48\linewidth}
\vspace{-15pt}
\centering
\captionsetup{
    width=\linewidth,
    justification=raggedright,
    singlelinecheck=false
}
\caption{Summary of the D4MORL dataset.}
\label{tab:D4MORL}
\vspace{0.1em}

\begingroup
\scriptsize
\setlength{\tabcolsep}{0.8pt}
\renewcommand{\arraystretch}{1.02}

\begin{tabular*}{\linewidth}{@{\extracolsep{\fill}}l|cccccc@{}}
\toprule
TASK  & \#OBJ & S-D & A-D & STEP-E/A & \#OPT-E/A & \#PRE  \\
\midrule
Ant         & 2 & 27 & 8 & 500/500 & 102/65    & \hbarwide{0.2391}{0.2979}{0.7021}{0.7609} \\ 
HalfCheetah & 2 & 17 & 6 & 499/482 & 1547/459  & \hbarwide{0.1668}{0.3768}{0.6232}{0.8332} \\
Hopper      & 2 & 11 & 3 & 499/387 & 223/52    & \hbarwide{0.3208}{0.4353}{0.5647}{0.6792} \\
Walker2d    & 2 & 17 & 6 & 499/442 & 2055/411  & \hbarwide{0.1668}{0.2583}{0.7417}{0.8332} \\
Swimmer     & 2 & 8  & 2 & 500/500 & 1475/58   & \hbarwide{0.0774}{0.0950}{0.9050}{0.9226} \\
Hopper-3obj & 3 & 11 & 3 & 499/442 & 8945/1925 & \hbarwide{0.1511}{0.0007}{0.6078}{0.5919}  \\
\bottomrule
\end{tabular*}

\vspace{0.2em}
\parbox{\linewidth}{%
\raggedright\tiny
\textit{Note.} 
\#OBJ: number of objectives.
S/A-D: state/action dimensions. 
STEP-E/A and \#OPT-E/A: average episode length and Pareto-front trajectories for \texttt{Expert}/\texttt{Amateur}. \#PRE: range of preferences; only the first two dimensions are shown for 3-objective tasks.
}

\endgroup
\vspace{-35pt}
\end{wraptable}

the latter serving as a noisier counterpart of the former; together, they correspond to the full dataset $\mathcal{D}_{\text{full}}$.
Furthermore, to assess generalization, we adopt the \texttt{Shattered} and \texttt{Narrow}
preference-deficient configurations $\mathcal{D}_{\text{pre}}$ from MODULI~\cite{yuan2024moduli}, which remove $30\%$ of the data.

\textbf{Baselines.}
We compare our pre-trained model (\textbf{FP2-Pre}), mid-trained model (\textbf{FP2-Mid}),
post-trained model (\textbf{FP2-Post}), and foundation model (\textbf{FP2-Fou}) trained on five MO task environments, with strong offline MORL baselines:
(1) MODT(P)~\cite{chen2021decision,zhu2023scaling}, a Decision Transformer-based approach;
(2) MORvS(P)~\cite{emmons2022rvs,zhu2023scaling}, a supervised reinforcement learning method;
(3) BC(P)~\cite{emmons2022rvs,zhu2023scaling}, a behavior cloning algorithm;
(4) CQL(P)~\cite{kumar2020conservative,zhu2023scaling}, a conservative Q-learning variant; and
(5) MODULI~\cite{yuan2024moduli}, the current state-of-the-art offline MORL algorithm.
Since COLA~\cite{li2026cola} and Q-chunking~\cite{li2026reinforcement} are designed for online and offline-to-online RL, respectively, we report comparisons with their fixed-data adaptations separately in Appendix~\ref{app:fixed_data_baselines}.

\textbf{Metric.} To comprehensively assess the performance of FP2 and baselines, we adopt metrics capturing Pareto-front convergence and diversity, as well as preference alignment.
\textbf{(1)} Hypervolume (\textbf{HV})$\uparrow$~\cite{zitzler2002multiobjective, auger2012hypervolume}: HV is widely-used to measure the volume of the objective space dominated by the obtained Pareto front solutions $\mathcal{F}$.
$\mathcal{HV}(\mathcal{F}) = \int_{\mathbb{R}^{D_o}} \boldsymbol{1}_{H(\mathcal{F})}(\bm{y}) \mathrm{d}y$, 
where $D_o$ is the number of objectives, 
\( H(\mathcal{F}) = \{ \bm{y} \in \mathbb{R}^{D_o} \mid \exists \bm{y}' \in \mathcal{F}, \bm{y}_0 \preceq \bm{y} \preceq \bm{y}' \} \), \( \bm{y}_0 \) is a reference point in the objective space ($\boldsymbol{0}^{D_o}$ for maximization problems), and $\bm{y}_0 \preceq \bm{y}$ means $\bm{y}_0$ weakly dominates $\bm{y}$, $\boldsymbol{1}_{H(\mathcal{F})}(\bm{y})=1$ iff $\bm{y} \in H(\mathcal{F})$ and 0 otherwise.
A higher HV is better.
\textbf{(2)} Sparsity (\textbf{SP})$\downarrow$~\cite{zhu2023scaling}: 
SP indicates how sparsely the obtained solutions cover the Pareto front:
$\mathcal{SP}(\mathcal{F})=\frac{1}{|\mathcal{F}|} \sum_{i=1}^{n} \sum_{k=1}^{|\mathcal{F}|-1} (\widetilde{\mathcal{F}}_i(k) - \widetilde{\mathcal{F}}_i(k+1))^2$,
where $\widetilde{\mathcal{F}}_i$ is the list of Pareto front solutions sorted by the $i$-th objective value.
A lower SP is better.
\textbf{(3)} Preference alignment score (\textbf{PAS})$\uparrow$: 
We propose a cosine similarity-based preference alignment metric PAS. 
For a preference vector $\bm{\omega}$, a reference Pareto frontier point  $\bm{y}_{\text{gt}}$, and a predicted point $\bm{y}_{\text{pred}}$, it is defined as:
$\mathcal{PAS} = \frac{1}{2} \left( 1 + \frac{(\bm{y}_{\text{gt}} - \bm{\omega}) \cdot (\bm{y}_{\text{pred}} - \bm{\omega})}{\left\| \bm{y}_{\text{gt}} - \bm{\omega} \right\|_2 \cdot \left\| \bm{y}_{\text{pred}} - \bm{\omega} \right\|_2} \right) \in [0,1]$, where larger values indicate better alignment.






\subsection{Effective preference generalization via pre- and mid-training}
\label{sec:exp_task_spec_pre_mid}

\begin{figure}[t]
\centering
\begin{subfigure}[t]{0.46\linewidth}
    \centering
    \includegraphics[width=\linewidth]{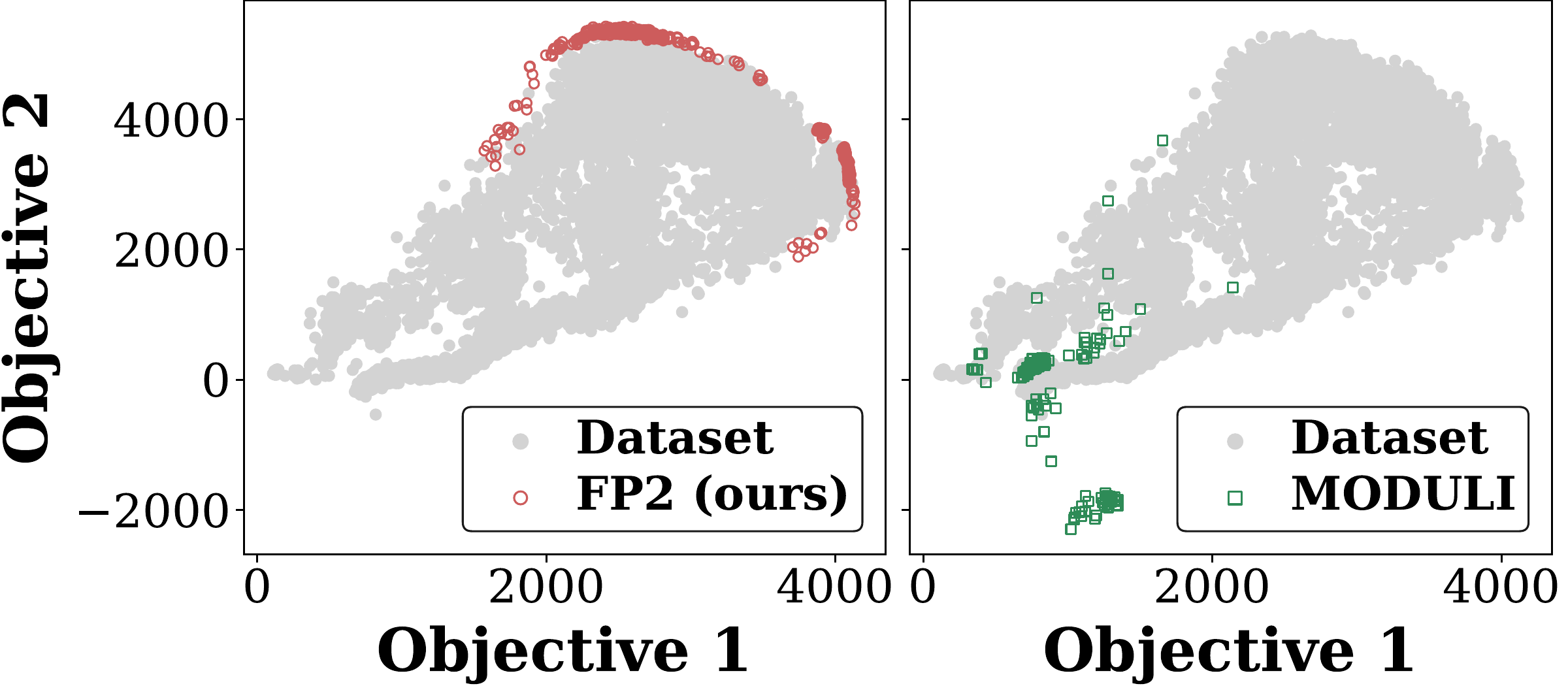}
\end{subfigure}
\hfill
\begin{subfigure}[t]{0.46\linewidth}
    \centering
    \includegraphics[width=\linewidth]{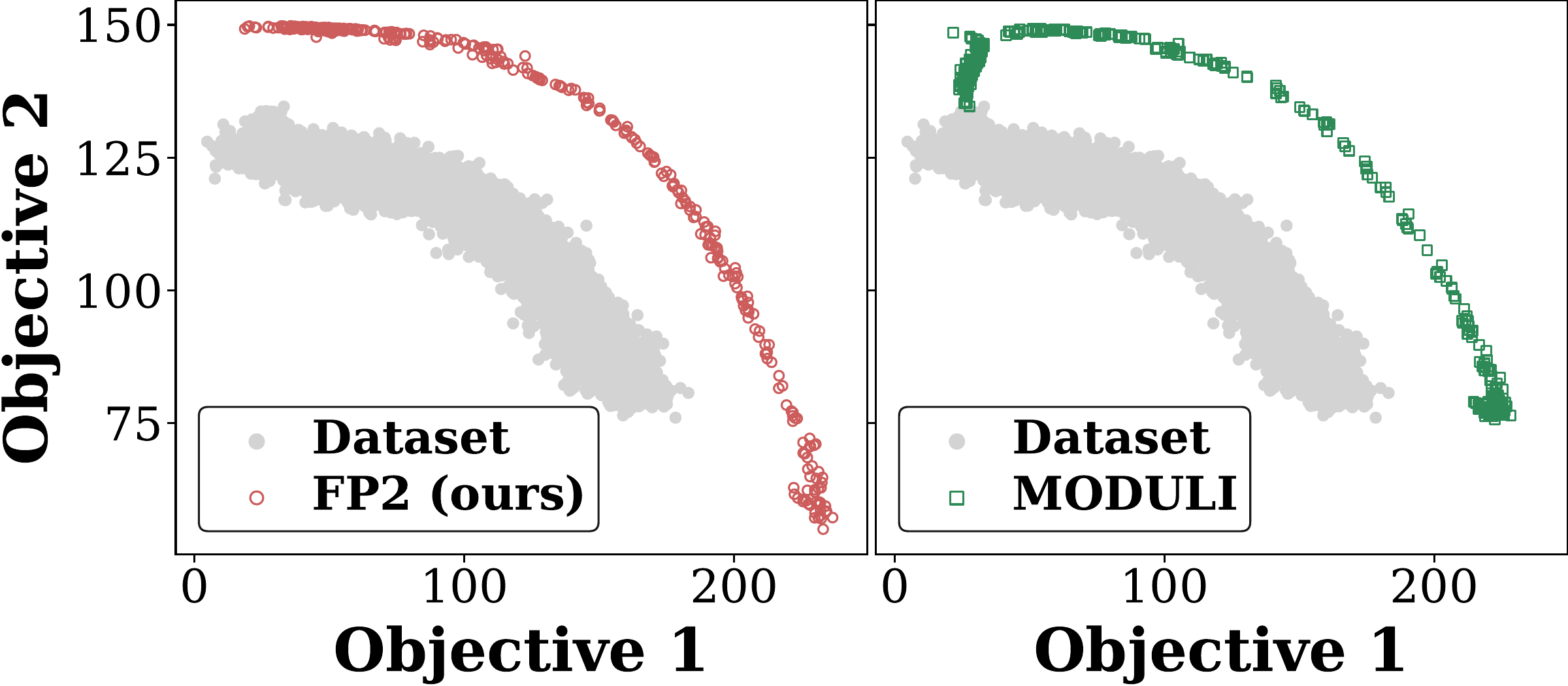}
\end{subfigure}
\caption{Preference generalization, with OOD points in color and dataset points in gray.}
\label{fig:ood_pref_visualize}
\end{figure}

Table~\ref{tab:pref_gen_compact} compares FP2-Pre, FP2-Mid, and prior methods on \texttt{Expert}-\texttt{Shattered}/\texttt{Narrow}. FP2-Mid outperforms MODULI on $11/12$ tasks in HV and $10/12$ in SP. 
The gain from FP2-Pre to FP2-Mid on most tasks highlights the role of mid-training in enhancing preference–behavior alignment from dataset-relative high-quality trajectories.
Fig.~\ref{fig:ood_pref_visualize} further shows improved interpolation and extrapolation of FP2-Mid on Pareto manifolds (Hopper on Amateur-Shattered and Swimmer on Amateur-Narrow).
We summarize all 36 comparisons in Appendix~\ref{app:add_results}: FP2-Mid achieves median gains of +0.98\% on HV, +29.44\% on SP, and +1.04\% on PAS, and favorable Win/Tie/Loss results of 30/3/3, 29/4/3, and 24/9/3, respectively (Table~\ref{tab:full_compare_36_statistics}). 
The gains are larger in SP than in HV, indicating that FP2 mainly improves the coverage and uniformity of the obtained front, while HV, bounded by the quality of the offline data, is maintained or modestly improved.

\begin{table*}[htbp]
\centering
\scriptsize
\setlength{\tabcolsep}{2.4pt}
\renewcommand{\arraystretch}{0.86}
\caption{
Task-specific preference generalization on \texttt{Expert}-\texttt{Shattered} (S) and \texttt{Expert}-\texttt{Narrow} (N).
We report average scores over three seeds; full mean $\pm$ std results are provided in Appendix~\ref{app:full_resutls}. 
}
\label{tab:pref_gen_compact}
\vspace{-0.4em}
\resizebox{\textwidth}{!}{
\begin{tabular}{llcccccc|cccccc}
\toprule
Metric & Method
& Ant-S & HalfC.-S & Hopper-S & Walker-S & Swim.-S & Hopper3-S
& Ant-N & HalfC.-N & Hopper-N & Walker-N & Swim.-N & Hopper3-N \\
\midrule

\multirow{6}{*}{HV$\uparrow$}
& MODT(P)  & 5.88 & 5.69 & 1.95 & 5.01 & 3.20 & 2.83 & 5.05 & 5.04 & 1.85 & 4.75 & 3.03 & 3.18 \\
& MORvS(P) & \cellcolor{gray!15}\textbf{6.38} & 5.73 & 2.06 & 5.14 & 3.24 & 3.28 & 6.06 & 5.46 & 1.98 & 4.85 & 3.10 & 3.32 \\
& BC(P)    & 4.76 & 5.66 & 1.49 & 3.47 & 3.19 & 1.95 & 4.90 & 4.88 & 1.29 & 1.88 & 3.06 & 2.32 \\
& MODULI   & \cellcolor{gray!15}\textbf{6.38} & 5.78 & 2.07 & 5.20 & 3.24 & 3.33 & 6.18 & 5.74 & 2.07 & 5.13 & 3.16 & 3.15 \\
& FP2-Pre  & 6.17 & 5.76 & 2.01 & 5.16 & 3.22 & \cellcolor{gray!15}\textbf{3.80} & 6.06 & 5.74 & 2.01 & 5.10 & 3.16 & 3.39 \\
& FP2-Mid  & 6.35 & \cellcolor{gray!15}\textbf{5.79} & \cellcolor{gray!15}\textbf{2.08} & \cellcolor{gray!15}\textbf{5.22} & \cellcolor{gray!15}\textbf{3.26} & 3.70 & \cellcolor{gray!15}\textbf{6.24} & \cellcolor{gray!15}\textbf{5.79} & \cellcolor{gray!15}\textbf{2.10} & \cellcolor{gray!15}\textbf{5.22} & \cellcolor{gray!15}\textbf{3.26} & \cellcolor{gray!15}\textbf{3.81} \\

\midrule

\multirow{6}{*}{SP$\downarrow$}
& MODT(P)  & 2.22 & 0.18 & 0.85 & 0.49 & 15.50 & 0.09 & 0.88 & 0.04 & 0.34 & 0.34 & 3.65 & 0.11 \\
& MORvS(P) & 0.75 & 0.15 & 0.20 & 0.27 & 7.36 & 0.06 & 0.79 & 0.05 & 0.16 & 0.19 & 4.38 & 0.22 \\
& BC(P)    & 5.16 & 0.34 & 2.58 & 16.61 & 9.05 & 0.17 & 1.56 & \cellcolor{gray!15}\textbf{0.02} & 0.59 & 10.76 & 5.53 & 0.14 \\
& MODULI   & 0.53 & 0.11 & 0.13 & 0.11 & 5.63 & 0.10 & 0.96 & 0.04 & 0.28 & 0.14 & 5.84 & 0.16 \\
& FP2-Pre  & 1.15 & 0.13 & \cellcolor{gray!15}\textbf{0.12} & 0.12 & 7.11 & 0.07 & 0.82 & 0.05 & 0.47 & 0.12 & 6.12 & \cellcolor{gray!15}\textbf{0.03} \\
& FP2-Mid  & \cellcolor{gray!15}\textbf{0.51} & \cellcolor{gray!15}\textbf{0.07} & 0.13 & \cellcolor{gray!15}\textbf{0.05} & \cellcolor{gray!15}\textbf{2.46} & \cellcolor{gray!15}\textbf{0.05} & \cellcolor{gray!15}\textbf{0.75} & 0.05 & \cellcolor{gray!15}\textbf{0.08} & \cellcolor{gray!15}\textbf{0.06} & \cellcolor{gray!15}\textbf{1.68} & 0.05 \\

\midrule

\multirow{3}{*}{PAS$\uparrow$}
& MODULI  & 0.94 & \cellcolor{gray!15}\textbf{0.99} & 0.43 & \cellcolor{gray!15}\textbf{0.99} & \cellcolor{gray!15}\textbf{0.99} & 0.98 & \cellcolor{gray!15}\textbf{0.93} & \cellcolor{gray!15}\textbf{0.99} & 0.92 & 0.93 & \cellcolor{gray!15}\textbf{0.99} & 0.96 \\
& FP2-Pre & 0.92 & \cellcolor{gray!15}\textbf{0.99} & \cellcolor{gray!15}\textbf{0.92} & 0.92 & 0.92 & 0.95 & 0.90 & \cellcolor{gray!15}\textbf{0.99} & 0.90 & 0.92 & 0.89 & 0.91 \\
& FP2-Mid & \cellcolor{gray!15}\textbf{0.96} & \cellcolor{gray!15}\textbf{0.99} & 0.56 & \cellcolor{gray!15}\textbf{0.99} & \cellcolor{gray!15}\textbf{0.99} & \cellcolor{gray!15}\textbf{0.99} & \cellcolor{gray!15}\textbf{0.92} & 0.98 & \cellcolor{gray!15}\textbf{0.96} & \cellcolor{gray!15}\textbf{0.98} & \cellcolor{gray!15}\textbf{0.99} & \cellcolor{gray!15}\textbf{0.98} \\
\bottomrule
\end{tabular}
}
\vspace{0.25em}
\parbox{\linewidth}{%
\scriptsize
\textit{Note.} HV/SP are rescaled for readability. 
HV: $\times10^6$ for Ant, HalfC., Walker; $\times10^7$ for Hopper; 
$\times10^4$ for Swim.; $\times10^{10}$ for Hopper3. 
SP: $\times10^4$ for Ant, HalfC., Swim.; $\times10^5$ for Hopper, Hopper3;  $\times10^0$ for Walker.
HalfC.: HalfCheetah;
Swim.: Swimmer;
Hopper3: Hopper-3obj.
}
\end{table*}

\subsection{Fast adaptation to new preferences via post-training}
\label{sec:exp_task_spec_post_train}
\begin{figure}[t]
\centering
\includegraphics[width=\linewidth]{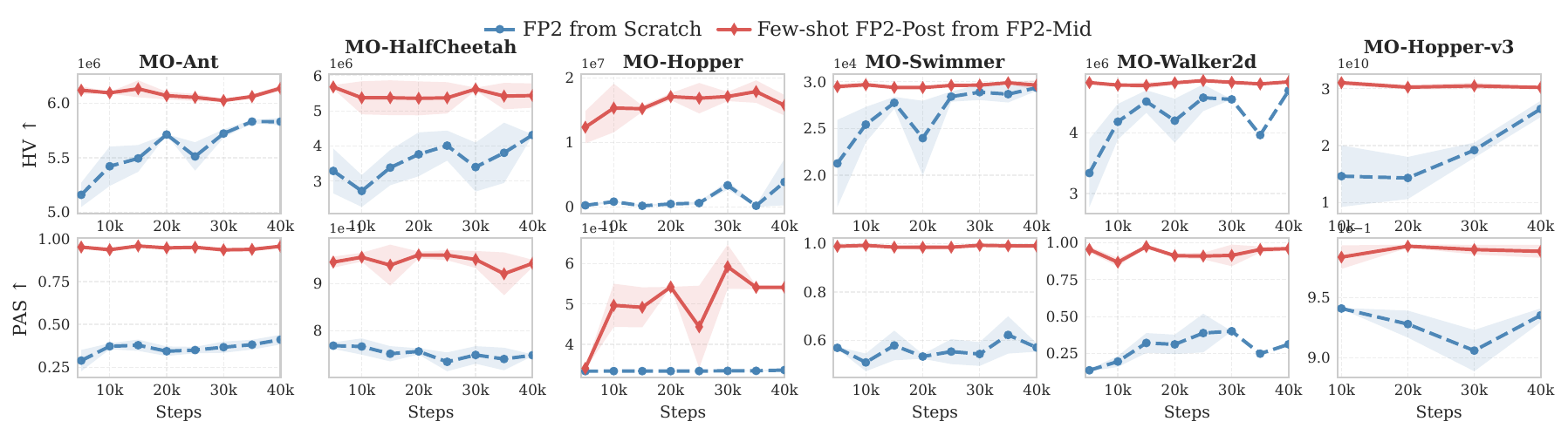}
\caption{
Training performance comparison for out-of-distribution (OOD) preference generalization in Table~\ref{tab:fewshot_main}: FP2 trained from scratch vs. few-shot FP2-Post initialized from FP2-Mid.
}
\label{fig:post_vs_scratch}
\vspace{-6pt}
\end{figure}

We evaluate few-shot adaptation to unseen preference configurations using a fixed budget of 1K gradient steps.
Specifically, we adapt on Pareto-front samples from the $30\%$ held-out preference regions in the \texttt{Expert}/\texttt{Amateur}-\texttt{Shattered}/\texttt{Narrow} datasets, and all metrics are computed on these held-out regions.
Table~\ref{tab:fewshot_main} compares FP2 trained from scratch (Scratch), zero-shot FP2-Mid, and post-trained FP2 (FP2-Post).
FP2-Post outperforms Scratch in HV and PAS on all six tasks, often by a large margin; notably, this all-task advantage over Scratch is also observed for zero-shot FP2-Mid, indicating that pre- and mid-training provide a strong initialization for fast adaptation.
Fig.~\ref{fig:post_vs_scratch} further shows that FP2-Post converges faster than Scratch.
Post-training does not always improve over zero-shot FP2-Mid; for example, HV slightly decreases on MO-Hopper (1.95 to 1.89).

\begin{table}[t]
\centering
\footnotesize
\setlength{\tabcolsep}{3pt}
\caption{
Few-shot post-training (1K steps) on the held-out $30\%$ preference regions of \texttt{Expert}-\texttt{Shattered}; all metrics are computed on the held-out regions.
}
\label{tab:fewshot_main}
\begin{tabular*}{\linewidth}{@{\extracolsep{\fill}}lccccccccc@{}}
\toprule
\multirow{2}{*}{Environment}
& \multicolumn{3}{c}{Scratch}
& \multicolumn{3}{c}{FP2-Mid (Zero-shot)}
& \multicolumn{3}{c}{FP2-Post (Few-shot)} \\
\cmidrule(lr){2-4} \cmidrule(lr){5-7} \cmidrule(lr){8-10}
& HV $\uparrow$ & SP $\downarrow$ & PAS $\uparrow$
& HV $\uparrow$ & SP $\downarrow$ & PAS $\uparrow$
& HV $\uparrow$ & SP $\downarrow$ & PAS $\uparrow$ \\
\midrule
MO-Ant         & 5.04 & 3.53 & 0.36 & 5.70 & 3.37 & 0.85 & \textbf{5.83} & \textbf{1.69} & \textbf{0.91} \\
MO-HalfCheetah & 4.00 & 0.34 & 0.79 & 5.69 & 0.09 & \textbf{0.98} & \textbf{5.73} & \textbf{0.08} & 0.97 \\
MO-Hopper      & 0.91 & \textbf{0.07} & 0.28 & \textbf{1.95} & 0.14 & 0.54 & 1.89 & 0.13 & \textbf{0.60} \\
MO-Walker2d    & 4.80 & 1.21 & 0.17 & 4.87 & 0.20 & \textbf{0.99} & \textbf{5.05} & \textbf{0.15} & \textbf{0.99} \\
MO-Swimmer     & 2.79 & 71.33 & 0.67 & 3.03 & 11.13 & \textbf{0.99} & \textbf{3.09} & \textbf{9.15} & \textbf{0.99} \\
MO-Hopper-3obj & 2.13 & 0.10 & 0.70 & 3.63 & 0.60 & \textbf{0.97} & \textbf{3.66} & \textbf{0.03} & \textbf{0.97} \\
\bottomrule
\end{tabular*}
\vspace{-6pt}
\end{table}

\subsection{Cross-task transfer of a reusable trade-off prior}
\label{sec:exp_cross_task}

We evaluate whether FP2 can learn a reusable multi-objective trade-off prior that transfers across environments.
To cover different locomotion characteristics, we adopt a systematic leave-one-task-out protocol: for each target environment, the model is trained on the other five multi-objective MuJoCo tasks and then adapted to the held-out task with a fixed budget of 1K gradient steps.
This out-of-distribution (OOD) task generalization setting is more challenging than OOD preference generalization because the target environment has different dynamics and is never observed during training.
The FP2 model trained on mixed-environment data is denoted as \textbf{FP2-Fou}, and its post-trained variant as \textbf{FP2-Fou-Post}.

\begin{wrapfigure}[11]{r}{0.50\linewidth}
\centering
\includegraphics[width=\linewidth]{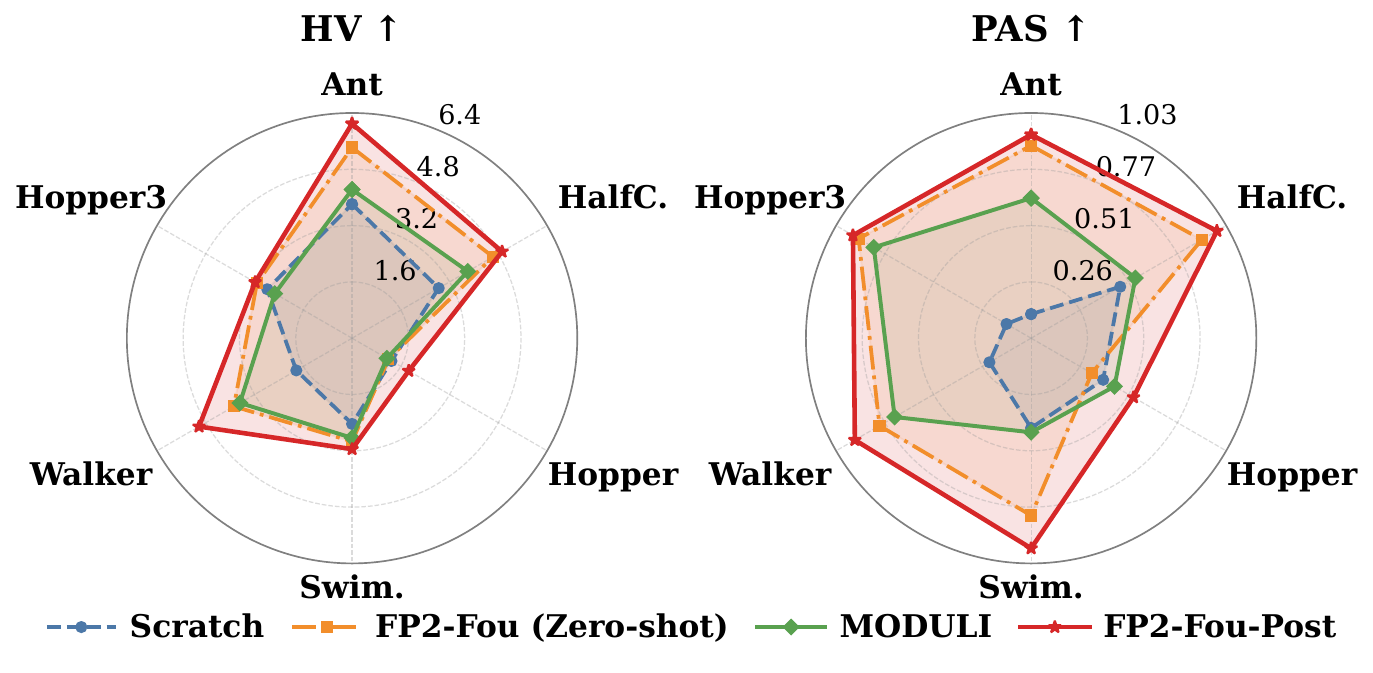}
\caption{Cross-task performance comparison.}
\label{fig:cross_task}
\end{wrapfigure}

For comparison, we adapt MODULI to the cross-task setting using the same zero-padding, feature/temporal masking, mixed-environment sampling, leave-one-task-out split, target-task data, and 1K-step budget as FP2.
Fig.~\ref{fig:cross_task} illustrates the adaptation behavior of FP2-Fou-Post, training from scratch (Scratch), zero-shot FP2-Fou, and MODULI, while Table~\ref{tab:cross_task_transfer} reports quantitative results.
Among these comparisons, FP2-Fou-Post vs.\ Scratch isolates the effect of pre- and mid-training, FP2-Fou vs.\ FP2-Fou-Post isolates post-training, and MODULI serves as an algorithmic offline-MORL reference.
FP2-Fou-Post consistently outperforms all baselines, demonstrating the effective cross-task transfer capability of FP2.
As an exploratory extension beyond locomotion, Appendix~\ref{app:metaworld} further evaluates zero-shot FP2 on ten multi-objective Meta-World manipulation environments~\cite{yu2020meta} with control-level speed and cost objectives.

\subsection{Ablation}
\label{sec:exp_ablation}

\begin{figure}[t]
\centering
\includegraphics[width=\linewidth]{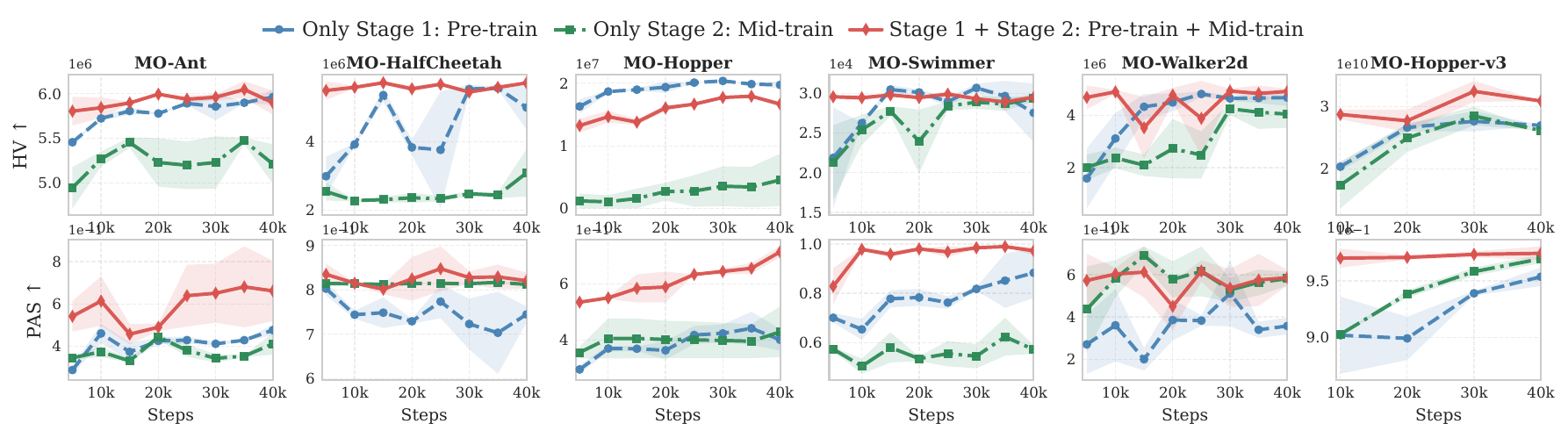}
\caption{Ablation study. Pre- and mid-training together achieve the best and fastest convergence, demonstrating the effectiveness of the multi-stage training paradigm.}
\label{fig:stages_ablation}
\vspace{-6pt}
\end{figure}

\textbf{Multi-stage training paradigm.}
Fig.~\ref{fig:stages_ablation} compares training-stage variants to isolate their contributions.
Stage~1 provides a strong generative prior; Stage~2 alone is constrained by its limited data coverage.
Their combination best improves preference--behavior alignment and performs strongest overall.

\textbf{Generative backbone.}
To isolate the effect of Rectified Flow (RF), Table~\ref{tab:ablation_diff_pref_loss} compares RF with DDIM~\cite{song2021denoising} under the same training stages (Pre and Pre+Mid) on Expert-Shattered. 
The two backbones perform comparably overall, with the clearest RF gain on MO-Hopper-3obj after pre-training (3.80 vs.\ 3.37); mid-training improves DDIM on all six tasks and RF on five, indicating that FP2's gains mainly arise from the staged training pipeline rather than from RF alone.

\begin{table}[t]
\centering
\small
\setlength{\tabcolsep}{3pt}
\caption{
Ablation on (1) generative backbone, (2) preference guidance, (3) contrastive learning, and (4) post-training strategy.
HV is reported in environment-specific scales.
}
\label{tab:ablation_diff_pref_loss}
\label{tab:ablation_finetuning}
\begin{tabular*}{\linewidth}{@{\extracolsep{\fill}}clcccccc@{}}
\toprule
& Method & Ant & HalfCheetah & Hopper & Swimmer & Walker2d & Hopper-3obj \\
\midrule
\multirow{4}{*}{(1)}
& \textbf{FP2-Pre} & $\mathbf{6.17}$ & $5.76$ & $\mathbf{2.01}$ & $3.22$ & $5.16$ & $\cellcolor{gray!15}\mathbf{3.80}$ \\
& $\circlearrowright$ DDIM-Pre & $6.15$ & $\mathbf{5.77}$ & $2.00$ & $\mathbf{3.24}$ & $\mathbf{5.17}$ & $3.37$ \\
& \textbf{FP2-Mid} & $6.35$ & $\cellcolor{gray!15}\mathbf{5.79}$ & $\cellcolor{gray!15}\mathbf{2.08}$ & $\cellcolor{gray!15}\mathbf{3.26}$ & $\cellcolor{gray!15}\mathbf{5.22}$ & $\mathbf{3.70}$ \\
& $\circlearrowright$ DDIM-Mid & $\cellcolor{gray!15}\mathbf{6.37}$ & $\cellcolor{gray!15}\mathbf{5.79}$ & $\cellcolor{gray!15}\mathbf{2.08}$ & $\cellcolor{gray!15}\mathbf{3.26}$ & $5.21$ & $3.69$ \\
\midrule
\multirow{3}{*}{(2)}
& \textbf{FP2-Mid} & $\mathbf{6.35}$ & $\mathbf{5.79}$ & $\mathbf{2.08}$ & $\mathbf{3.26}$ & $\mathbf{5.22}$ & $\mathbf{3.70}$ \\
& w/o Text.Pref. & $6.33$ & $5.76$ & $\mathbf{2.08}$ & $3.24$ & $\mathbf{5.22}$ & $3.53$ \\
& w/o Num.Pref. & $6.07$ & $5.74$ & $1.92$ & $3.16$ & $5.14$ & $3.21$ \\
\midrule
\multirow{3}{*}{(3)}
& \textbf{FP2-Pre} & $\mathbf{6.17}$ & $\mathbf{5.76}$ & $2.01$ & $\mathbf{3.22}$ & $\mathbf{5.16}$ & $\mathbf{3.80}$ \\
& w/o Triplet & $6.11$ & $5.74$ & $1.96$ & $3.20$ & $5.15$ & $3.33$ \\
& $\circlearrowright$ InfoNCE & $5.63$ & $5.46$ & $\mathbf{2.08}$ & $3.13$ & $4.97$ & $3.57$ \\
\midrule
\multirow{3}{*}{(4)}
& \textbf{FP2-Post} & $5.83$ & $\mathbf{5.73}$ & $\mathbf{1.89}$ & $\mathbf{3.09}$ & $\mathbf{5.05}$ & $3.66$ \\
& w/o $\ell_2$-Post & $5.65$ & $5.10$ & $1.35$ & $3.02$ & $4.88$ & $2.11$ \\
& $\circlearrowright$ LoRA & $\mathbf{5.85}$ & $5.71$ & $1.88$ & $3.08$ & $\mathbf{5.05}$ & $\mathbf{3.67}$ \\
\bottomrule
\end{tabular*}
\vspace{-6pt}
\end{table}

\textbf{Numerical and textual preferences.}
Table~\ref{tab:ablation_diff_pref_loss} shows that removing numerical preferences causes a clear drop, whereas removing textual preferences changes performance only slightly except on MO-Hopper-3obj (3.70 vs.\ 3.53), indicating that numerical preferences drive trade-off control, while textual preferences mainly serve as a semantic interface.





\textbf{Effect of contrastive objective.}
Table~\ref{tab:ablation_diff_pref_loss} shows that removing contrastive learning (w/o Triplet) degrades performance.
Triplet loss outperforms InfoNCE~\cite{oord2018representation}, which treats all other in-batch samples as negatives, by using nearby-preference negatives to capture subtle behavior differences, improving preference awareness.

\textbf{Effect of post-training strategies.}
FP2-Post uses the $\ell_2$ penalty in Eq.~(9), with LoRA~\cite{hu2022lora} as an alternative.
As shown in Table~\ref{tab:ablation_finetuning}, both outperform w/o $\ell_2$-Post on all six tasks and perform comparably, suggesting that FP2 mainly benefits from post-training itself rather than from a specific fine-tuning technique.

\section{Conclusion and limitation}
\label{sec:conclusion}
We studied generalization across preferences and continuous-control tasks in offline multi-objective reinforcement learning.
Accordingly, we proposed \textbf{F}oundation \textbf{P}areto \textbf{F}low \textbf{P}olicy (\textbf{FP2}), an offline MORL foundation policy that provides a reusable multi-objective trade-off prior for continuous control.
FP2 is built on rectified flow and trained via a principled three-stage pipeline, with mid-training and a contrastive triplet loss to enhance fine-grained preference alignment.
Experiments show that FP2 generalizes to unseen preferences and supports few-shot cross-task adaptation of a shared multi-objective trade-off prior with a unified policy.
These results highlight the benefits of mid-training with high-quality Pareto-optimal data in data-scarce offline settings and the triplet regularization over nearby negative samples for fine-grained control in continuous preference space.
\textbf{Limitations.}
There are two main limitations.
First, textual preferences are constructed from templates, which ensures consistency through deterministic rules but limits flexibility for free-form language expressions.
Second, FP2 is evaluated only in state-based continuous control, and its cross-task gains are not yet disentangled into multi-objective structure versus general trajectory pretraining; extending it to richer modalities, such as image inputs, and isolating these factors remain interesting directions for future work.

\begin{ack}
This work was supported by the Natural Science Foundation of Top Talent of SZTU under Grant No. GDRC202521.
\end{ack}

\bibliographystyle{plainnat}
\bibliography{my_references}

\newpage
\appendix

\appendix
\makeatletter
\setlength{\@fptop}{0pt}
\setlength{\@fpsep}{12pt plus 2pt}
\setlength{\@fpbot}{0pt plus 1fil}
\makeatother
\renewcommand{\topfraction}{0.9}
\renewcommand{\bottomfraction}{0.6}
\renewcommand{\textfraction}{0.1}
\renewcommand{\floatpagefraction}{0.75}
\setcounter{topnumber}{3}
\setcounter{totalnumber}{5}

\section{Dataset}
\subsection{Raw data}
\begin{table}[ht]
\begin{center}
\small
\setlength{\tabcolsep}{5pt}
\caption{Basic statistics of the full \texttt{Expert} and \texttt{Amateur} datasets. Preference ranges are reported for the first two dimensions of the preference vector.}
\label{tab:raw_dataset_stats_full}
\resizebox{\linewidth}{!}{
\begin{tabular}{lccccc}
\toprule
$\mathcal{D}_{\text{full}}$  & \#Objectives & Preference range & \#Preference & \#Optimal traj. & Total traj.  \\
\midrule
MO-Ant         & 2 & $[0.2391, 0.2979]\xrightarrow{}[0.7021, 0.7609]$ & 501 & 102/65 & 50k  \\ 
MO-HalfCheetah & 2 & $[0.1668, 0.3768]\xrightarrow{}[0.6232, 0.8332]$ & 501  &1547/459 &50k \\
MO-Hopper      & 2 & $[0.3208, 0.4353] \xrightarrow{} [0.5647, 0.6792]$ & 501  & 223/52 &50k \\
MO-Walker2d    & 2 & $[0.1668, 0.2583]\xrightarrow{}[0.7417, 0.8332]$ & 501 &2055/411 &50k  \\
MO-Swimmer     & 2 & $[0.0774, 0.0950] \xrightarrow{} [0.9050, 0.9226]$ & 501 &1475/58 &50k   \\
MO-Hopper-3obj  & 3 & $[0.1511, 0.0007] \xrightarrow{} [0.6078, 0.5919]$ & 501 & 8945/1925 &50k   \\
\bottomrule
\end{tabular}
}
\end{center}
\vskip -0.1in
\end{table}

We use the \textbf{D4MORL benchmark}~\cite{zhu2023scaling}, which consists of six multi-objective
robotic control environments with two or three objectives.
Each environment provides trajectories generated under a dense set of preference vectors.

Table~\ref{tab:raw_dataset_stats_full} summarizes the basic statistics of the full dataset,
which we denote as $\mathcal{D}_{\text{full}}$.
Following common practice, trajectories are segmented into fixed-length windows during training;
therefore, the number of trajectories does not directly correspond to the number of training samples.

\subsection{Dataset construction}

All datasets used in this work are derived from the full dataset $\mathcal{D}_{\text{full}}$,
consistent with the description in the main paper.

\paragraph{Pre-train split.}
Following the protocol of~\cite{yuan2024moduli}, we randomly retain $70\%$ of trajectories
from $\mathcal{D}_{\text{full}}$ to construct the pre-training dataset $\mathcal{D}_{\text{pre}}$,
with the remaining $30\%$ forming the dropped subset $\mathcal{D}_{\text{drop}}$.
The dropped data are not used during pre-training.

\paragraph{Mid-train dataset.}
From $\mathcal{D}_{\text{pre}}$, we further construct a dataset
$\mathcal{D}_{\text{mid}}$ by retaining only the trajectories on top two Pareto fronts ranked by dominance.
Specifically, we first compute the Pareto-optimal set for each environment, remove it, and then recompute the next Pareto front to obtain a set of near-optimal trajectories.
The union of these two fronts constitutes $\mathcal{D}_{\text{mid}}$, which is used for
preference alignment in Stage~2.
Table~\ref{tab:raw_dataset_stats_mid} reports the number of Pareto-optimal and suboptimal trajectories
retained in $\mathcal{D}_{\text{mid}}$ for each environment.

\begin{table}[ht]
\begin{center}
\small
\setlength{\tabcolsep}{5pt}
\caption{Basic statistics of $\mathcal{D}_{\text{mid}}$, filtered from the \texttt{Expert} and \texttt{Amateur} datasets (abbreviated as ``E'' and ``A'', respectively, in the table).}
\label{tab:raw_dataset_stats_mid}
\begin{tabular}{lccccc}
\toprule
$\mathcal{D}_{\text{mid}}$  & \#Optimal traj. (E/A) & \#Suboptimal traj. (E/A) & Total traj. (E/A) \\
\midrule
MO-Ant & 98/58 & 166/92 & 264/150\\ 
MO-HalfCheetah & 970/393 & 1245/531 & 2215/924\\
MO-Hopper &182/47 & 250/71 & 432/118   \\
MO-Walker2d & 1688/360 & 2077/590 & 3765/950  \\
MO-Swimmer  &1032/55 &1230/92 &2262/147    \\
MO-Hopper-3obj &6573/1583 &5179/1911 &11752/3494  \\
\bottomrule
\end{tabular}
\end{center}
\vskip -0.1in
\end{table}

\paragraph{Task-level split.}
For foundation models, we use the data of five environments for training and leave the remaining one environment for out-of-distribution evaluation, denoted as $\mathcal{D}_{\text{task}}$.

\subsection{Preprocessing}

Before slicing trajectories into fixed-horizon windows, we apply the following preprocessing steps to the raw trajectories.

\begin{itemize}
    \item \textbf{State normalization.}
    States are normalized using dataset-wide Gaussian statistics.
    Let $\mu_s$ and $\sigma_s$ denote the mean and standard deviation of states.
    The normalized state at time step $t$ is given by
    $\tilde{\bm{s}}_t = (\bm{s}_t - \mu_s) / \sigma_s$.

    \item \textbf{Action normalization.}
    Actions are scaled to the range $[-1, 1]$ using min--max normalization based on the dataset
    action bounds.

    \item \textbf{Objective-aware reward normalization.}
    To mitigate scale discrepancies across objectives and preference configurations, we adopt
    a neighborhood-based normalization strategy.
    For a trajectory $\tau^i$ with preference vector $\bm{\omega}^i$, we define its neighborhood as
    $\mathcal{N}(i) = \{ j \mid \|\bm{\omega}^j - \bm{\omega}^i\| < \epsilon \}$ with $\epsilon$ be a constant.
    The normalized reward is computed as
    \[
    \tilde{\bm{r}}_t^i =
    \frac{
        \bm{r}_t^i - \min_{j \in \mathcal{N}(i)} \bm{r}^j
    }{
        \max_{k \in \mathcal{N}(i)} \bm{r}^k -
        \min_{j \in \mathcal{N}(i)} \bm{r}^j + \delta
    },
    \]
    where $\delta$ is a small constant for numerical stability.
\end{itemize}

After normalization, trajectories are sliced into overlapping windows of horizon $H$,
yielding training samples of the form
$\{\bm{s}_t, \bm{a}_t, \bm{r}_t, \bm{\omega}, \bm{e}_{\text{txt}}\}_{t=t_i}^{t_i+H}$.

\section{Implementation details}
\label{app:details}
This section provides implementation details of \textbf{FP2} that complement the method
description in Sec.~\ref{sec:archi} and Sec.~\ref{Sec:training_objective}.
All components described below instantiate the architecture and training pipeline
introduced in the main paper and do not introduce additional modeling assumptions.
We organize the details into a \textbf{Common Setup} (Sec.~\ref{sec:appendix_common_setup}), shared across all training stages,
and a \textbf{Mixed-Environment Training Setup} (Sec.~\ref{sec:appedix_mixed_training_setup}), which is enabled only when jointly training on multiple
environments.

\subsection{Common setup}
\label{sec:appendix_common_setup}

\paragraph{Backbone and embeddings.}
All diffusion models in FP2 use a 1D rectified-flow backbone based on a
multi-modal diffusion transformer (MMDiT-1D), consistent with the architecture shown in
Fig.~\ref{fig:network}.
The network has a model dimension $d_{\text{model}} = 384$, $6$ attention heads,
and $4$ MMDiT blocks.
Input trajectory tokens are projected to the model dimension via a linear embedding layer.
Each transformer block employs adaptive layer normalization (adaLN-Zero),
where the modulation parameters are conditioned on a joint time--condition embedding.
The conditioning signal $\bm{c}$ (including preference, return, and text guidance)
is encoded by a two-layer MLP (128 $\rightarrow$ 384) and summed with the sinusoidal
time embedding to form the final time--condition embedding.
\paragraph{Text prompt generation.}
D4MORL provides numerical preference vectors but no natural language annotations.
For each trajectory, the preference vector $\omega$ is already stored in the dataset and indicates the objective trade-off used to generate that trajectory.
Therefore, our text prompts are not inferred from data or manually annotated; they are automatically generated from the existing $\omega$. Specifically, we convert each $\omega$ into a short text description using fixed rules. We first compare the relative weights in $\omega$ to determine which objectives are more important.
We then assign the preference to several priority levels, such as balanced, moderate and strong. Preferences with similar weight ratios are mapped to the same level and share the same text template. Finally, we fill the template with environment-specific objective names. For example, in a two-objective task, $\omega=[0.5,0.5]$ is mapped to \texttt{``balanced trade-off between speed and energy efficiency''}, while $\omega=[0.8,0.2]$ is mapped to \texttt{``strongly prioritize speed over energy efficiency''}. Thus, the text prompt is a deterministic description of the same preference vector, rather than an additional source of supervision.

\paragraph{Text-based preference guidance.}
As described in Sec.~\ref{sec:archi}, textual preferences are encoded using dual text encoders.
Specifically, we concatenate a 384-dimensional SentenceTransformer embedding
(\texttt{all-MiniLM-L6-v2}) and a 512-dimensional CLIP embedding (ViT-B/32),
resulting in an 896-dimensional text representation $\bm{e}_{\text{txt}}$.
Textual preferences are incorporated through classifier guidance (CG).
We train a lightweight text-preference classifier $C_\phi$, whose network $f_\phi$, a small MMDiT-style 1D transformer
(128-dimensional diffusion embedding, model width 192, 2 transformer blocks, and 6 attention heads),
predicts the preference text embedding from a noisy trajectory $\bm{z}_t$.
Following Eq.~\eqref{eq:cg}, the classifier models $\log p_\phi(\bm{e}_{\text{txt}} \mid \bm{z}_t, t)$ with a Gaussian likelihood and is trained by minimizing the negative log-likelihood, which reduces to a mean-squared-error objective:
\[
\mathcal{L}_{\text{cls}} = -\mathbb{E}_{\bm{x},t}\left[C_\phi(\bm{z}_t,t,\bm{e}_{\text{txt}})\right],\quad
C_\phi(\bm{z}_t,t,\bm{e}_{\text{txt}}) = \log p_\phi(\bm{e}_{\text{txt}}\mid\bm{z}_t,t) \propto -\left\|f_\phi(\bm{z}_t,t)-\bm{e}_{\text{txt}}\right\|_2^2,
\]
where $\bm{e}_{\text{txt}}$ denotes the offline text embedding of the preference.
During inference, an exponential moving average (EMA) copy of the classifier provides $\nabla_{\bm{z}_t} C_\phi$ to guide the vector field (Eq.~\eqref{eq:cg}) and scores sampled trajectories by preference alignment; the scores are aggregated along the sampled horizon to select the final action.

\paragraph{Step-dependent loss weighting.}
Errors at the beginning of a generated trajectory segment affect all subsequent steps and thus have a disproportionate influence on long-horizon returns.
We therefore weight the rectified-flow loss in Eq.~\eqref{eq:loss_rf} over trajectory steps and feature dimensions:
\[
\mathcal{L}_{\text{RF}}
=
\mathbb{E}_{\bm{z}_0,\, \bm{x},\, t}
\left[
\left\|
\bigl(
v_{\theta}(\bm{z}_t, t \mid \bm{c}) - (\bm{x} - \bm{z}_0)
\bigr)
\odot \bm{W}
\right\|_2^2
\right],
\]
where $\odot$ denotes the element-wise product and $\bm{W} \in \mathbb{R}^{H_{\text{max}} \times D}$ assigns a weight to each step $h$ and feature $d$ of a (padded) segment with horizon $H_{\text{max}}$ and feature dimension $D$.
We set $W_{h=1,d} = 10$ for the state dimensions of the first step and $W_{h>1,d} = 1$ for all other steps; padded entries are excluded by the feature mask (Appendix~\ref{sec:appedix_mixed_training_setup}).
This weighting emphasizes the first step of each segment and empirically stabilizes optimization across heterogeneous tasks.

\subsection{Mixed-environment training setup}
\label{sec:appedix_mixed_training_setup}

The following mechanisms are enabled only in the joint multi-environment training setting, where state and action spaces are heterogeneous.

\paragraph{Feature-space masking.}
To handle varying state and action dimensions across tasks, we construct a binary
feature mask $\mathbf{M}_{\text{feat}} \in \{0,1\}^{H_{\text{max}} \times (D_s^{\max}+D_a^{\max})}$.
For a given environment, entries corresponding to valid features are set to 1, while 
those corresponding to padded dimensions are set to 0.
The mask is applied element-wise to the rectified-flow loss to ensure gradients are
propagated only through task-relevant features.

\paragraph{Temporal masking.}
We additionally apply temporal masking to handle variable trajectory horizons.
A boolean attention mask $\mathbf{M}_{\text{attn}} \in \{0,1\}^{H_{\text{max}}}$ is injected
into the self-attention layers of the MMDiT backbone to prevent attention to padded
timesteps.
In parallel, the feature mask $\mathbf{M}_{\text{feat}}$ is modulated by temporal validity,
such that all features at invalid timesteps are excluded from the training objective.
Together, these mechanisms confine both attention and supervision to valid temporal contexts.

\section{Training-stage analysis}
\begin{table}[!htbp]
\centering
\caption{Training hyperparameters.}
\label{tab:training_parameter}
\begin{tabular}{lccc}
\toprule
\multirow{2}{*}{\textbf{Parameter}}
 & \multicolumn{3}{c}{\textbf{Value}} \\
\cmidrule(lr){2-4}
& \textbf{Pre-train} & \textbf{Mid-train} & \textbf{Post-train} \\
\midrule
Learning Rate & 2e-4 & 2e-4 & 1e-5 \\
Weight Decay & 1e-5 & 1e-5 & 1e-5 \\
Batch Size/Mini Batch & 2048/256 & 64/64 & 64/64 \\
Training Steps & 400K & 200K & 100K\\
\bottomrule
\end{tabular}
\end{table}

The hyperparameters employed in the three-stage training are summarized in Tab.~\ref{tab:training_parameter}. We adopted a fixed learning rate and weight decay rate for each stage.

\subsection{Single-environment training}
A relatively large batch size is employed in the pre-training stage to accelerate convergence, where we retain the weights that maximize the Hypervolume (HV) metric to prioritize diversity coverage. During mid-training, a small batch size is adopted to enhance alignment quality—accompanied by a corresponding reduction in the HV metric for certain environments—and we retain the weights that jointly optimize both HV and Preference Alignment Score (PAS) metrics. Finally, in the post-training stage, we use the same batch size as mid-training but a smaller learning rate to enable more stable gradient updates, while retaining the weights that yield optimal HV and PAS performance.

\subsection{Mixed-environment training}
\begin{figure}[htbp!]
\centering
\includegraphics[width=\linewidth]{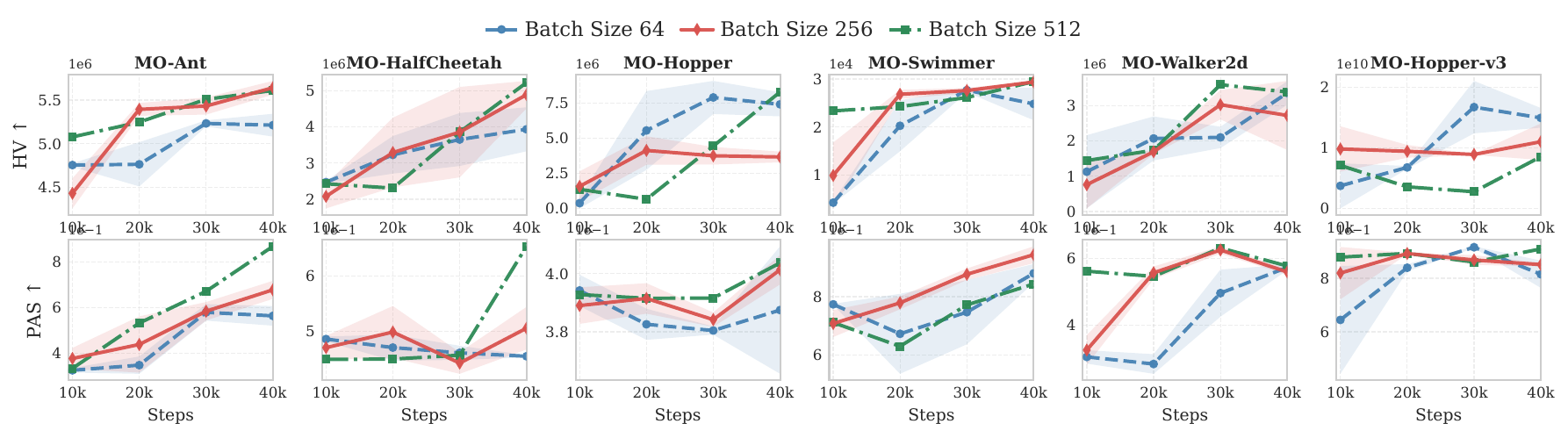}
\caption{Ablation on the batch size of mixed-environment training.}
\label{fig:bs_ablation}
\vspace{-6pt}
\end{figure}

 Mini-batches are utilized, where each batch contains an equal amount of data from different environments for gradient updates. As shown in Fig.~\ref{fig:bs_ablation}, we illustrate the metric evolution across all environments during the mixed training process, where the low PAS in the Hopper environment is primarily attributed to conflicts with the Hopper-3obj environment.

\subsection{Effect of mid-training on preference--policy alignment}

\begin{figure}[htbp!]
    \centering
    \includegraphics[width=0.9\linewidth]{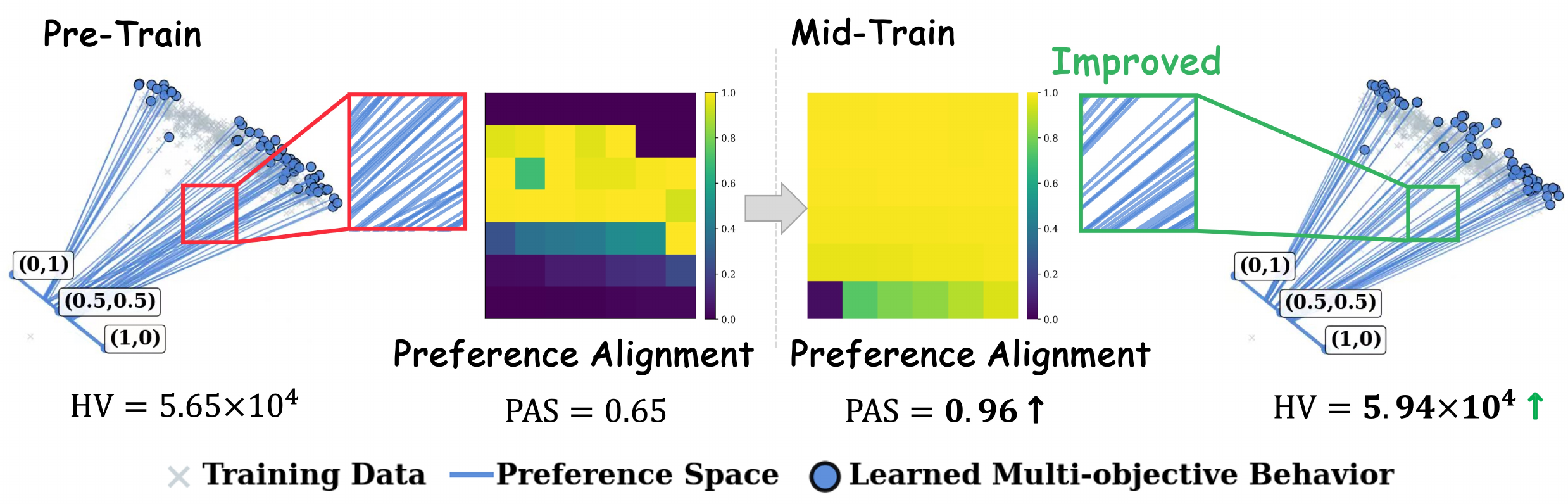}
    \caption{Effect of mid-training on preference--policy alignment. Mid-trained FP2 better aligns policy and preference than pre-trained FP2, achieving higher PAS and more uniform preference alignment distribution.}
    \label{fig:pref_align_analysis}
\end{figure}

Preference alignment is critical for preference generalization.
As a case study, Fig.~\ref{fig:pref_align_analysis} compares alignment on MO-Ant \texttt{Amateur}-\texttt{Narrow}.
By leveraging high-quality Pareto-optimal trajectories during mid-training, FP2-Mid improves preference alignment over FP2-Pre.

\section{Additional results}
\label{app:add_results}

\subsection{Controlled comparison and resource fairness}
\label{app:resource_fairness}

To ensure a fair comparison with prior generative MORL methods, we summarize the model capacity, training budget, and inference efficiency of our proposed FP2 and the state-of-the-art method MODULI~\cite{yuan2024moduli} in \cref{tab:resource_fairness}.
This resource-level comparison complements the controlled experiments in the
main text, where we isolate the effects of the generative backbone and the
multi-stage training pipeline.

As shown in \cref{tab:resource_fairness}, FP2 does not rely on a larger model or a more expensive inference procedure than MODULI. 
Both methods use the same DiT-1D architecture, while FP2 has fewer trainable parameters and lower inference latency. 
These results suggest that the empirical gains of FP2 are not simply due to increased model capacity or computational advantage.

\begin{table}[!htbp]
\centering
\caption{
Resource-level comparison between FP2 and state-of-the-art method MODULI~\cite{yuan2024moduli}. Both methods use a DiT-1D architecture, while FP2 uses fewer trainable parameters and achieves lower inference latency under a comparable training budget.
}
\label{tab:resource_fairness}
\resizebox{\linewidth}{!}{
\begin{tabular}{lcc}
\toprule
\textbf{Aspect} & \textbf{FP2} & \textbf{MODULI} \\
\midrule
Architecture & DiT-1D & DiT-1D \\
Generative backbone & RF & Diffusion \\
Sampler & 10-step Euler & 10-step DDIM \\
Additional modules & Classifier + text encoders & ControlNet $\times 2$ + inverse-dynamics MLP \\
Trainable parameters & 12.8M & 23.4M \\
Training steps & 600K & 400K \\
Inference latency & 112 ms / action & 126 ms / action \\
\bottomrule
\end{tabular}
}
\end{table}

\subsection{Task-specific preference generalization comparison with state-of-the-art methods}
\label{app:full_resutls}

In total, we conduct 36 comparison experiments with state-of-the-art methods across six Expert datasets and six Amateur datasets under three evaluation metrics.
The statistical summary is reported in Table~\ref{tab:full_compare_36_statistics}, and the complete comparison results are provided in Tables~\ref{tab:full_compare_expert_amatuer}, \ref{tab:full_compare_expert_shat_nar}, and \ref{tab:full_compare_amateur_shat_nar}.

\begin{figure}[p]
\centering
\setlength{\tabcolsep}{2pt}
\begin{subfigure}{0.49\linewidth}
    \centering
    \includegraphics[width=\linewidth]{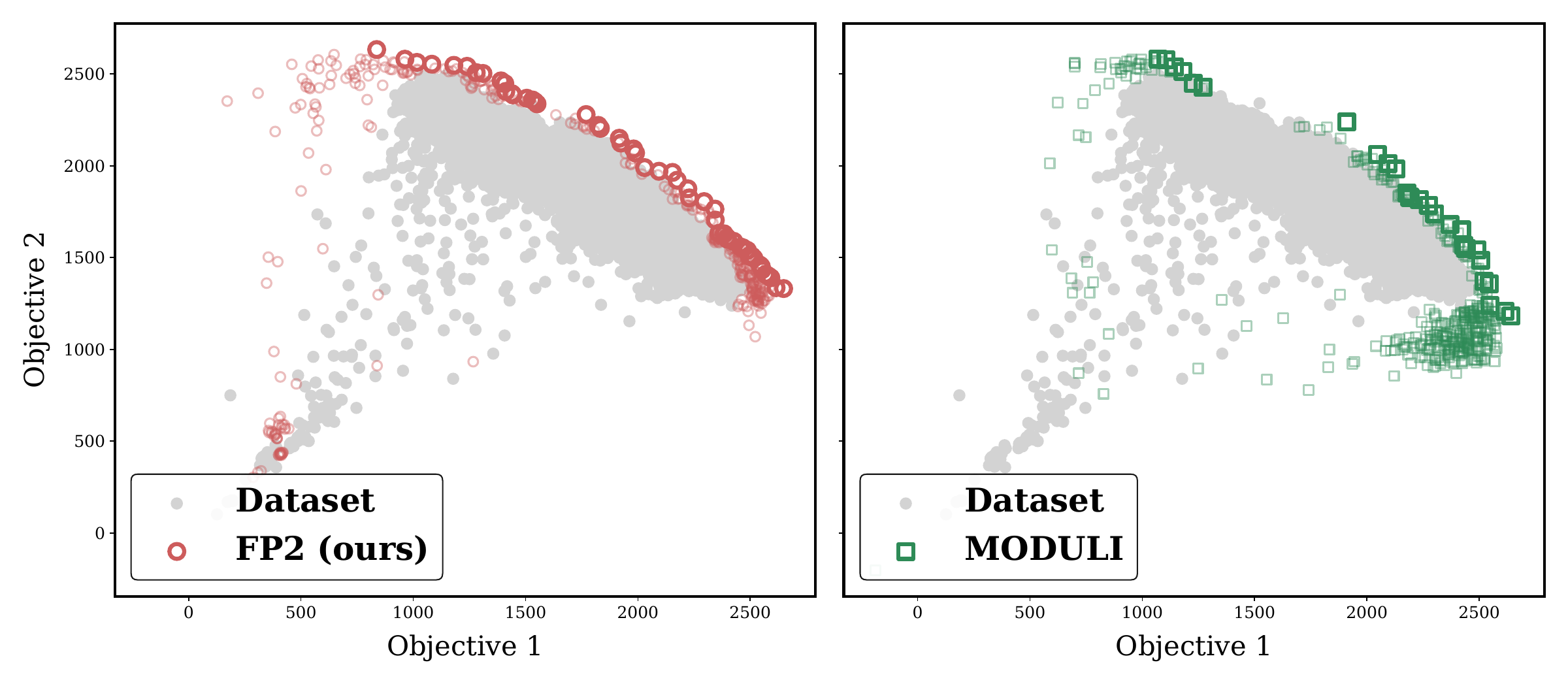}
    \caption{MO-Ant (\texttt{Amateur}-\texttt{Narrow})}
\end{subfigure}
\hfill
\begin{subfigure}{0.49\linewidth}
    \centering
    \includegraphics[width=\linewidth]{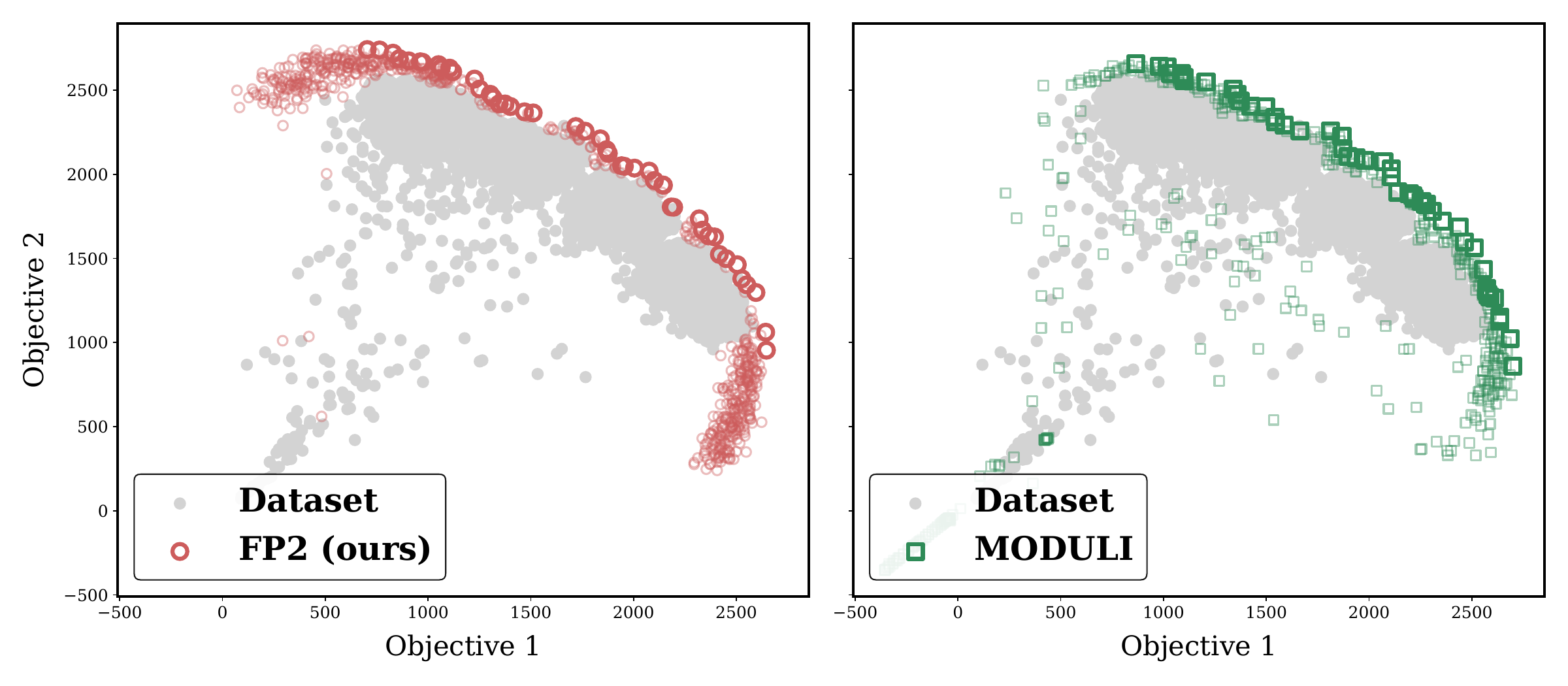}
    \caption{MO-Ant (\texttt{Amateur}-\texttt{Shattered})}
\end{subfigure}
\begin{subfigure}{0.49\linewidth}
    \centering
    \includegraphics[width=\linewidth]{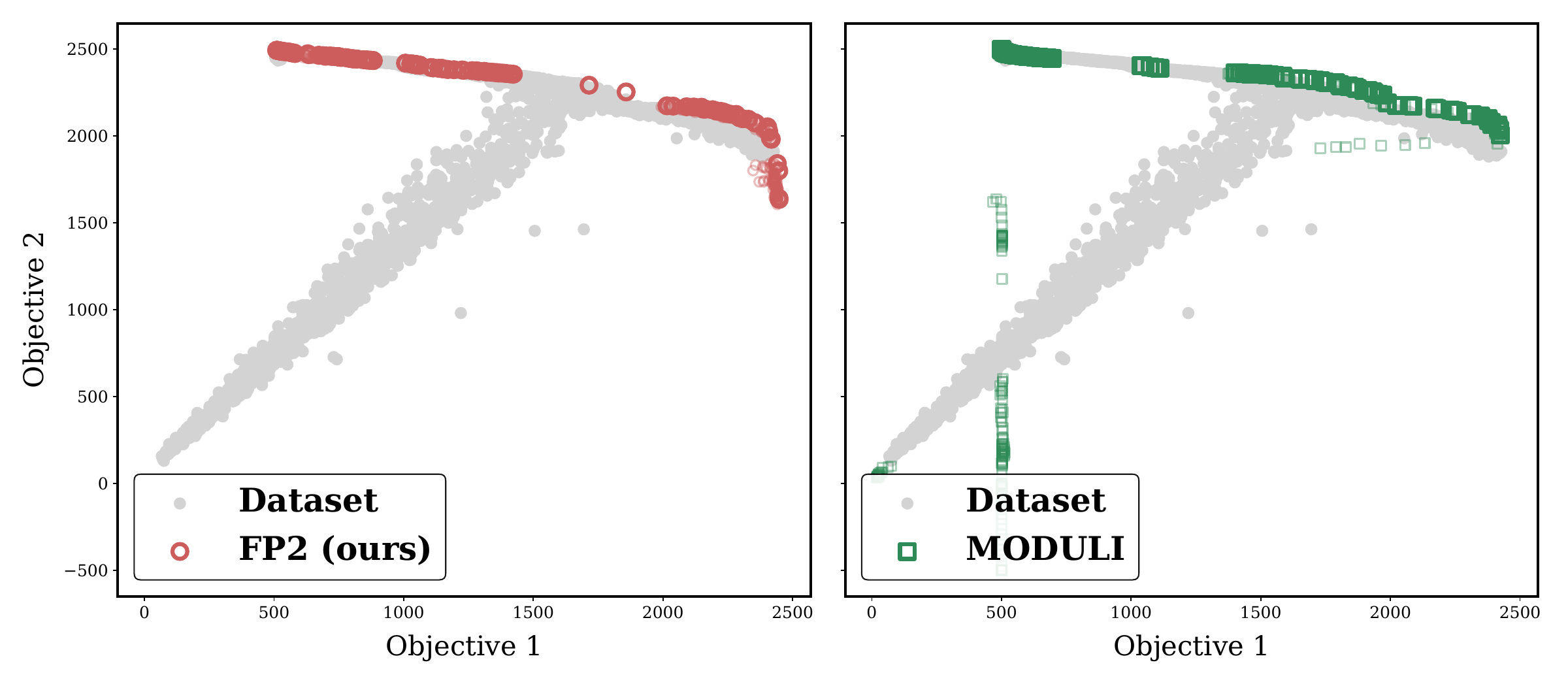}
    \caption{MO-HalfCheetah (\texttt{Amateur}-\texttt{Narrow})}
\end{subfigure}
\hfill
\begin{subfigure}{0.49\linewidth}
    \centering
    \includegraphics[width=\linewidth]{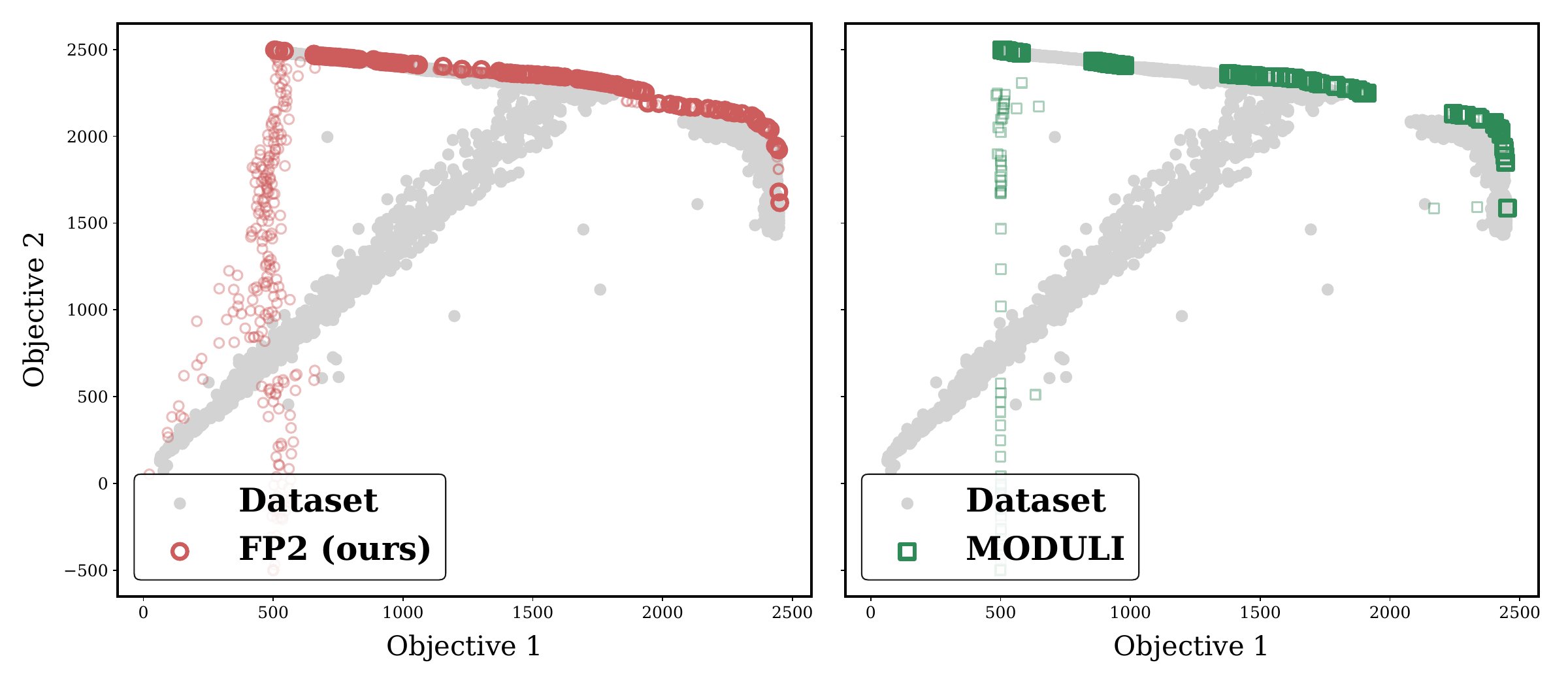}
    \caption{MO-HalfCheetah (\texttt{Amateur}-\texttt{Shattered})}
\end{subfigure}

\begin{subfigure}{0.49\linewidth}
    \centering
    \includegraphics[width=\linewidth]{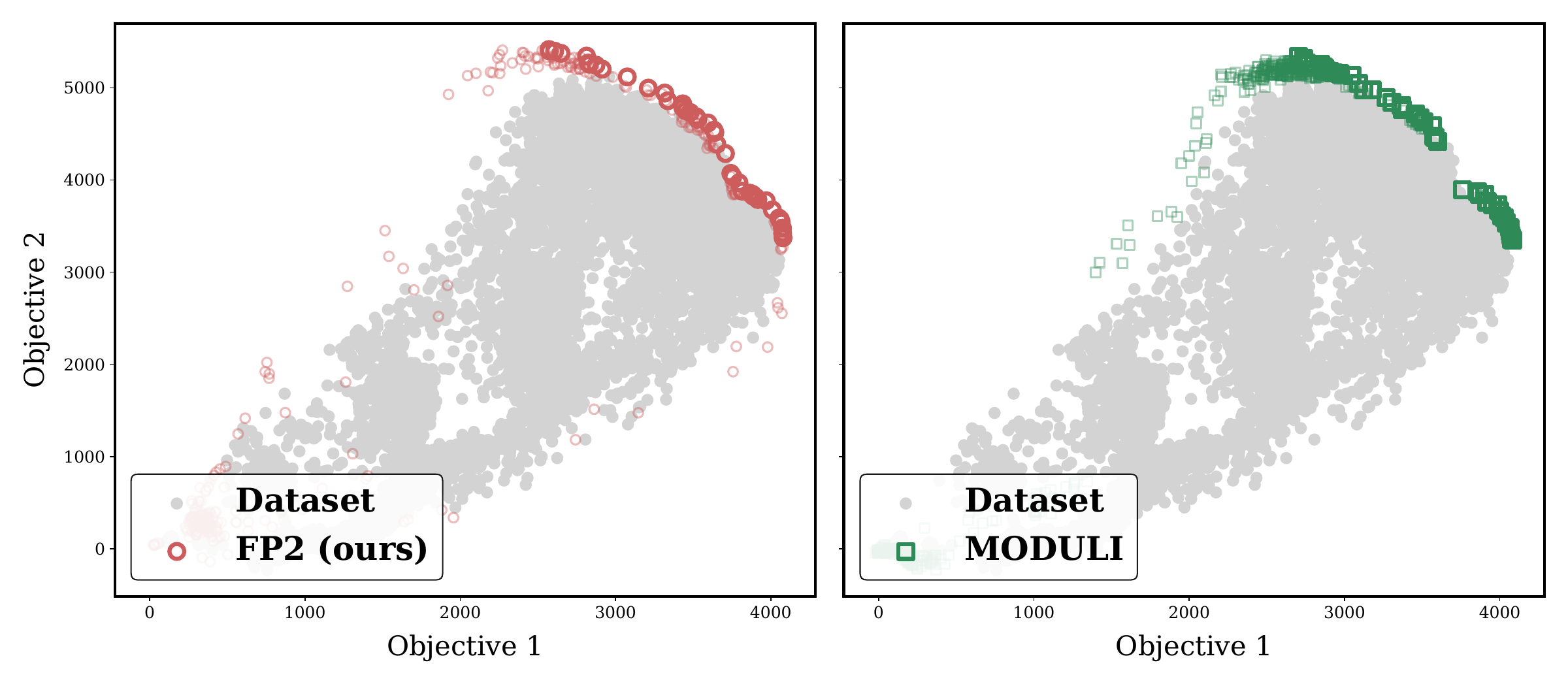}
    \caption{MO-Hopper (\texttt{Amateur}-\texttt{Narrow})}
\end{subfigure}
\hfill
\begin{subfigure}{0.49\linewidth}
    \centering
    \includegraphics[width=\linewidth]{MO-Hopper_amateur_shattered_ood_comparison.pdf}
    \caption{MO-Hopper (\texttt{Amateur}-\texttt{Shattered})}
\end{subfigure}

\begin{subfigure}{0.49\linewidth}
    \centering
    \includegraphics[width=\linewidth]{MO-Swimmer_amateur_narrow_ood_comparison.pdf}
    \caption{MO-Swimmer (\texttt{Amateur}-\texttt{Narrow})}
\end{subfigure}
\hfill
\begin{subfigure}{0.49\linewidth}
    \centering
    \includegraphics[width=\linewidth]{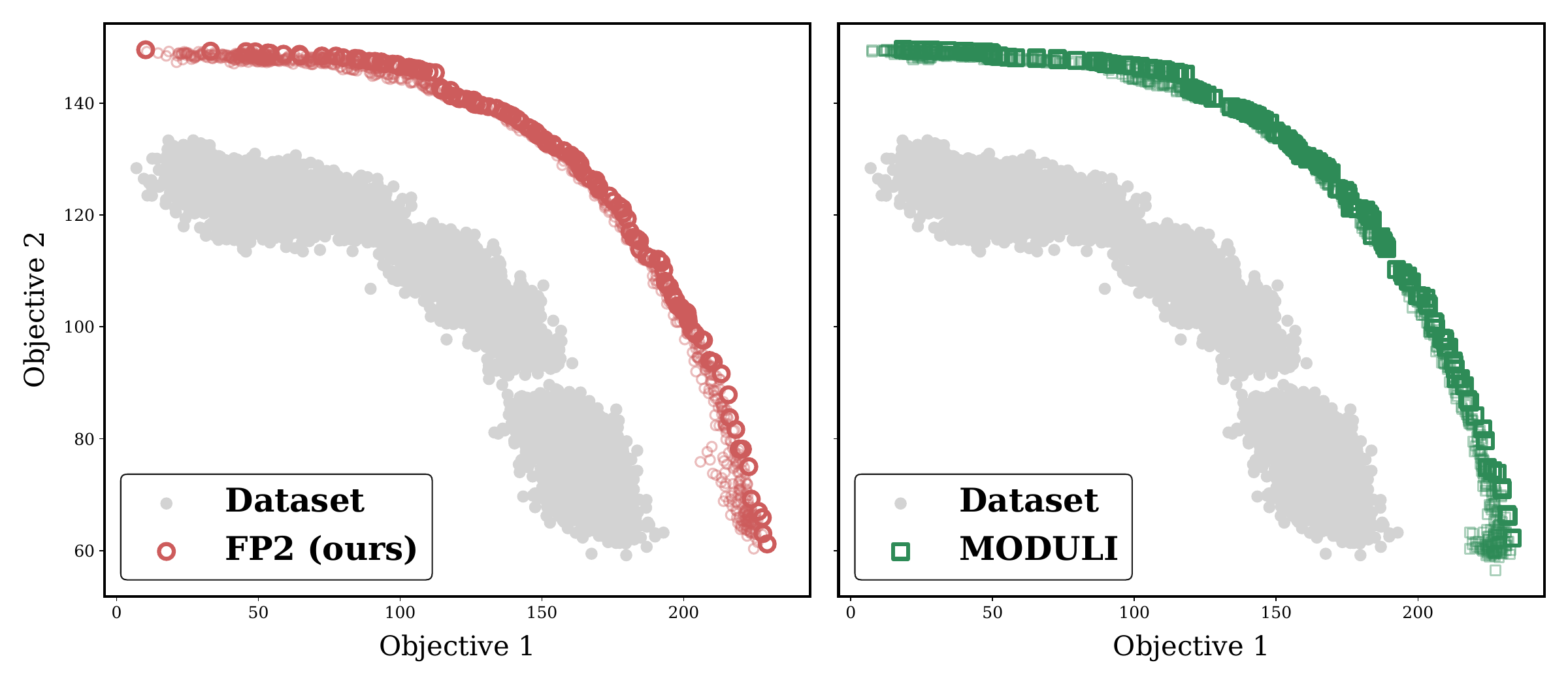}
    \caption{MO-Swimmer (\texttt{Amateur}-\texttt{Shattered})}
\end{subfigure}

\begin{subfigure}{0.49\linewidth}
    \centering
    \includegraphics[width=\linewidth]{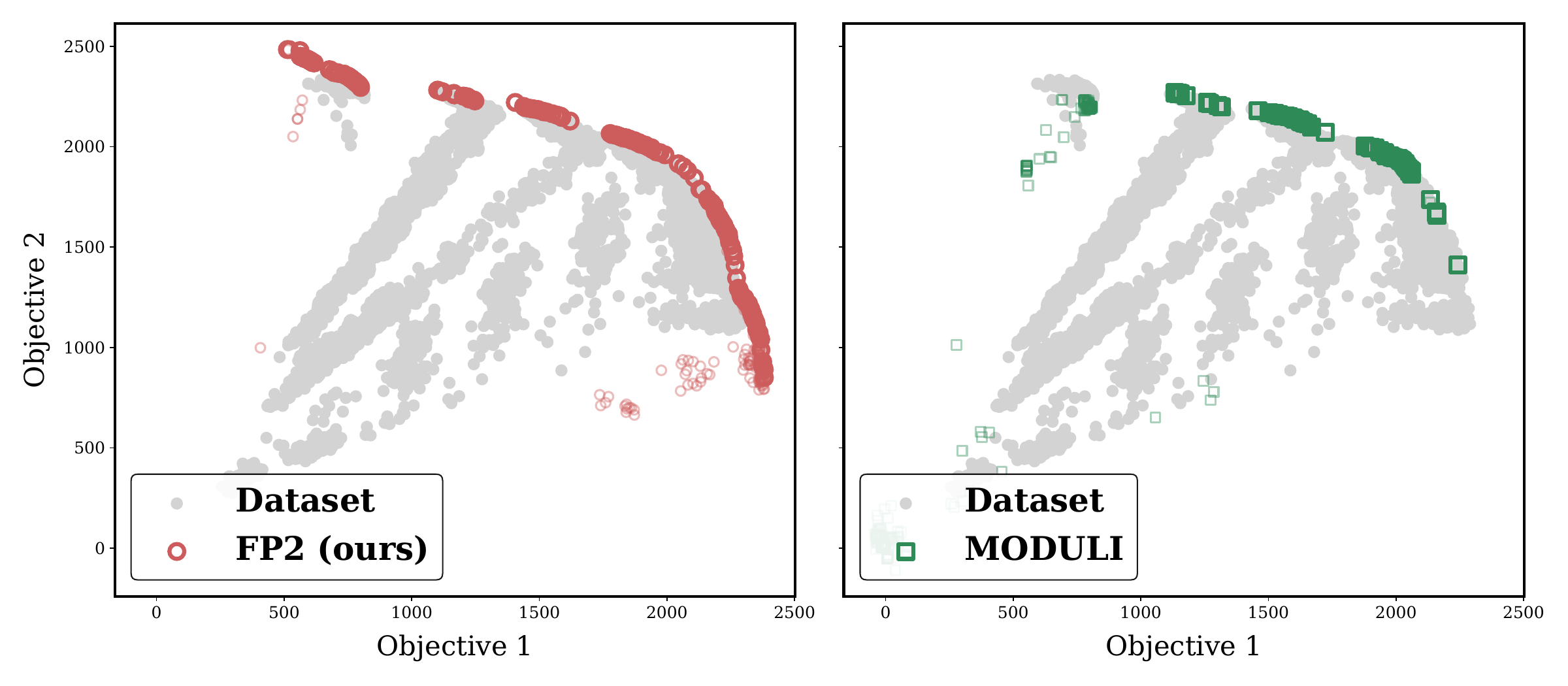}
    \caption{MO-Walker2d (\texttt{Amateur}-\texttt{Narrow})}
\end{subfigure}
\hfill
\begin{subfigure}{0.49\linewidth}
    \centering
    \includegraphics[width=\linewidth]{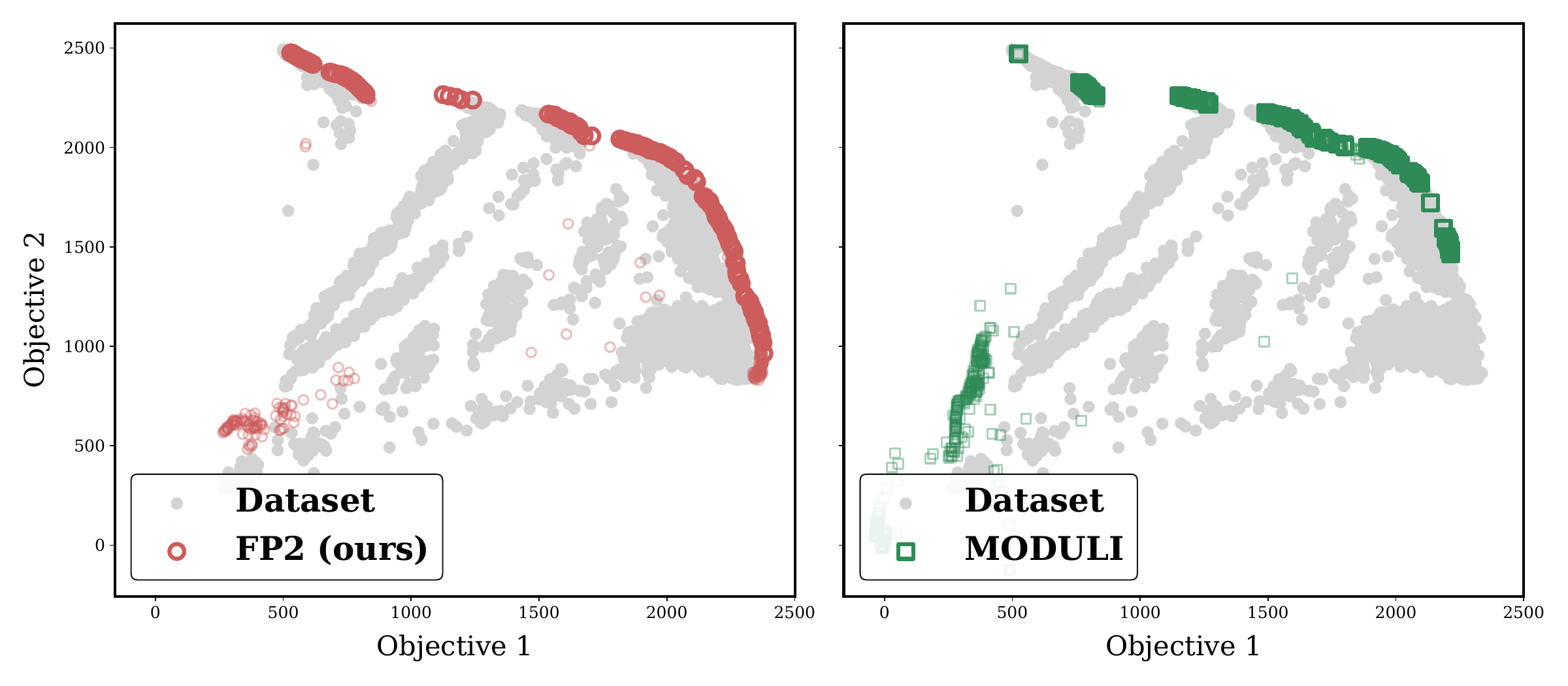}
    \caption{MO-Walker2d (\texttt{Amateur}-\texttt{Shattered})}
\end{subfigure}
\begin{subfigure}{0.49\linewidth}
    \centering
    \includegraphics[width=\linewidth]{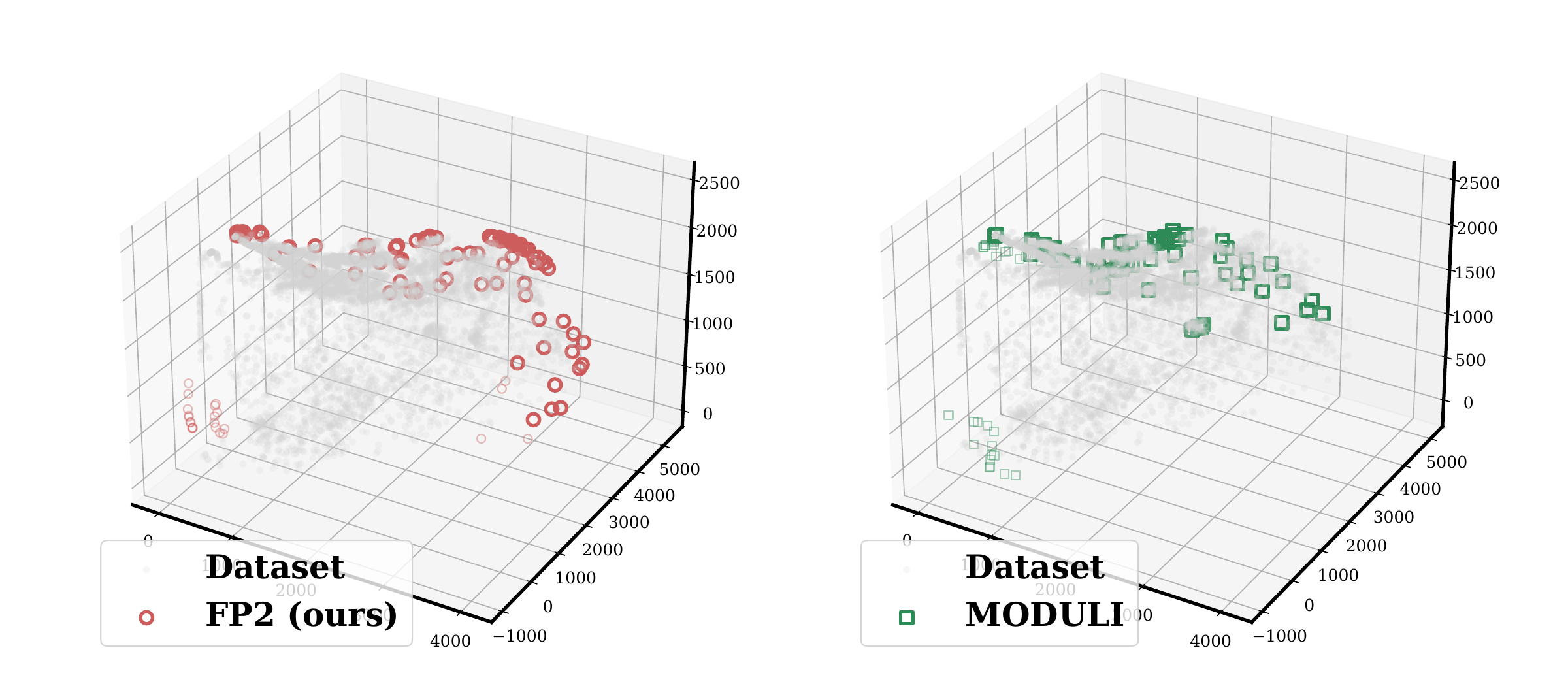}
    \caption{MO-Hopper-3obj (\texttt{Amateur}-\texttt{Narrow})}
\end{subfigure}
\hfill
\begin{subfigure}{0.49\linewidth}
    \centering
    \includegraphics[width=\linewidth]{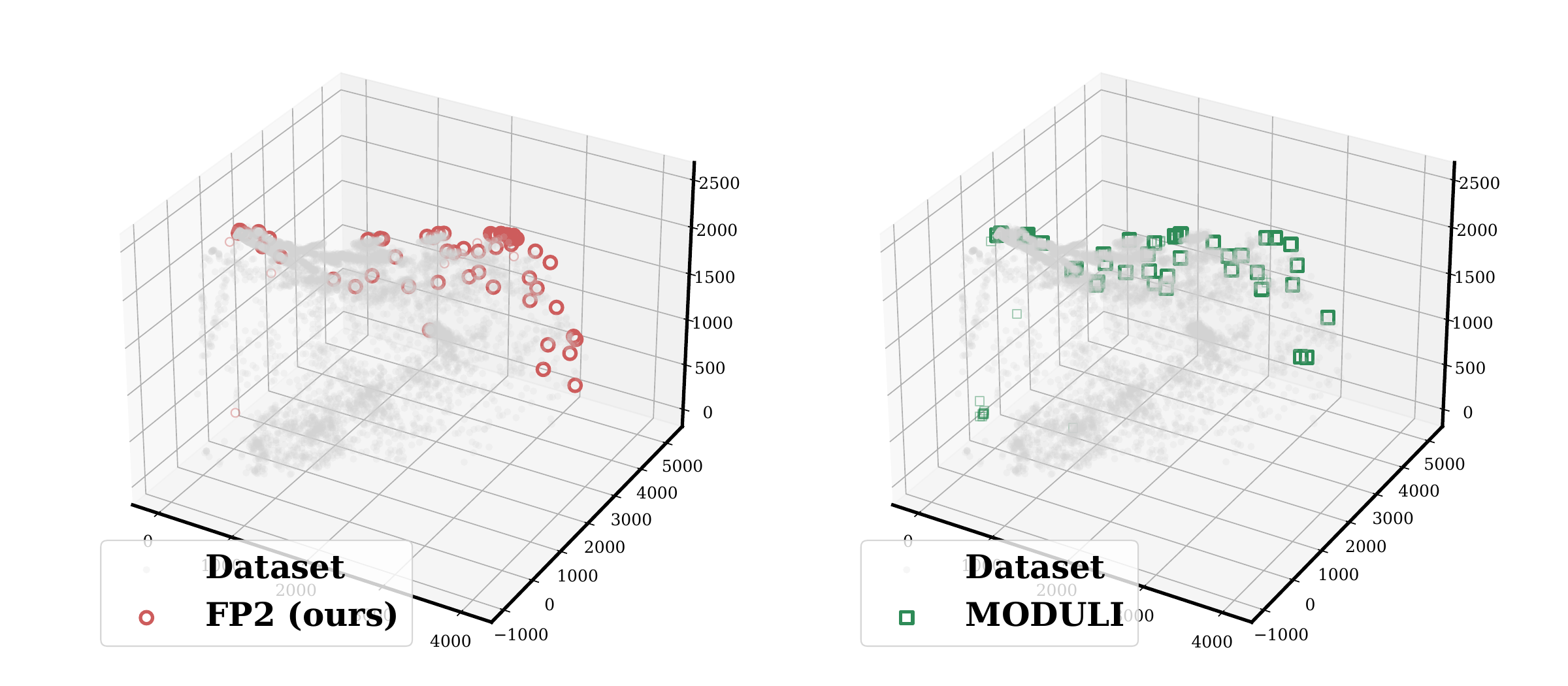}
    \caption{MO-Hopper-3obj (\texttt{Amateur}-\texttt{Shattered})}
\end{subfigure}
\caption{Preference generalization on the \texttt{Amateur} datasets. In each panel, FP2 (left) and MODULI (right) are shown with OOD points in color and dataset points in gray.}
\label{fig:ood_apx_amateur}
\end{figure}

\begin{figure}[p]
\centering
\setlength{\tabcolsep}{2pt}
\begin{subfigure}{0.49\linewidth}
    \centering
    \includegraphics[width=\linewidth]{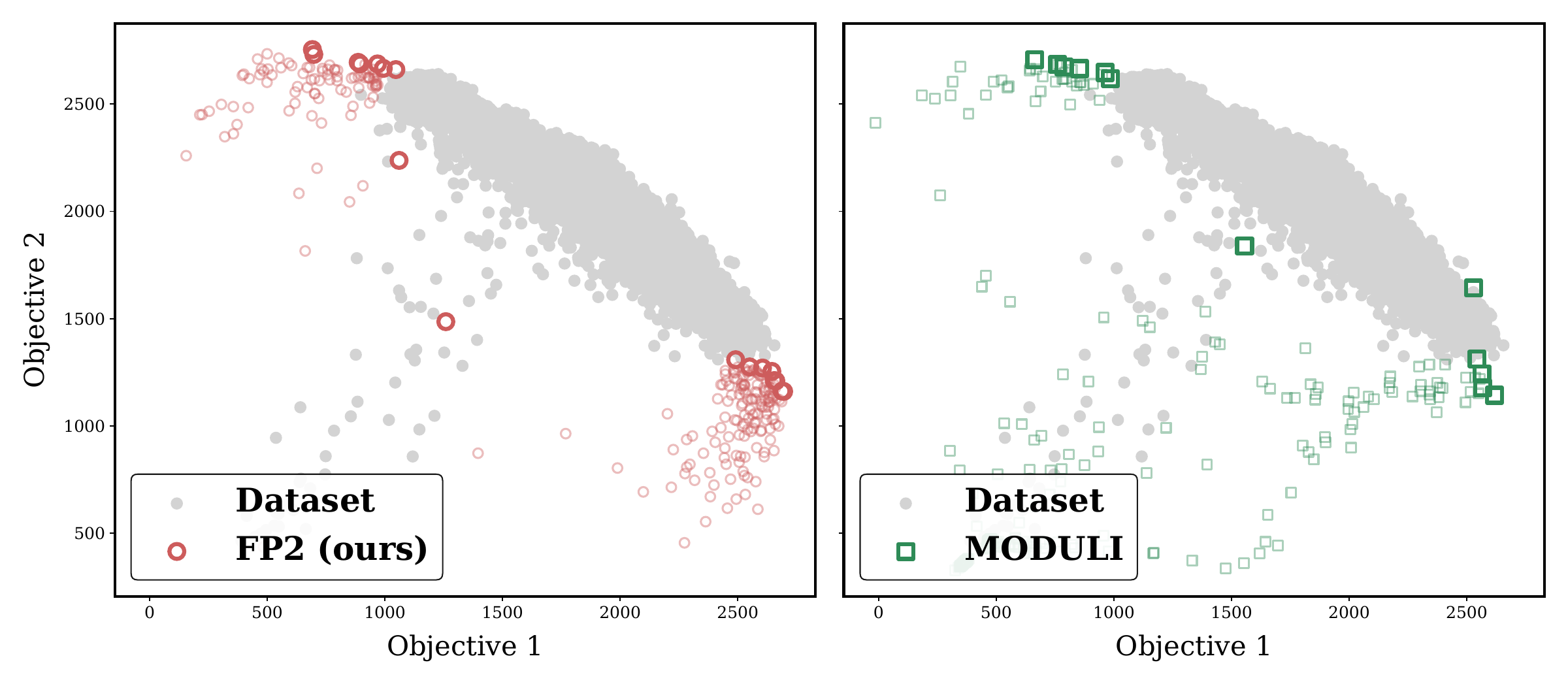}
    \caption{MO-Ant (\texttt{Expert}-\texttt{Narrow})}
\end{subfigure}
\hfill
\begin{subfigure}{0.49\linewidth}
    \centering
    \includegraphics[width=\linewidth]{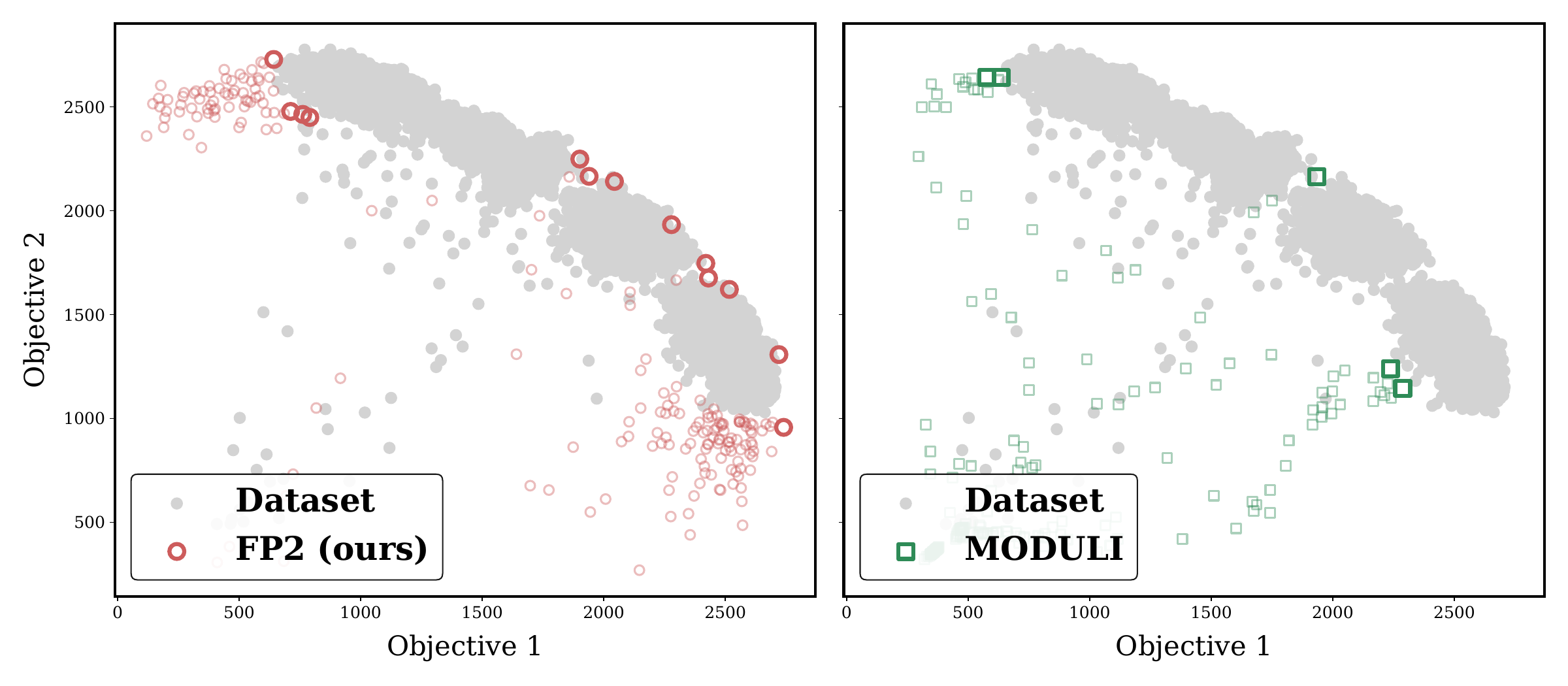}
    \caption{MO-Ant (\texttt{Expert}-\texttt{Shattered})}
\end{subfigure}

\begin{subfigure}{0.49\linewidth}
    \centering
    \includegraphics[width=\linewidth]{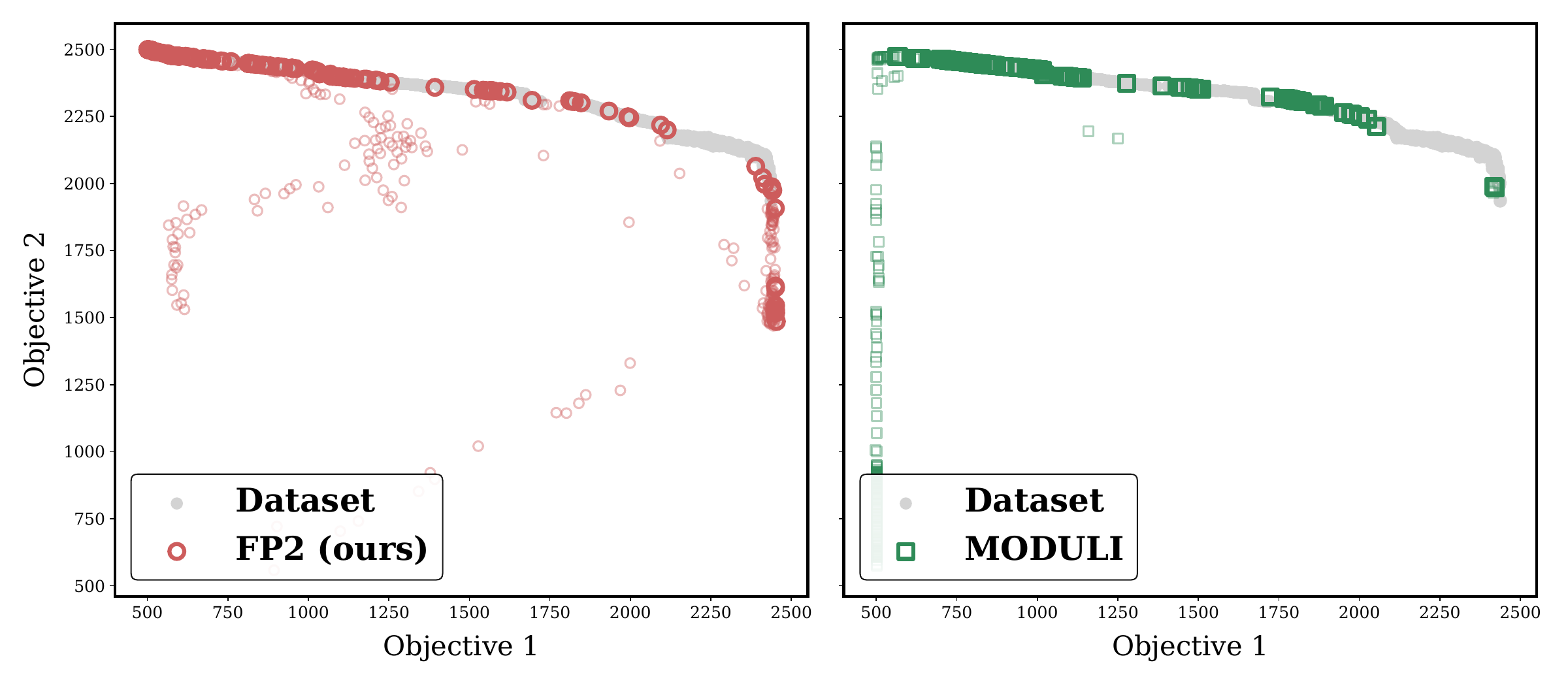}
    \caption{MO-HalfCheetah (\texttt{Expert}-\texttt{Narrow})}
\end{subfigure}
\hfill
\begin{subfigure}{0.49\linewidth}
    \centering
    \includegraphics[width=\linewidth]{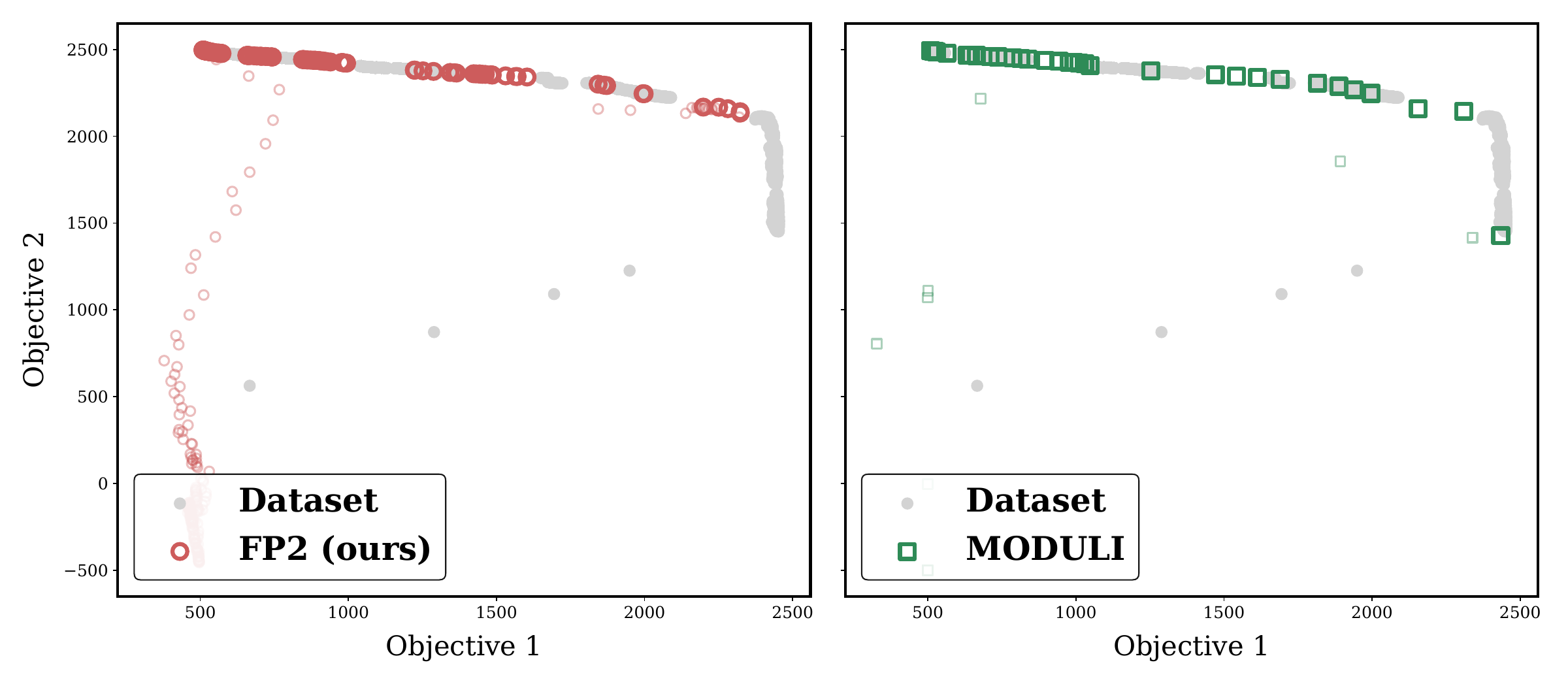}
    \caption{MO-HalfCheetah (\texttt{Expert}-\texttt{Shattered})}
\end{subfigure}

\begin{subfigure}{0.49\linewidth}
    \centering
    \includegraphics[width=\linewidth]{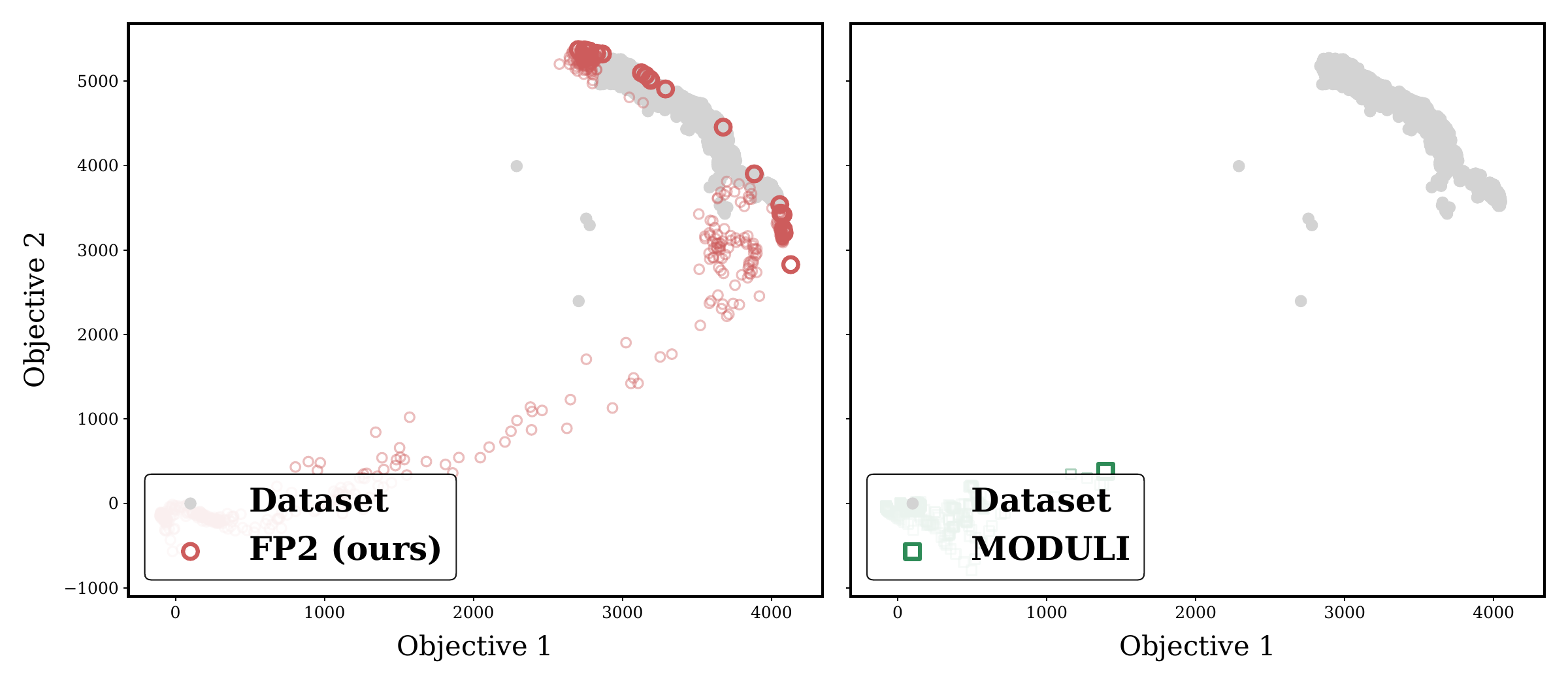}
    \caption{MO-Hopper (\texttt{Expert}-\texttt{Narrow})}
\end{subfigure}
\hfill
\begin{subfigure}{0.49\linewidth}
    \centering
    \includegraphics[width=\linewidth]{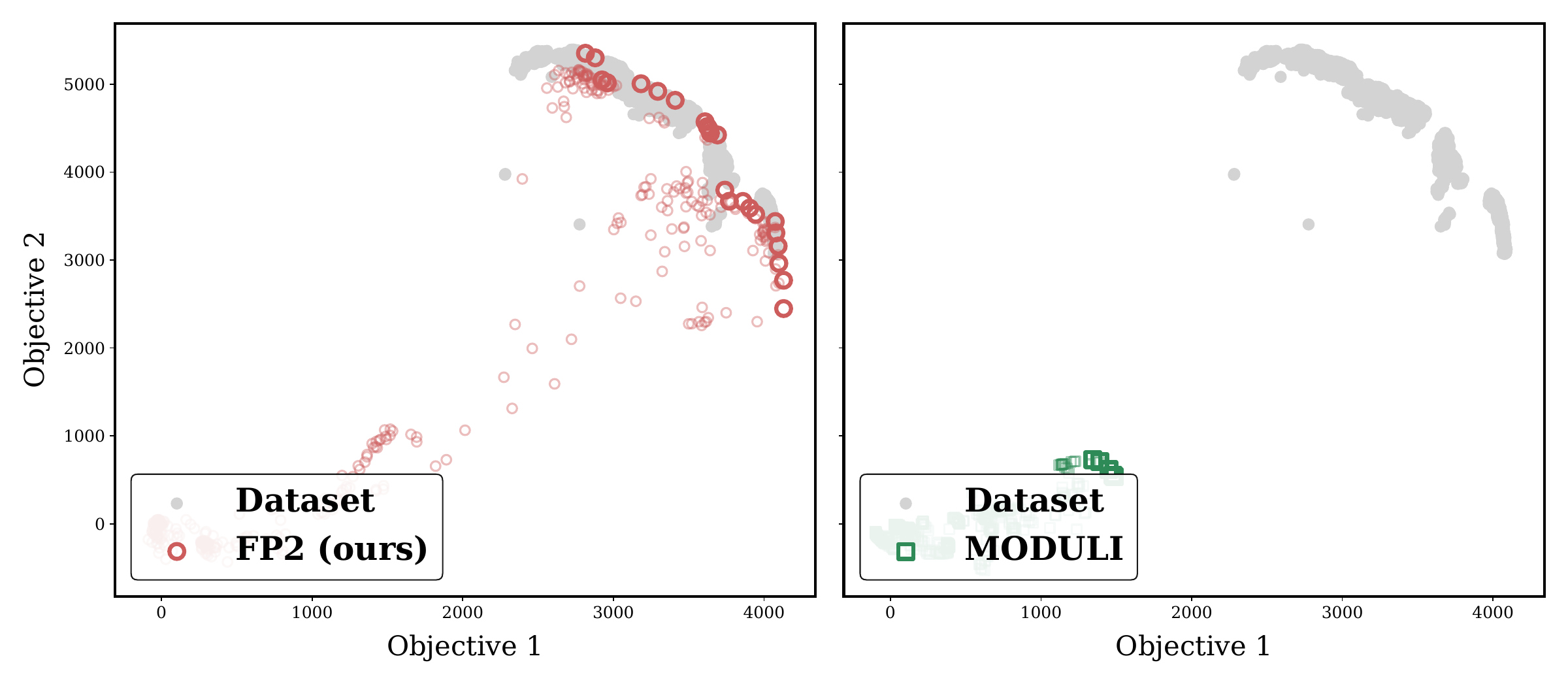}
    \caption{MO-Hopper (\texttt{Expert}-\texttt{Shattered})}
\end{subfigure}

\begin{subfigure}{0.49\linewidth}
    \centering
    \includegraphics[width=\linewidth]{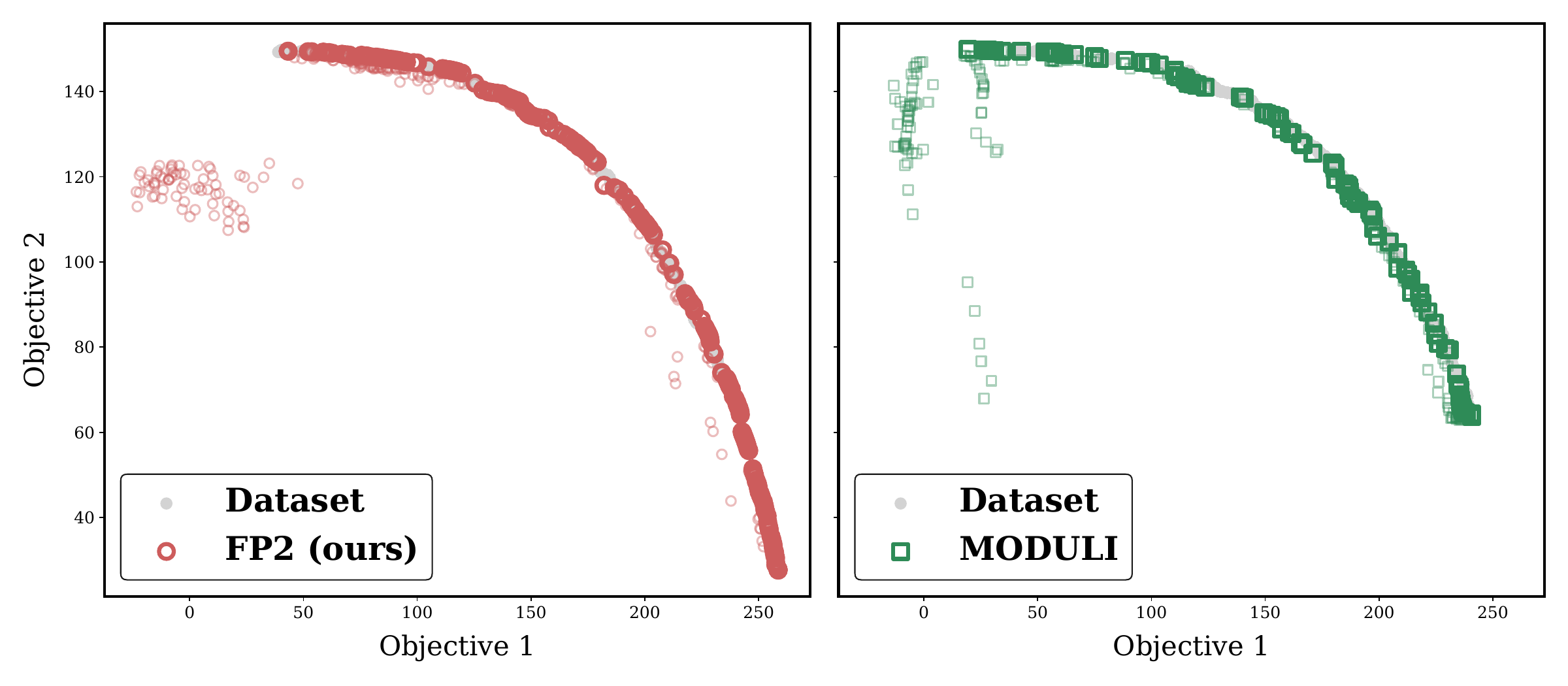}
    \caption{MO-Swimmer (\texttt{Expert}-\texttt{Narrow})}
\end{subfigure}
\hfill
\begin{subfigure}{0.49\linewidth}
    \centering
    \includegraphics[width=\linewidth]{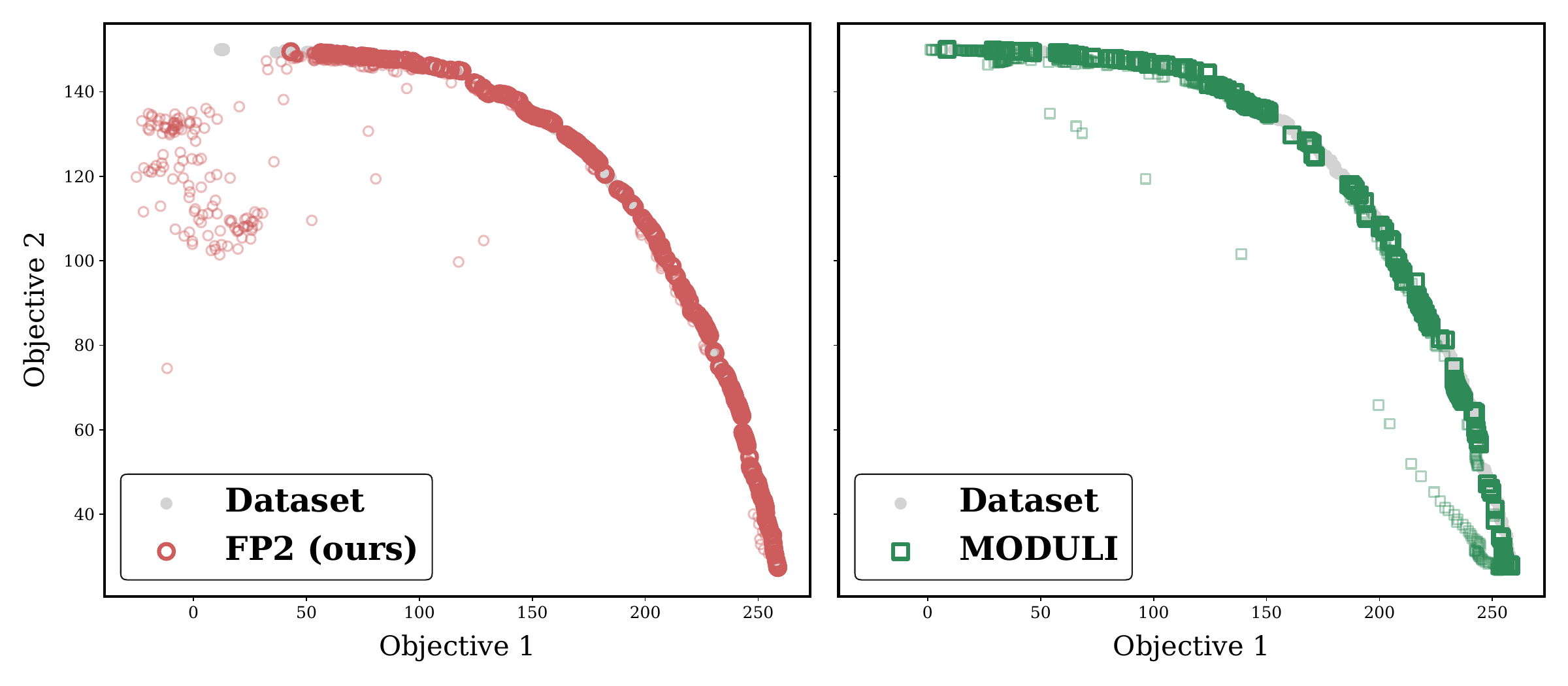}
    \caption{MO-Swimmer (\texttt{Expert}-\texttt{Shattered})}
\end{subfigure}

\begin{subfigure}{0.49\linewidth}
    \centering
    \includegraphics[width=\linewidth]{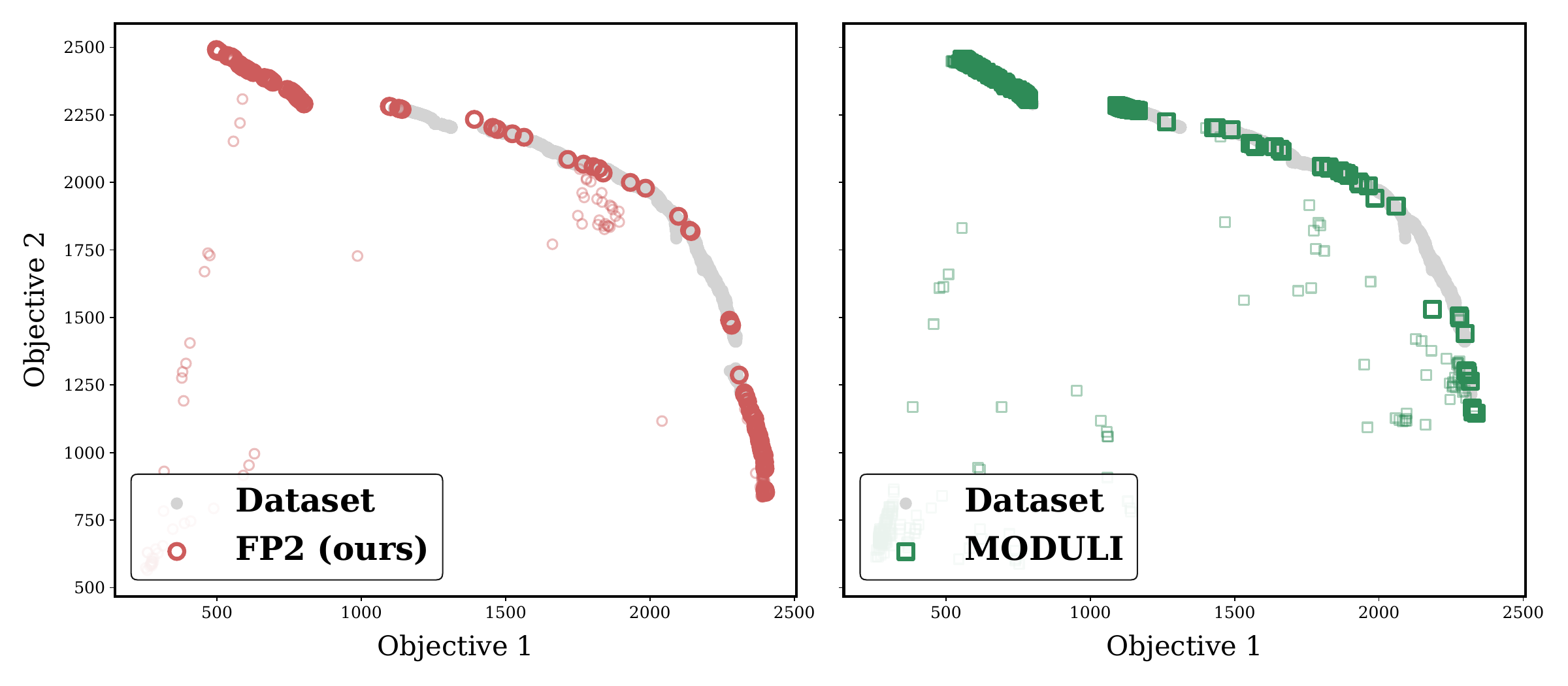}
    \caption{MO-Walker2d (\texttt{Expert}-\texttt{Narrow})}
\end{subfigure}
\hfill
\begin{subfigure}{0.49\linewidth}
    \centering
    \includegraphics[width=\linewidth]{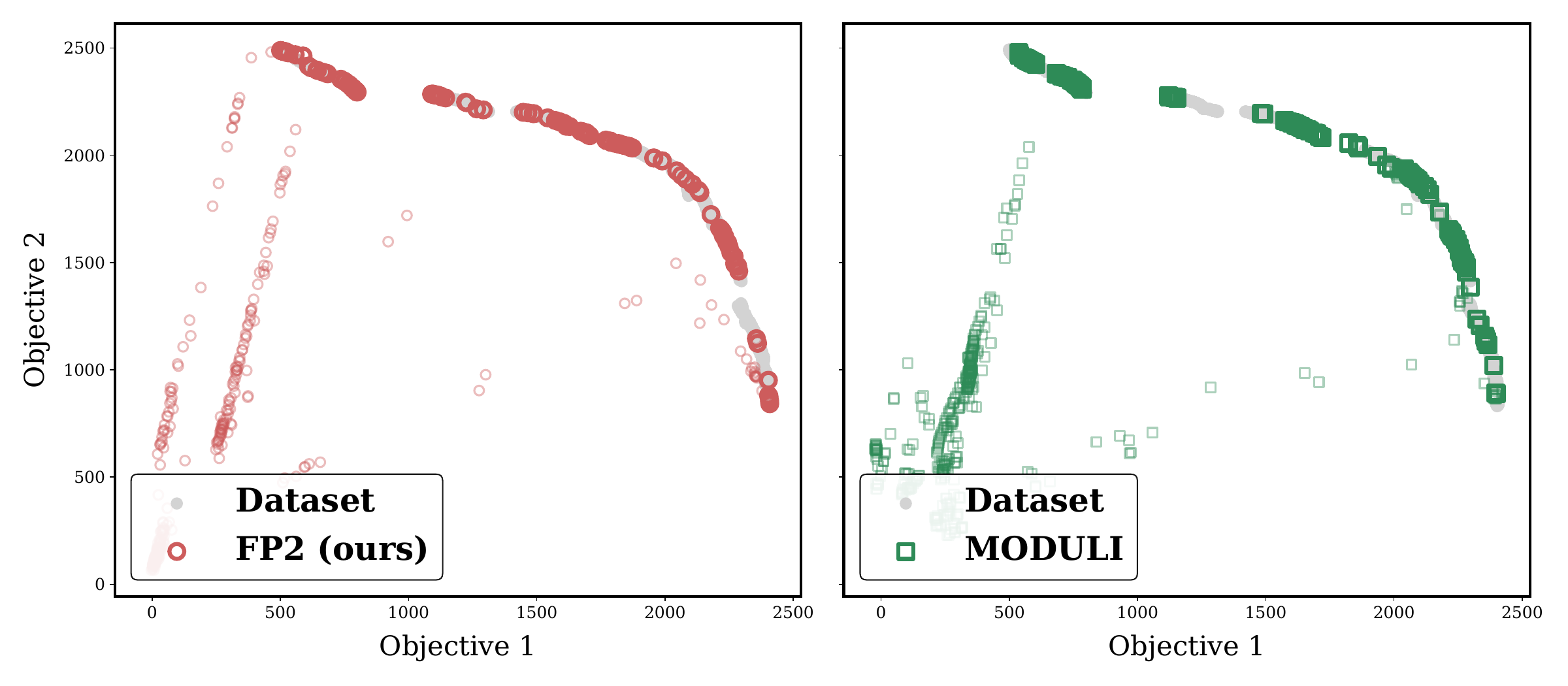}
    \caption{MO-Walker2d (\texttt{Expert}-\texttt{Shattered})}
\end{subfigure}
\begin{subfigure}{0.49\linewidth}
    \centering
    \includegraphics[width=\linewidth]{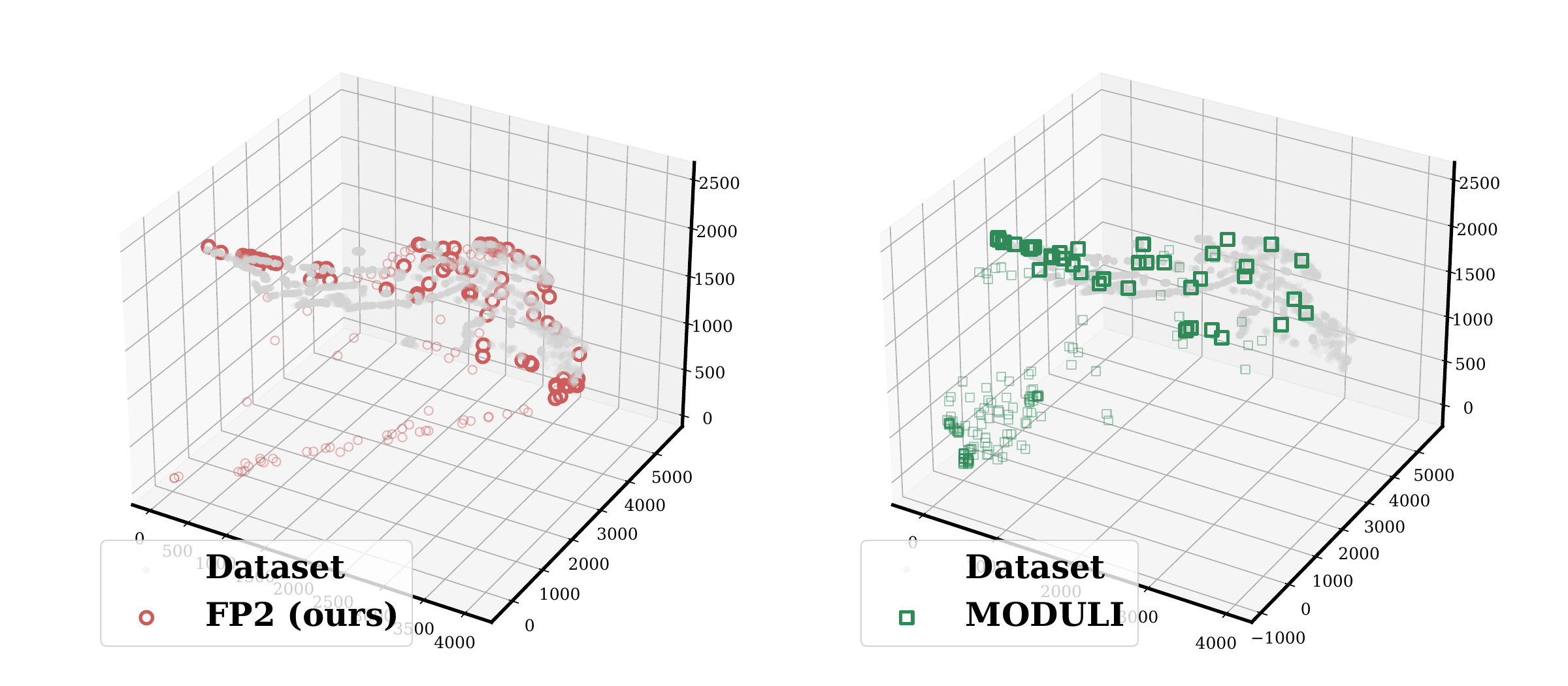}
    \caption{MO-Hopper-3obj (\texttt{Expert}-\texttt{Narrow})}
\end{subfigure}
\hfill
\begin{subfigure}{0.49\linewidth}
    \centering
    \includegraphics[width=\linewidth]{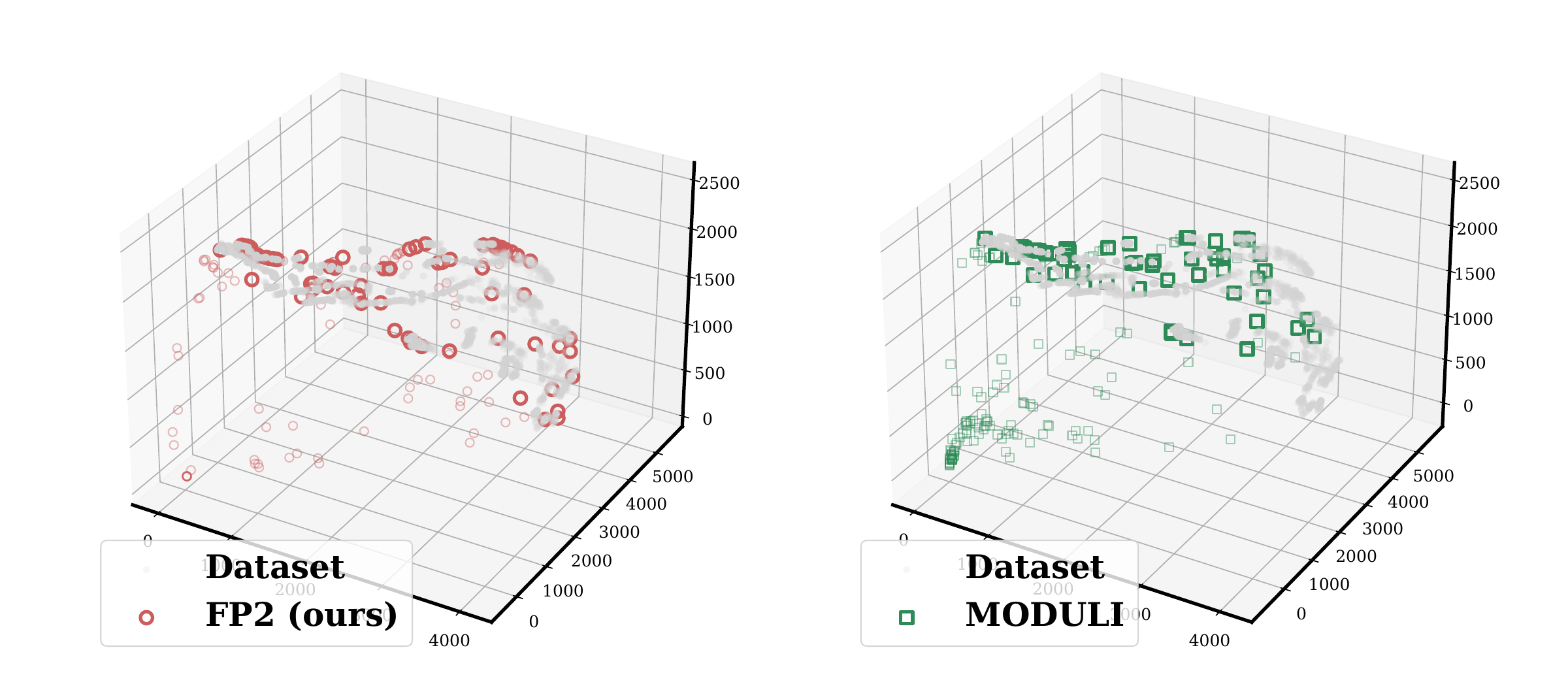}
    \caption{MO-Hopper-3obj (\texttt{Expert}-\texttt{Shattered})}
\end{subfigure}
\caption{Preference generalization on the \texttt{Expert} datasets. In each panel, FP2 (left) and MODULI (right) are shown with OOD points in color and dataset points in gray.}
\label{fig:ood_apx_expert}
\end{figure}

Overall, FP2-Mid achieves the best task-specific preference generalization performance.
It improves over MODULI across all three metrics, with median gains of +0.98\% on HV, +29.44\% on SP, and +1.04\% on PAS, and favorable Win/Tie/Loss results of 30/3/3, 29/4/3, and 24/9/3, respectively.
These results demonstrate that FP2-Mid generalizes more robustly to task-specific preferences than prior state-of-the-art methods.
Figs.~\ref{fig:ood_apx_amateur} and~\ref{fig:ood_apx_expert} further visualize the OOD preference generalization of FP2 and MODULI on all tasks.

\begin{table}[!htbp]
\centering
\caption{
Relative improvements (\%) of FP2-Mid over state-of-the-art method MODULI across 36 comparisons.
Positive values indicate improvements over MODULI~\cite{yuan2024moduli} after accounting for the optimization direction of each metric.
Positive improvements are highlighted.
}
\label{tab:full_compare_36_statistics}
\begin{tabular}{lccccccc}
\toprule
Metric & Min & P25 & Median & Mean & P75 & Max & Win/Tie/Loss \\
\midrule
HV$\uparrow$  
& -3.75  
& \cellcolor{gray!15}+0.35  
& \cellcolor{gray!15}+0.98  
& \cellcolor{gray!15}+2.96  
& \cellcolor{gray!15}+3.62  
& \cellcolor{gray!15}+20.95  
& 30/3/3 \\

SP$\downarrow$ 
& -25.00 
& \cellcolor{gray!15}+11.04 
& \cellcolor{gray!15}+29.44 
& \cellcolor{gray!15}+33.78 
& \cellcolor{gray!15}+59.52 
& \cellcolor{gray!15}+90.00  
& 29/4/3 \\

PAS$\uparrow$ 
& -7.07  
& +0.00  
& \cellcolor{gray!15}+1.04  
& \cellcolor{gray!15}+5.51  
& \cellcolor{gray!15}+3.17  
& \cellcolor{gray!15}+110.71 
& 24/9/3 \\
\bottomrule
\end{tabular}
\end{table}

\begin{table}[!htbp]
\centering
\caption{
Complete results for task-specific preference generalization on the full \texttt{Expert} and \texttt{Amateur} datasets. $*$ indicates results taken from ~\cite{yuan2024moduli}.}
\label{tab:full_compare_expert_amatuer}
\resizebox{\linewidth}{!}{
\begin{tabular}{l l c c c c c c c c c c}
\toprule
& \textbf{Environment} & \textbf{Metric} & \textbf{B} & \textbf{MODT(P)} & \textbf{MORvS(P)} & \textbf{BC(P)} & \textbf{CQL(P)} & \textbf{MODULI$^*$} & \textbf{MODULI} & \textbf{FP2-Pre} & \cellcolor{gray!15}\textbf{FP2-Mid} \\
\midrule
\multirow{18}{*}{\rotatebox[origin=c]{90}{\textbf{Expert}}}
& \multirow{3}{*}{MO-Ant}
& HV ($\times 10^6$) $\uparrow$ &$6.32$ &$6.21\pm0.01 $ &$6.36\pm0.02 $ &$4.88\pm0.17 $ &$5.76\pm0.10 $ &$6.39\pm0.02$&$6.40\pm0.01$ &$6.26\pm0.05$&$\cellcolor{gray!15}\boldsymbol{6.40\pm0.01}$ \\
& & SP ($\times 10^4$) $\downarrow$ &$-$&$8.26\pm2.22$&$0.87\pm0.19 $ &$46.20\pm16.40$ &$0.58\pm0.10 $&$0.79\pm0.12$ &$0.53\pm0.12$ &$0.50\pm0.04$ &$\cellcolor{gray!15}\boldsymbol{0.47\pm0.04}$ \\
& & PAS $\uparrow$ &$-$ &$-$ &$-$ &$-$ &$-$ &$-$&$0.96\pm0.00$ &$0.94\pm0.04$&$\cellcolor{gray!15}\boldsymbol{0.97\pm0.02}$ \\
\cmidrule(lr){2-12}
& \multirow{3}{*}{MO-HalfCheetah}
& HV ($\times 10^6$) $\uparrow$ &$5.79$ &$5.73\pm0.00$ &$5.78\pm0.00$ &$5.54\pm0.05 $ &$5.63\pm0.04 $ &$5.79\pm0.00$&$5.79\pm0.00$ &$5.77\pm0.00$&$\cellcolor{gray!15}\boldsymbol{5.79\pm0.01}$ \\
& & SP ($\times 10^4$) $\downarrow$ &$-$&$1.24\pm0.23 $&$0.67\pm0.05 $ &$1.78\pm0.39 $&$0.10\pm0.00 $ &$0.07\pm0.00$&$0.09\pm0.01$ &$0.09\pm0.01$ &$\cellcolor{gray!15}\boldsymbol{0.05\pm0.01}$ \\
& & PAS $\uparrow$ &$-$ &$-$ &$-$ &$-$ & $-$&$-$&$0.99\pm0.00$ &$0.99\pm0.00$&$\cellcolor{gray!15}\boldsymbol{0.99\pm0.00}$ \\
\cmidrule(lr){2-12}
& \multirow{3}{*}{MO-Hopper}
& HV ($\times 10^7$) $\uparrow$ &$2.09$ &$2.00\pm0.02 $ &$2.02\pm0.02 $ &$1.23\pm0.10 $ &$0.33\pm0.39 $ &$2.09\pm0.01$&$2.09\pm0.01$ &$2.01\pm0.03$ &$\cellcolor{gray!15}\boldsymbol{2.10\pm0.01}$\\
& & SP ($\times 10^5$) $\downarrow$ &$-$ &$16.30\pm10.60 $&$3.03\pm0.36 $ &$52.50\pm4.88 $ &$2.84\pm2.46 $ &$0.09\pm0.01$&$0.12\pm0.02$ &$0.08\pm0.00$ &$\cellcolor{gray!15}\boldsymbol{0.07\pm0.01}$ \\
& & PAS $\uparrow$ &$-$ &$-$ & $-$& $-$&$-$&$-$ &$0.99\pm0.00$ &$0.98\pm0.00$&$\cellcolor{gray!15}\boldsymbol{0.99\pm0.00}$ \\
\cmidrule(lr){2-12}
& \multirow{3}{*}{MO-Walker}
& HV ($\times 10^4$) $\uparrow$ &$5.21$ &$4.89\pm0.05 $ &$5.14\pm0.01 $ &$3.74\pm0.11$ &$3.21\pm0.32 $&$5.20\pm0.00$ &$5.19\pm0.00$ &$5.17\pm0.03$ &$\cellcolor{gray!15}\boldsymbol{5.22\pm0.01}$\\
& & SP ($\times 10^0$) $\downarrow$ &$-$ &$0.99\pm0.44 $&$3.22\pm0.73 $ &$75.60\pm52.30 $ &$6.23\pm10.70 $ &$0.11\pm0.01$&$0.34\pm0.42$ &$0.11\pm0.02$ &$\cellcolor{gray!15}\boldsymbol{0.06\pm0.01}$ \\
& & PAS $\uparrow$ &$-$ &$-$ &$-$ & $-$&$-$& $-$&$0.96\pm0.00$ &$0.90\pm0.02$&$\cellcolor{gray!15}\boldsymbol{0.99\pm0.00}$ \\
\cmidrule(lr){2-12}
& \multirow{3}{*}{MO-Swimmer}
& HV ($\times 10^6$) $\uparrow$ & $3.25$&$3.15\pm0.02 $ &$3.24\pm0.00 $ &$3.21\pm0.00 $ &$3.22\pm0.08 $ &$3.24\pm0.00$&$3.25\pm0.00$ &$3.24\pm0.01$&$\cellcolor{gray!15}\boldsymbol{3.26\pm0.01}$ \\
& & SP ($\times 10^4$) $\downarrow$ &$-$ &$15.00\pm7.49 $&$4.39\pm0.07 $ &$4.50\pm0.39$ &$13.60\pm5.31 $&$4.43\pm0.38$&$4.43\pm0.21$ &$9.92\pm2.50$ &$\cellcolor{gray!15}\boldsymbol{4.14\pm0.11}$ \\
& & PAS $\uparrow$ &$-$ &$-$ &$-$ &$-$ &$-$&$-$ &$0.99\pm0.00$ &$0.97\pm0.00$&$\cellcolor{gray!15}\boldsymbol{0.99\pm0.00}$ \\
\cmidrule(lr){2-12}
& \multirow{3}{*}{MO-Hopper-3obj}
& HV ($\times 10^{10}$) $\uparrow$ &$3.73$ &$3.38\pm0.05$ &$3.42\pm0.10 $ &$2.29\pm0.07 $ &$0.78\pm0.24 $ &$3.57\pm0.02$&$3.44\pm0.04$ &$3.72\pm0.01$&$\cellcolor{gray!15}\boldsymbol{3.76\pm0.02}$ \\
& & SP ($\times 10^5$) $\downarrow$ &$-$ &$1.40\pm0.44$&$2.72\pm1.93$ &$0.72\pm0.09$ &$2.60\pm3.14$ &$0.07\pm0.00$&$0.07\pm0.00$ &$0.07\pm0.02$ &$\cellcolor{gray!15}\boldsymbol{0.05\pm0.01}$ \\
& & PAS $\uparrow$ &$-$ &$-$ &$-$ &$-$ &$-$&$-$ &$0.98\pm0.01$ &$0.99\pm0.00$&$\cellcolor{gray!15}\boldsymbol{0.99\pm0.00}$ \\
\midrule
\multirow{18}{*}{\rotatebox[origin=c]{90}{\textbf{Amateur}}}
& \multirow{3}{*}{MO-Ant}
& HV ($\times 10^6$) $\uparrow$ &$5.61$ &$5.92\pm0.04 $ &$6.07\pm0.02 $ &$4.37\pm0.06 $ &$5.62\pm0.23$ &$6.08\pm0.03$&$6.10\pm0.01$&$6.10\pm0.06$&$\cellcolor{gray!15}\boldsymbol{6.14\pm0.01}$ \\
& & SP ($\times 10^4$) $\downarrow$ &$-$ &$8.72\pm0.77  $&$5.24\pm0.52 $ &$25.90\pm16.40 $ &$1.06\pm0.28 $ &$0.53\pm0.05$&$0.66\pm0.08$ &$0.67\pm0.21$ &$\cellcolor{gray!15}\boldsymbol{0.39\pm0.06}$ \\
& & PAS $\uparrow$ &$-$ &$-$ &$-$ &$-$ &$-$& $-$&$0.94\pm0.01$&$0.91\pm0.00$&$\cellcolor{gray!15}\boldsymbol{0.97\pm0.00}$ \\
\cmidrule(lr){2-12}
& \multirow{3}{*}{MO-HalfCheetah}
& HV ($\times 10^6$) $\uparrow$ &$5.68$ &$5.69\pm0.01  $ &$5.77\pm0.00  $ &$5.46\pm0.02  $ &$5.54\pm0.02$&$5.76\pm0.00$ &$5.76\pm0.00$&$5.77\pm0.01$&$\cellcolor{gray!15}\boldsymbol{5.78\pm0.01}$ \\
& & SP ($\times 10^4$) $\downarrow$ &$-$ &$1.16\pm0.42  $&$0.57\pm0.09 $ &$2.22\pm0.91  $ &$0.45\pm0.27  $ &$0.07\pm0.02$&$0.07\pm0.01$ &$0.06\pm0.02$ &$\cellcolor{gray!15}\boldsymbol{0.05\pm0.01}$ \\
& & PAS $\uparrow$ &$-$ &$-$ &$-$&$-$ &$-$ &$-$ &$0.96\pm0.01$ &$0.99\pm0.00$&$\cellcolor{gray!15}\boldsymbol{0.99\pm0.00}$\\
\cmidrule(lr){2-12}
& \multirow{3}{*}{MO-Hopper}
& HV ($\times 10^7$) $\uparrow$ &$1.97$ &$1.81\pm0.05 $ &$1.76\pm0.03  $ &$1.35\pm0.03  $ &$1.64\pm0.01 $ &$2.01\pm0.01$&$2.01\pm0.01$&$\cellcolor{gray!15}\boldsymbol{2.03\pm0.01}$&$2.01\pm0.01$\\
& & SP ($\times 10^5$) $\downarrow$ &$-$ &$1.61\pm0.29  $&$3.50\pm1.54  $ &$2.42\pm1.08  $ &$3.30\pm5.25 $ &$\cellcolor{gray!15}\boldsymbol{0.10\pm0.01}$&$0.11\pm0.01$ &$0.13\pm0.01$ &$0.13\pm0.02$ \\
& & PAS $\uparrow$ &$-$ &$-$ &$-$ &$-$ &$-$&$-$ &$0.69\pm0.01$&$\cellcolor{gray!15}\boldsymbol{0.92\pm0.01}$&$0.77\pm0.01$\\
\cmidrule(lr){2-12}
& \multirow{3}{*}{MO-Walker}
& HV ($\times 10^4$) $\uparrow$ &$4.99$ &$3.10\pm0.34  $ &$4.98\pm0.01  $ &$3.42\pm0.42  $ &$1.78\pm0.33  $ &$5.06\pm0.00$ &$5.07\pm0.01$&$5.10\pm0.02$&$\cellcolor{gray!15}\boldsymbol{5.15\pm0.01}$\\
&$-$ & SP ($\times 10^0$) $\downarrow$ & &$164.20\pm13.50  $&$1.94\pm0.06 $ &$53.10\pm34.60 $ &$7.33\pm5.89 $ &$0.25\pm0.03$&$0.28\pm0.02$ &$0.10\pm0.02$ &$\cellcolor{gray!15}\boldsymbol{0.09\pm0.02}$ \\
& & PAS $\uparrow$ &$-$ &$-$ &$-$&$-$ &$-$ & $-$&$0.94\pm0.01$&$0.90\pm0.01$&$\cellcolor{gray!15}\boldsymbol{0.99\pm0.00}$\\
\cmidrule(lr){2-12}
& \multirow{3}{*}{MO-Swimmer}
& HV ($\times 10^6$) $\uparrow$ & $2.11$&$1.67\pm0.22 $ &$2.79\pm0.03 $ &$2.82\pm0.04 $ &$1.69\pm0.93$ &$\cellcolor{gray!15}\boldsymbol{3.20\pm0.00}$ &$3.12\pm0.00$&$3.05\pm0.04$&$3.08\pm0.00$ \\
& & SP ($\times 10^4$) $\downarrow$ & $-$&$2.87\pm1.32 $&$\cellcolor{gray!15}\boldsymbol{1.03\pm0.20}$ &$5.05\pm1.82 $ &$9.50\pm0.59$&$8.87\pm6.24 $ &$9.72   \pm0.69$ &$5.51\pm0.70$ &$5.82\pm0.12$ \\
& & PAS $\uparrow$ &$-$ &$-$ &$-$ &$-$ &$-$&$-$ &$0.95\pm0.00$ &$0.98\pm0.01$&$\cellcolor{gray!15}\boldsymbol{0.99\pm0.00}$\\
\cmidrule(lr){2-12}
& \multirow{3}{*}{MO-Hopper-3obj}
& HV ($\times 10^{10}$) $\uparrow$ & $3.09$&$1.04\pm0.16 $ &$2.77\pm0.24 $ &$2.42\pm0.18 $ &$0.59\pm0.42 $ &$3.33\pm0.06$&$3.30\pm0.06$ &$3.52\pm0.01$&$\cellcolor{gray!15}\boldsymbol{3.60\pm0.02}$\\
& & SP ($\times 10^5$) $\downarrow$ & $-$&$10.23\pm2.78$&$1.03\pm0.11 $ &$0.87\pm0.29$ &$2.00\pm1.72 $ &$0.10\pm0.00$&$0.12\pm0.02$ &$0.09\pm0.01$ &$\cellcolor{gray!15}\boldsymbol{0.06\pm0.01}$ \\
& & PAS $\uparrow$ &$-$ &$-$ &$-$ &$-$&$-$ & $-$&$0.98\pm0.00$&$0.99\pm0.00$&$\cellcolor{gray!15}\boldsymbol{0.99\pm0.00}$\\
\bottomrule
\end{tabular}
}
\end{table}

\begin{table}[!htbp]
\centering
\footnotesize
\setlength{\tabcolsep}{5pt}
\caption{
Complete results for task-specific preference generalization on the \texttt{Expert}-\texttt{Shattered} and \texttt{Expert}-\texttt{Narrow} datasets.
$*$ indicates results taken from~\cite{yuan2024moduli}.}
\label{tab:full_compare_expert_shat_nar}
\resizebox{\linewidth}{!}{
\begin{tabular}{l l c c c c c c c c}
\toprule
& \textbf{Environment} & \textbf{Metric} & \textbf{MODT(P)} & \textbf{MORvS(P)} & \textbf{BC(P)} & \textbf{MODULI$^*$}& \textbf{MODULI} & \textbf{FP2-Pre} & \cellcolor{gray!15}\textbf{FP2-Mid}\\
\midrule
\multirow{18}{*}{\rotatebox[origin=c]{90}{\textbf{Expert}-\textbf{Shattered}}}
& \multirow{3}{*}{MO-Ant}
& HV ($\times 10^6$) $\uparrow$ &$5.88\pm0.02 $ &$6.38\pm0.01 $ &$4.76\pm0.05 $ &$\cellcolor{gray!15}\boldsymbol{6.39\pm0.01}$&$6.38\pm0.01$ &$6.17\pm0.12$ &$6.35\pm0.01$\\
& & SP ($\times 10^4$) $\downarrow$ &$2.22\pm0.50 $ &$0.75\pm0.13 $&$5.16\pm1.18$ &$0.53\pm0.13$&$0.53\pm0.11$ &$1.15\pm0.65$ &$\cellcolor{gray!15}\boldsymbol{0.51\pm0.07}$ \\
& & PAS $\uparrow$ &$-$ &$-$ &$-$ &$-$&$0.94\pm0.01$ &$0.92\pm0.01$ &$\cellcolor{gray!15}\boldsymbol{0.96\pm0.01}$\\
\cmidrule(lr){2-10}
& \multirow{3}{*}{MO-HalfCheetah}
& HV ($\times 10^6$) $\uparrow$ &$5.69\pm0.00 $ &$5.73\pm0.00 $ &$5.66\pm0.00 $ &$5.78\pm0.00$&$5.78\pm0.00$ &$5.76\pm0.01$&$\cellcolor{gray!15}\boldsymbol{5.79\pm0.01}$ \\
& & SP ($\times 10^4$) $\downarrow$ &$0.18\pm0.01 $&$0.15\pm0.02 $ & $0.34\pm0.03 $&$0.11\pm0.01$& $0.11\pm0.01$&$0.13\pm0.04$ &$\cellcolor{gray!15}\boldsymbol{0.07\pm0.01}$\\
& & PAS $\uparrow$ &$-$ &$-$ &$-$ &$-$&$0.99\pm0.00$ &$0.99\pm0.00$&$\cellcolor{gray!15}\boldsymbol{0.99\pm0.00}$\\
\cmidrule(lr){2-10}
& \multirow{3}{*}{MO-Hopper}
& HV ($\times 10^7$) $\uparrow$ &$1.95\pm0.03 $ &$2.06\pm0.01 $ &$1.49\pm0.01$ &$2.07\pm0.00$&$2.07\pm0.00$ &$2.01\pm0.11$&$\cellcolor{gray!15}\boldsymbol{2.08\pm0.00}$ \\
& & SP ($\times 10^5$) $\downarrow$ &$0.85\pm0.27 $&$0.20\pm0.06$ &$2.58\pm1.01 $ &$\cellcolor{gray!15}\boldsymbol{0.11\pm0.01}$&$0.13\pm0.01$ &$0.12\pm0.08$ &$0.13\pm0.02$ \\
& & PAS $\uparrow$ &$-$ &$-$ &$-$ &$-$ &$0.43\pm0.02$ &$\cellcolor{gray!15}\boldsymbol{0.92\pm0.01}$ &$0.56\pm0.01$\\
\cmidrule(lr){2-10}
& \multirow{3}{*}{MO-Walker}
& HV ($\times 10^4$) $\uparrow$ &$5.01\pm0.01$ &$5.14\pm0.01 $ &$3.47\pm0.12$ &$5.20\pm0.00$&$5.20\pm0.00$ &$5.16\pm0.03$&$\cellcolor{gray!15}\boldsymbol{5.22\pm0.00}$ \\
& & SP ($\times 10^0$) $\downarrow$ &$0.49\pm0.03 $&$0.27\pm0.04 $ &$16.61\pm13.78$ &$0.12\pm0.01$&$0.11\pm0.01$ &$0.12\pm0.01$ &$\cellcolor{gray!15}\boldsymbol{0.05\pm0.01}$\\
& & PAS $\uparrow$ &$-$ &$-$ &$-$ &$-$&$0.99\pm0.00$ &$0.92\pm0.02$ &$\cellcolor{gray!15}\boldsymbol{0.99\pm0.00}$\\
\cmidrule(lr){2-10}
& \multirow{3}{*}{MO-Swimmer}
& HV ($\times 10^6$) $\uparrow$ &$3.20\pm0.00 $ &$3.24\pm0.00 $ &$3.19\pm0.00 $ &$3.24\pm0.00$&$3.24\pm0.00$ &$3.22\pm0.02$ &$\cellcolor{gray!15}\boldsymbol{3.26\pm0.01}$\\
& & SP ($\times 10^4$) $\downarrow$ &$15.50\pm0.33 $&$7.36\pm0.81$&$9.05\pm0.30$&$5.79\pm0.43$ &$5.63\pm0.47$ &$7.11\pm0.07$&$\cellcolor{gray!15}\boldsymbol{2.46\pm0.21}$ \\
& & PAS $\uparrow$ &$-$ &$-$ &$-$ &$-$ &$0.99\pm0.00$ &$0.92\pm0.01$&$\cellcolor{gray!15}\boldsymbol{0.99\pm0.00}$\\
\cmidrule(lr){2-10}
& \multirow{3}{*}{MO-Hopper-3obj}
& HV ($\times 10^{10}$) $\uparrow$ &$2.83\pm0.06 $ &$3.28\pm0.07 $ &$1.95\pm0.16 $ &$3.43\pm0.02$&$3.33\pm0.02$ &$\cellcolor{gray!15}\boldsymbol{3.80\pm0.08}$ &$3.70\pm0.00$\\
& & SP ($\times 10^5$) $\downarrow$ &$0.09\pm0.01 $&$0.06\pm0.00$ &$0.17\pm0.06 $ &$0.10\pm0.01$&$0.10\pm0.02$ &$0.07\pm0.02$&$\cellcolor{gray!15}\boldsymbol{0.05\pm0.01}$ \\
& & PAS $\uparrow$ &$-$ &$-$ &$-$ &$-$&$0.98\pm0.00$ &$0.95\pm0.01$ &$\cellcolor{gray!15}\boldsymbol{0.99\pm0.00}$\\
\midrule
\multirow{18}{*}{\rotatebox[origin=c]{90}{\textbf{Expert}-\textbf{Narrow}}}
& \multirow{3}{*}{MO-Ant}
& HV ($\times 10^6$) $\uparrow$ &$5.05\pm0.08$ &$6.06\pm0.00$ &$4.90\pm0.02 $ &$\cellcolor{gray!15}\boldsymbol{6.36\pm0.00}$&$6.18\pm0.01$ &$6.06\pm0.19$ &$6.24\pm0.01$ \\
& & SP ($\times 10^4$) $\downarrow$ &$0.88\pm0.41 $&$0.79\pm0.11$ &$1.56\pm0.28$ &$0.80\pm0.25$&$0.96\pm0.21$&$0.82\pm0.41$ &$0.75\pm0.52$ \\
& & PAS $\uparrow$ &$-$ &$-$ &$-$ &$-$ &$0.93\pm0.01$ &$0.90\pm0.01$ 
&$\cellcolor{gray!15}\boldsymbol{0.92\pm0.01}$\\
\cmidrule(lr){2-10}
& \multirow{3}{*}{MO-HalfCheetah}
& HV ($\times 10^6$) $\uparrow$ &$5.04\pm0.00  $ &$5.46\pm0.01 $ &$4.88\pm0.02 $ &$5.76\pm0.00$&$5.74\pm0.01$ &$5.74\pm0.00$&$\cellcolor{gray!15}\boldsymbol{5.79\pm0.00}$ \\
& & SP ($\times 10^4$) $\downarrow$ &$0.04\pm0.00$&$0.05\pm0.00$ &$\cellcolor{gray!15}\boldsymbol{0.02\pm0.00 }$ &$0.04\pm0.00$ &$0.04\pm0.01$ &$0.05\pm0.01$ &$0.05\pm0.01$ \\
& & PAS $\uparrow$ &$-$ &$-$ &$-$ &$-$&$0.99\pm0.00$ &$\cellcolor{gray!15}\boldsymbol{0.99\pm0.00}$ &$0.98\pm0.00$ \\
\cmidrule(lr){2-10}
& \multirow{3}{*}{MO-Hopper}
& HV ($\times 10^7$) $\uparrow$ &$1.85\pm0.02 $ &$1.98\pm0.00 $&$1.29\pm0.01 $ &$2.04\pm0.01$&$2.07\pm0.01$ &$2.01\pm0.03$ &$\cellcolor{gray!15}\boldsymbol{2.10\pm0.00}$\\
& & SP ($\times 10^5$) $\downarrow$ &$0.34\pm0.07$&$0.16\pm0.02 $&$0.59\pm0.24 $&$0.25\pm0.05$ &$0.28\pm0.02$ &$0.47\pm0.70$ &$\cellcolor{gray!15}\boldsymbol{0.08\pm0.01}$ \\
& & PAS $\uparrow$ &$-$ &$-$ &$-$ &$-$&$0.92\pm0.00$ &$0.90\pm0.00$ &$\cellcolor{gray!15}\boldsymbol{0.96\pm0.00}$\\
\cmidrule(lr){2-10}
& \multirow{3}{*}{MO-Walker}
& HV ($\times 10^4$) $\uparrow$ &$4.75\pm0.01 $ &$4.85\pm0.01 $ &$1.88\pm0.95 $ &$5.10\pm0.01$&$5.13\pm0.01$ &$5.10\pm0.01$ &$\cellcolor{gray!15}\boldsymbol{5.22\pm0.00}$ \\
& & SP ($\times 10^0$) $\downarrow$ &$0.34\pm0.09 $&$0.19\pm0.02 $ &$10.76\pm11.58 $&$0.13\pm0.01$ &$0.14\pm0.01$ &$0.12\pm0.01$ &$\cellcolor{gray!15}\boldsymbol{0.06\pm0.01}$ \\
& & PAS $\uparrow$ &$-$ &$-$ &$-$ &$-$&$0.93\pm0.01$ &$0.92\pm0.01$ &$\cellcolor{gray!15}\boldsymbol{0.98\pm0.00}$ \\
\cmidrule(lr){2-10}
& \multirow{3}{*}{MO-Swimmer}
& HV ($\times 10^6$) $\uparrow$ &$3.03\pm0.00 $ &$3.10\pm0.00 $  &$3.06\pm0.00 $ &$3.21\pm0.00$&$3.16\pm0.00$ &$3.16\pm0.00$ &$\cellcolor{gray!15}\boldsymbol{3.26\pm0.00}$ \\
& & SP ($\times 10^4$) $\downarrow$ &$3.65\pm0.22 $&$4.38\pm0.80 $ &$5.53\pm0.20 $ &$3.28\pm0.15$&$5.84\pm0.01$ &$6.12\pm0.68$ &$\cellcolor{gray!15}\boldsymbol{1.68\pm0.07}$ \\
& & PAS $\uparrow$ &$-$ &$-$ &$-$ &$-$ &$0.99\pm0.00$ &$0.89\pm0.04$ &$\cellcolor{gray!15}\boldsymbol{0.99\pm0.00}$\\
\cmidrule(lr){2-10}
& \multirow{3}{*}{MO-Hopper-3obj}
& HV ($\times 10^{10}$) $\uparrow$ &$3.18\pm0.05 $ &$3.32\pm0.00 $&$2.32\pm0.10$ &$3.37\pm0.05$&$3.15\pm0.13$ &$3.39\pm0.29$ &$\cellcolor{gray!15}\boldsymbol{3.81\pm0.00}$ \\
& & SP ($\times 10^5$) $\downarrow$ &$0.11\pm0.01 $&$0.22\pm0.02$ &$0.14\pm0.02$ &$0.12\pm0.01$&$0.16\pm0.00$ &$\cellcolor{gray!15}\boldsymbol{0.03\pm0.02}$ &$0.05\pm0.01$ \\
& & PAS $\uparrow$ &$-$ &$-$ &$-$ &$-$&$0.96\pm0.00$ &$0.91\pm0.01$ &$\cellcolor{gray!15}\boldsymbol{0.98\pm0.01}$ \\
\bottomrule
\end{tabular}
}
\end{table}

\begin{table}[!htbp]
\centering
\caption{
Complete results for task-specific preference generalization on the \texttt{Amateur}-\texttt{Shattered} and \texttt{Amateur}-\texttt{Narrow} datasets. 
$*$ indicates results taken from~\cite{yuan2024moduli}.}
\label{tab:full_compare_amateur_shat_nar}
\resizebox{\linewidth}{!}{
\begin{tabular}{l l c c c c c c c c}
\toprule
& \textbf{Environment} & \textbf{Metric}& \textbf{MODT(P)} & \textbf{MORvS(P)} & \textbf{BC(P)} & \textbf{MODULI$^*$} & \textbf{MODULI} & \textbf{FP2-Pre} & \cellcolor{gray!15}\textbf{FP2-Mid} \\
\midrule
\multirow{18}{*}{\rotatebox[origin=c]{90}{\textbf{Amateur}-\textbf{Shattered}}}
& \multirow{3}{*}{MO-Ant}
& HV ($\times 10^6$) $\uparrow$ &$5.74\pm0.04$ &$6.09\pm0.01$ &$4.20\pm0.05$ &$6.04\pm0.01$  &$6.04\pm0.01$ &$6.07\pm0.06$& $\cellcolor{gray!15}\boldsymbol{6.12\pm0.00}$\\
& & SP ($\times 10^4$) $\downarrow$ &$2.47\pm0.61$ &$0.79\pm0.14$ &$2.29\pm1.38$ & $0.63\pm0.23$ &$0.59\pm0.05$ &$1.15\pm0.65$  & $\cellcolor{gray!15}\boldsymbol{0.51\pm0.01}$\\
& & PAS $\uparrow$ &$-$ &$-$ &$-$ &$-$  &$0.96\pm0.00$ &$0.92\pm0.00$ & $\cellcolor{gray!15}\boldsymbol{0.97\pm0.00}$\\
\cmidrule(lr){2-10}
& \multirow{3}{*}{MO-HalfCheetah}
& HV ($\times 10^6$) $\uparrow$ &$5.61\pm0.01$ &$5.74\pm0.00$ &$5.54\pm0.00$ &$\cellcolor{gray!15}\boldsymbol{5.75\pm0.00}$  &$5.74\pm0.01$ &$5.74\pm0.01$&$5.74\pm0.01$\\
& & SP ($\times 10^4$) $\downarrow$ &$0.11\pm0.01$&$0.09\pm0.01$ &$0.09\pm0.01$ & $0.13\pm0.01$&$0.07\pm0.02$ &$0.12\pm0.03$& $\cellcolor{gray!15}\boldsymbol{0.07\pm0.02}$\\
& & PAS $\uparrow$ &$-$ &$-$ &$-$ &$-$ &$0.98\pm0.00$ &$0.95\pm0.04$ &$\cellcolor{gray!15}\boldsymbol{0.99\pm0.00}$\\
\cmidrule(lr){2-10}
& \multirow{3}{*}{MO-Hopper}
& HV ($\times 10^7$) $\uparrow$ &$1.46\pm0.00$ &$1.69\pm0.01$ &$1.63\pm0.00$ &$2.01\pm0.01$ &$2.01\pm0.00$ &$2.01\pm0.04$&$\cellcolor{gray!15}\boldsymbol{2.11\pm0.00}$\\
& & SP ($\times 10^5$) $\downarrow$ &$0.54\pm0.58$&$0.18\pm0.01$ &$0.11\pm0.05$ & $0.15\pm0.04$&$0.16\pm0.01$ &$\cellcolor{gray!15}\boldsymbol{0.14\pm0.02}$&$0.16\pm0.01$ \\
& & PAS $\uparrow$ &$-$ &$-$ &$-$ &$-$ &$0.28\pm0.00$ &$\cellcolor{gray!15}\boldsymbol{0.92\pm0.01}$ &$0.59\pm0.01$\\
\cmidrule(lr){2-10}
& \multirow{3}{*}{MO-Walker}
& HV ($\times 10^4$) $\uparrow$ &$3.86\pm0.07$ &$4.99\pm0.00$ &$3.42\pm0.02$ &$5.03\pm0.00$ &$5.04\pm0.03$ &$5.09\pm0.04$&$\cellcolor{gray!15}\boldsymbol{5.15\pm0.01}$\\
& & SP ($\times 10^0$) $\downarrow$  &$11.13\pm3.58$&$0.20\pm0.03$ &$4.56\pm0.08$ & $0.29\pm0.01$&$0.12\pm0.02$ &$0.18\pm0.03$ &$\cellcolor{gray!15}\boldsymbol{0.10\pm0.01}$ \\
& & PAS $\uparrow$  &$-$ &$-$ & $-$& $-$&$\cellcolor{gray!15}\boldsymbol{0.99\pm0.00}$ &$0.90\pm0.01$&$0.92\pm0.01$ \\
\cmidrule(lr){2-10}
& \multirow{3}{*}{MO-Swimmer}
& HV ($\times 10^6$) $\uparrow$ &$0.75\pm0.00$ &$2.81\pm0.00$ &$2.81\pm0.00$ &$\cellcolor{gray!15}\boldsymbol{3.17\pm0.00}$ &$3.09\pm0.00$& $3.03\pm0.02$&$3.05\pm0.01$ \\
& & SP ($\times 10^4$) $\downarrow$ &$6.75\pm0.23$&$\cellcolor{gray!15}\boldsymbol{1.26\pm0.06}$&$1.89\pm0.05$ &$8.85\pm0.77$ &$4.96\pm0.08$ &$8.05\pm2.92$&$5.04\pm0.05$  \\
& & PAS $\uparrow$ &$-$ &$-$ &$-$ &$-$ &$0.94\pm0.00$ &$0.87\pm0.03$&$\cellcolor{gray!15}\boldsymbol{0.99\pm0.00}$ \\
\cmidrule(lr){2-10}
& \multirow{3}{*}{MO-Hopper-3obj}
& HV ($\times 10^{10}$) $\uparrow$  &$1.48\pm0.02$ &$2.84\pm0.04$ & $1.34\pm0.02$&$3.02\pm0.03$ &$3.23\pm0.01$ &$3.46\pm0.06$&$\cellcolor{gray!15}\boldsymbol{3.49\pm0.04}$ \\
& & SP ($\times 10^5$) $\downarrow$  &$0.56\pm0.08$&$0.13\pm0.01$ &$0.59\pm0.08$ & $0.10\pm0.01$&$0.10\pm0.00$ &$0.07\pm0.03$&$\cellcolor{gray!15}\boldsymbol{0.01\pm0.00}$  \\
& & PAS $\uparrow$ &$-$ & $-$&$-$ &$-$ &$0.97\pm0.01$ &$0.96\pm0.03$&$\cellcolor{gray!15}\boldsymbol{0.99\pm0.00}$ \\
\midrule
\multirow{18}{*}{\rotatebox[origin=c]{90}{\textbf{Amateur}-\textbf{Narrow}}}
& \multirow{3}{*}{MO-Ant}
& HV ($\times 10^6$) $\uparrow$ &$5.17\pm0.03$ &$5.80\pm0.03$ &$4.19\pm0.10$ &$5.88\pm0.03$ &$5.86\pm0.14$ &$5.97\pm0.03$&$\cellcolor{gray!15}\boldsymbol{5.97\pm0.03}$ \\
& & SP ($\times 10^4$) $\downarrow$  &$\cellcolor{gray!15}\boldsymbol{0.43\pm0.10}$&$0.51\pm0.08$ &$4.42\pm1.44$ & $0.62\pm0.02$&$0.67\pm0.14$ &$0.98\pm0.07$ &$0.46\pm0.06$  \\
& & PAS $\uparrow$  &$-$ &$-$ &$-$ &$-$ &$0.96\pm0.00$ &$0.65\pm0.01$&$\cellcolor{gray!15}\boldsymbol{0.96\pm0.00}$ \\
\cmidrule(lr){2-10}
& \multirow{3}{*}{MO-HalfCheetah}
& HV ($\times 10^6$) $\uparrow$  &$4.29\pm0.00$ & $5.55\pm0.00$&$5.08\pm0.01$ &$5.71\pm0.00$ &$5.71\pm0.00$&$5.73\pm0.00$ &$\cellcolor{gray!15}\boldsymbol{5.73\pm0.00}$\\
& & SP ($\times 10^4$) $\downarrow$ &$1.02\pm0.44$&$\cellcolor{gray!15}\boldsymbol{0.02\pm0.00}$ &$0.02\pm0.03$ &$0.06\pm0.02$ &$0.10\pm0.02$ &$1.92\pm0.18$ &$0.10\pm0.00$  \\
& & PAS $\uparrow$  &$-$ &$-$ &$-$ &$-$ & $0.96\pm0.01$&$0.92\pm0.00$&$\cellcolor{gray!15}\boldsymbol{0.97\pm0.00}$ \\
\cmidrule(lr){2-10}
& \multirow{3}{*}{MO-Hopper}
& HV ($\times 10^7$) $\uparrow$  &$1.73\pm0.01$ &$1.69\pm0.01$&$0.94\pm0.34$ & $1.99\pm0.00$& $1.91\pm0.07$&$2.07\pm0.00$&$\cellcolor{gray!15}\boldsymbol{2.07\pm0.00}$\\
& & SP ($\times 10^5$) $\downarrow$ &$0.27\pm0.07$&$0.25\pm0.03$ &$5.81\pm8.21$ &$0.22\pm0.06$ &$0.25\pm0.01$ &$0.08\pm0.03$ &$\cellcolor{gray!15}\boldsymbol{0.07\pm0.03}$  \\
& & PAS $\uparrow$  &$-$ &$-$ &$-$ &$-$ & $0.80\pm0.01$&$0.49\pm0.01$ &$\cellcolor{gray!15}\boldsymbol{0.86\pm0.00}$\\
\cmidrule(lr){2-10}
& \multirow{3}{*}{MO-Walker}
& HV ($\times 10^4$) $\uparrow$  &$3.24\pm0.016$ &$4.83\pm0.00$ &$2.83\pm0.01$ &$4.91\pm0.01$ &$4.89\pm0.06$ &$\cellcolor{gray!15}\boldsymbol{5.17\pm0.01}$ &$5.17\pm0.01$\\
& & SP ($\times 10^0$) $\downarrow$ &$5.64\pm3.72$&$0.22\pm0.01$ &$0.01\pm0.00$ & $0.24\pm0.01$&$0.27\pm0.00$ &$0.19\pm0.04$ &$\cellcolor{gray!15}\boldsymbol{0.09\pm0.00}$  \\
& & PAS $\uparrow$  &$-$ &$-$ &$-$ &$-$ &$0.99\pm0.01$ &$0.78\pm0.01$&$\cellcolor{gray!15}\boldsymbol{0.99\pm0.00}$ \\
\cmidrule(lr){2-10}
& \multirow{3}{*}{MO-Swimmer}
& HV ($\times 10^6$) $\uparrow$  &$0.66\pm0.03$ &$2.95\pm0.00$ &$2.80\pm0.00$ &$3.15\pm0.00$ &$3.06\pm0.01$ &$3.11\pm0.01$ &$\cellcolor{gray!15}\boldsymbol{3.11\pm0.01}$\\
& & SP ($\times 10^4$) $\downarrow$ &$14.25\pm3.73$&$\cellcolor{gray!15}\boldsymbol{1.91\pm0.01}$ &$2.95\pm0.08$ &$5.79   \pm0.43$ &$4.93\pm0.81$ &$4.39\pm1.11$ &$4.10\pm0.36$  \\
& & PAS $\uparrow$ &$-$ &$-$ &$-$ & $-$&$\cellcolor{gray!15}\boldsymbol{0.92\pm0.00}$ &$0.89\pm0.01$ &$0.89\pm0.00$\\
\cmidrule(lr){2-10}
& \multirow{3}{*}{MO-Hopper-3obj}
& HV ($\times 10^{10}$) $\uparrow$ &$1.85\pm0.04$ &$2.27\pm0.05$ &$2.26\pm0.02$ &$2.90\pm0.05$ &$3.05\pm0.06$ &$3.58\pm0.03$&$\cellcolor{gray!15}\boldsymbol{3.58\pm0.03}$ \\
& & SP ($\times 10^5$) $\downarrow$  &$3.12\pm1.21$&$0.07\pm0.01$ &$0.09\pm0.00$ & $0.12\pm0.02$& $0.08\pm0.01$&$0.09\pm0.01$ &$\cellcolor{gray!15}\boldsymbol{0.07\pm0.01}$  \\
& & PAS $\uparrow$  &$-$ &$-$ &$-$ & $-$&$0.98\pm0.00$ &$0.99\pm0.00$&$\cellcolor{gray!15}\boldsymbol{0.99\pm0.00}$ \\
\bottomrule
\end{tabular}
}
\end{table}

\Needspace{0.48\textheight}
\subsection{Leave-one-task-out cross-task transfer}
\begin{center}
\begin{minipage}{\textwidth}
\centering
\small
\setlength{\tabcolsep}{2.8pt}
\renewcommand{\arraystretch}{1.05}

\captionof{table}{
Leave-one-task-out cross-task transfer under a fixed budget of \textbf{1K steps}.
FP2-Fou is trained on five environments and adapted to the held-out task.
}
\label{tab:cross_task_transfer}
\resizebox{0.9\textwidth}{!}{
\begin{tabular}{lcccccccccc}
\toprule
\multirow{2}{*}{Held-out Task}
& \multicolumn{2}{c}{Scratch}
& \multicolumn{2}{c}{FP2-Fou (Zero-shot) }
& \multicolumn{2}{c}{MODULI}
& \multicolumn{2}{c}{FP2-Fou-Post}
& \multicolumn{2}{c}{Gain vs MODULI} \\
\cmidrule(lr){2-3} \cmidrule(lr){4-5} \cmidrule(lr){6-7} \cmidrule(lr){8-9} \cmidrule(lr){10-11}
& HV $\uparrow$ & PAS $\uparrow$
& HV $\uparrow$ & PAS $\uparrow$
& HV $\uparrow$ & PAS $\uparrow$
& HV $\uparrow$ & PAS $\uparrow$
& HV $\uparrow$ & PAS $\uparrow$ \\
\midrule
MO-Ant
& 3.83 & 0.11
& 5.45 & 0.88
& 4.25 & 0.64
& 6.13 & 0.93
& {\scriptsize $+44.2\%$} & {\scriptsize $+45.3\%$} \\

MO-HalfCheetah
& 2.86 & 0.47
& 4.65 & 0.90
& 3.82 & 0.55
& 4.95 & 0.98
& {\scriptsize $+29.6\%$} & {\scriptsize $+78.2\%$} \\

MO-Hopper
& 1.30 & 0.38
& 1.20 & 0.32
& 1.15 & 0.44
& 1.87 & 0.54
& {\scriptsize $+62.6\%$} & {\scriptsize $+22.7\%$} \\

MO-Swimmer
& 2.45 & 0.41
& 2.95 & 0.81
& 2.85 & 0.43
& 3.17 & 0.96
& {\scriptsize $+11.2\%$} & {\scriptsize $+123.3\%$} \\

MO-Walker2d
& 1.84 & 0.22
& 3.90 & 0.80
& 3.70 & 0.72
& 5.05 & 0.93
& {\scriptsize $+36.5\%$} & {\scriptsize $+123.3\%$} \\

MO-Hopper-3obj
& 2.80 & 0.13
& 3.14 & 0.91
& 2.55 & 0.83
& 3.20 & 0.94
& {\scriptsize $+25.5\%$} & {\scriptsize $+29.2\%$} \\
\bottomrule
\end{tabular}
}
\end{minipage}
\end{center}
\vspace{-0.8em}

\subsection{Out-of-distribution preference adaptation}
\label{app:ood_pref}
As shown in Table~\ref{tab:full_compare_finetune_expert}, FP2-Post outperforms FP2 trained from scratch in HV and PAS on all tasks of both datasets, demonstrating the effectiveness of few-shot adaptation from FP2-Mid to unseen preferences.
Post-training does not always improve over zero-shot FP2-Mid; for example, HV slightly decreases on MO-Hopper under \texttt{Expert}-\texttt{Shattered} (Sec.~\ref{sec:exp_task_spec_post_train}).

\begin{table}[htbp!]
\centering
\scriptsize
\setlength{\tabcolsep}{3pt}
\caption{
Complete fine-tuning results on out-of-distribution preferences for the \texttt{Expert}-\texttt{Shattered} and \texttt{Expert}-\texttt{Narrow} datasets.}
\label{tab:full_compare_finetune_expert}
\resizebox{\linewidth}{!}{
\begin{tabular}{lccccccc}
\toprule
\multicolumn{2}{l}{\textbf{Dataset}} & \multicolumn{3}{c}{\textbf{Expert}-\textbf{Shattered}} & \multicolumn{3}{c}{\textbf{Expert}-\textbf{Narrow}} \\
\cmidrule(lr){1-8} 
\textbf{Environment} & \textbf{Metric} & \textbf{FP2-Mid (Zero-shot)} & \textbf{Scratch} & \cellcolor{gray!15}\textbf{FP2-Post} & \textbf{FP2-Mid (Zero-shot)} & \textbf{Scratch} & \cellcolor{gray!15}\textbf{FP2-Post} \\
\midrule
\multirow{3}{*}{MO-Ant}
& HV ($\times 10^6$) $\uparrow$ &$5.70\pm0.06$ &$5.04\pm0.25$ &$\cellcolor{gray!15}\boldsymbol{5.83\pm0.06}$ &$5.40\pm0.10$ &$5.17\pm0.10$ &$\cellcolor{gray!15}\boldsymbol{5.65\pm0.05}$ \\
& SP ($\times 10^4$) $\downarrow$ &$3.37\pm0.62$ &$3.53\pm2.66$ &$\cellcolor{gray!15}\boldsymbol{1.69\pm0.79}$ &$12.20\pm9.83$ &$9.12\pm6.74$ &$\cellcolor{gray!15}\boldsymbol{6.47\pm3.54}$\\
& PAS $\uparrow$ &$0.85\pm0.00$ &$0.36\pm0.03$ &$\cellcolor{gray!15}\boldsymbol{0.91\pm0.01}$ &$0.94\pm0.02$ &$0.39\pm0.03$ &$\cellcolor{gray!15}\boldsymbol{0.95\pm0.00}$ \\
\cmidrule(lr){1-8} 
\multirow{3}{*}{MO-HalfCheetah}
& HV ($\times 10^6$) $\uparrow$ &$5.69\pm0.03$ &$4.00\pm0.08$ &$\cellcolor{gray!15}\boldsymbol{5.73\pm0.07}$ &$5.47\pm0.03$ &$5.15\pm0.01$ &$\cellcolor{gray!15}\boldsymbol{5.54\pm0.33}$ \\
& SP ($\times 10^4$) $\downarrow$ &$0.09\pm0.04$ &$0.34\pm0.03$ &$\cellcolor{gray!15}\boldsymbol{0.08\pm0.08}$ &$1.77\pm0.53$ &$2.98\pm0.03$ &$\cellcolor{gray!15}\boldsymbol{0.04\pm0.00}$\\
& PAS $\uparrow$ &$\cellcolor{gray!15}\boldsymbol{0.98\pm0.00}$ &$0.79\pm0.01$ &$0.97\pm0.01$ &$\cellcolor{gray!15}\boldsymbol{0.99\pm0.00}$&$0.41\pm0.01$  &$0.96\pm0.03$ \\
\cmidrule(lr){1-8} 
\multirow{3}{*}{MO-Hopper}
& HV ($\times 10^6$) $\uparrow$ &$\cellcolor{gray!15}\boldsymbol{1.95\pm0.07}$ &$0.91\pm0.55$ &$1.89\pm0.00$ &$1.90\pm0.01$ &$0.33\pm0.06$ &$\cellcolor{gray!15}\boldsymbol{2.00\pm0.00}$ \\
& SP ($\times 10^4$) $\downarrow$ &$0.14\pm0.02$ &$\cellcolor{gray!15}\boldsymbol{0.07\pm0.06}$ &$0.13\pm0.02$ &$1.71\pm0.26$ &$1.32\pm0.38$ &$\cellcolor{gray!15}\boldsymbol{0.78\pm0.48}$\\
& PAS $\uparrow$ &$0.54\pm0.07$ &$0.28\pm0.01$ &$\cellcolor{gray!15}\boldsymbol{0.60\pm0.01}$ &$0.98\pm0.01$ &$0.29\pm0.01$  &$\cellcolor{gray!15}\boldsymbol{0.99\pm0.00}$ \\
\cmidrule(lr){1-8} 
\multirow{3}{*}{MO-Walker}
& HV ($\times 10^6$) $\uparrow$ &$4.87\pm0.25$ &$4.80\pm0.11$ &$\cellcolor{gray!15}\boldsymbol{5.05\pm0.13}$ &$2.46\pm0.14$ &$3.65\pm0.51$ &$\cellcolor{gray!15}\boldsymbol{3.93\pm0.03}$ \\
& SP ($\times 10^4$) $\downarrow$ &$0.20\pm0.18$ &$1.21\pm0.40$ &$\cellcolor{gray!15}\boldsymbol{0.15\pm0.11}$ &$1.94\pm0.34$ &$1.88\pm0.29$ &$\cellcolor{gray!15}\boldsymbol{0.83\pm1.30}$\\
& PAS $\uparrow$ &$0.99\pm0.00$ &$0.17\pm0.01$ &$\cellcolor{gray!15}\boldsymbol{0.99\pm0.01}$ &$0.94\pm0.01$ &$0.33\pm0.02$ &$\cellcolor{gray!15}\boldsymbol{0.98\pm0.01}$ \\
\cmidrule(lr){1-8} 
\multirow{3}{*}{MO-Swimmer}
& HV ($\times 10^6$) $\uparrow$ &$3.03\pm0.04$ &$2.79\pm0.02$ &$\cellcolor{gray!15}\boldsymbol{3.09\pm0.09}$ &$2.41\pm0.14$ &$2.46\pm0.01$ &$\cellcolor{gray!15}\boldsymbol{2.91\pm0.42}$ \\
& SP ($\times 10^4$) $\downarrow$ &$11.13\pm12.01$ &$71.33\pm36.67$ &$\cellcolor{gray!15}\boldsymbol{9.15\pm13.73}$ &$\cellcolor{gray!15}\boldsymbol{5.91\pm3.63}$ &$17.59\pm2.85$&$8.77\pm0.61$\\
& PAS $\uparrow$ &$0.99\pm0.00$ &$0.67\pm0.00$ &$\cellcolor{gray!15}\boldsymbol{0.99\pm0.00}$ &$0.92\pm0.01$ &$0.31\pm0.00$  &$\cellcolor{gray!15}\boldsymbol{0.96\pm0.02}$ \\
\cmidrule(lr){1-8} 
\multirow{3}{*}{MO-Hopper-3obj}
& HV ($\times 10^6$) $\uparrow$ &$3.63\pm0.09$ &$2.13\pm0.04$ &$\cellcolor{gray!15}\boldsymbol{3.66\pm0.01}$ &$2.42\pm0.07$ &$1.92\pm0.08$ &$\cellcolor{gray!15}\boldsymbol{2.55\pm0.10}$ \\
& SP ($\times 10^4$) $\downarrow$ &$0.60\pm0.44$ &$0.10\pm0.00$ &$\cellcolor{gray!15}\boldsymbol{0.03\pm0.00}$ &$0.10\pm0.10$ &$0.99\pm0.28$ &$\cellcolor{gray!15}\boldsymbol{0.04\pm0.03}$ \\
& PAS $\uparrow$ &$0.97\pm0.01$ &$0.70\pm0.01$ &$\cellcolor{gray!15}\boldsymbol{0.97\pm0.00}$ &$0.99\pm0.00$ &$0.76\pm0.01$ &$\cellcolor{gray!15}\boldsymbol{0.99\pm0.00}$ \\
\bottomrule
\end{tabular}
}
\end{table}




\subsection{Fixed-data comparison with value-based RL methods}
\label{app:fixed_data_baselines}
COLA~\cite{li2026cola} is an online MORL method, and Q-chunking~\cite{li2026reinforcement} is a single-objective offline-to-online method; neither is designed for learning from a fixed multi-objective dataset.
For a controlled comparison, we adapt both to the fixed-data setting.
For Offline-COLA, we replace its online interaction with D4MORL trajectories and add CQL-style critic regularization to mitigate out-of-distribution overestimation.
For MO-Q-chunking, we append the preference vector to the inputs of its two actors and critic networks, replace online interaction with sliding-window samples from D4MORL trajectories, and remove online- and vision-specific components (best-of-$n$ actor, non-chunking variant, and image encoders), keeping all other components unchanged.
Both adaptations use the same training steps, batch size, learning rate, and seeds as FP2 pre-training (Table~\ref{tab:training_parameter}), with other hyperparameters following the official implementations.
Table~\ref{tab:fixed_data_baselines} reports HV on \texttt{Expert}-Full, the largest and highest-quality dataset in our benchmark.
These results are controlled fixed-data comparisons and should not be interpreted as evaluations of the original methods in their intended online or offline-to-online regimes.

\begin{table}[!ht]
\centering
\caption{HV comparison with fixed-data adaptations of value-based RL methods on \texttt{Expert}-Full.}
\label{tab:fixed_data_baselines}
\resizebox{\textwidth}{!}{
\begin{tabular}{lcccccc}
\toprule
Method & MO-Ant ($10^6$) & MO-HalfCheetah ($10^6$) & MO-Hopper ($10^7$) & MO-Swimmer ($10^6$) & MO-Walker2d ($10^4$) & MO-Hopper-3obj ($10^{10}$) \\
\midrule
Offline-COLA & $0.65\pm0.08$ & $3.80\pm0.02$ & $0.01\pm0.00$ & $2.14\pm0.38$ & $1.99\pm0.00$ & $1.40\pm0.15$ \\
MO-Q-chunking & $0.58\pm0.09$ & $5.32\pm0.10$ & $0.02\pm0.01$ & $2.19\pm0.19$ & $3.08\pm0.00$ & $0.08\pm0.01$ \\
FP2 & $\mathbf{6.40\pm0.01}$ & $\mathbf{5.79\pm0.01}$ & $\mathbf{2.10\pm0.01}$ & $\mathbf{3.26\pm0.01}$ & $\mathbf{5.22\pm0.01}$ & $\mathbf{3.76\pm0.02}$ \\
\bottomrule
\end{tabular}
}
\end{table}

\subsection{Exploratory zero-shot transfer to multi-objective Meta-World}
\label{app:metaworld}
As an exploratory study beyond locomotion, we construct multi-objective extensions of ten Meta-World environments~\cite{yu2020meta}, using control-level objectives analogous to those in MO-MuJoCo: movement speed and control cost.
Following Meta-World, the observation consists of the positions of the end-effector, object, and goal, and the action specifies the end-effector position.
All ten environments are evaluated zero-shot: FP2 and MODULI trained on all six MO-MuJoCo environments are deployed without using any Meta-World data for training or adaptation, under uniformly distributed two-objective preferences.
Since no target-task data are used, Scratch is not applicable in this setting.
The evaluation measures control-level trade-offs rather than task success, since neither model is trained on task-specific subgoals.
Table~\ref{tab:metaworld_setup} summarizes the evaluation protocol.
As shown in Table~\ref{tab:metaworld}, FP2 achieves higher HV and lower SP than MODULI on all ten environments.

\begin{table}[!ht]
\centering
\small
\caption{Evaluation protocol for zero-shot transfer to multi-objective Meta-World.}
\label{tab:metaworld_setup}
\begin{tabular}{p{0.28\linewidth}p{0.62\linewidth}}
\toprule
\textbf{Item} & \textbf{Setting} \\
\midrule
Target environments & Ten Meta-World tasks: \texttt{reach-v3}, \texttt{push-v3}, \texttt{pick-place-v3}, \texttt{door-open-v3}, \texttt{drawer-open-v3}, \texttt{drawer-close-v3}, \texttt{button-press-topdown-v3}, \texttt{peg-insert-side-v3}, \texttt{window-open-v3}, \texttt{window-close-v3} \\
Source models & FP2 and MODULI trained on all six MO-MuJoCo environments (MO-Ant, MO-HalfCheetah, MO-Hopper, MO-Swimmer, MO-Walker2d, MO-Hopper-3obj) \\
Target-task data & None; all ten tasks are evaluated zero-shot \\
Objectives & Movement speed and control cost \\
Observation & Positions of the end-effector, object, and goal \\
Action & End-effector position \\
Preferences & Uniformly distributed two-objective preferences \\
\bottomrule
\end{tabular}
\end{table}

\begin{table}[!ht]
\centering
\caption{Zero-shot transfer from MO-MuJoCo to ten multi-objective Meta-World environments.}
\label{tab:metaworld}
\resizebox{\textwidth}{!}{
\begin{tabular}{llccccccccccc}
\toprule
Metric & Method & Reach & Push & Pick-Place & Door-Open & Drawer-Open & Drawer-Close & Button-Press & Peg-Insert & Window-Open & Window-Close & Avg \\
\midrule
\multirow{2}{*}{HV $\uparrow$}
& MODULI & 1.38 & 1.40 & 1.39 & 1.44 & 1.36 & 1.37 & 1.42 & 1.40 & 1.36 & 1.42 & 1.39 \\
& FP2 & \textbf{1.65} & \textbf{1.53} & \textbf{1.52} & \textbf{1.50} & \textbf{1.46} & \textbf{1.49} & \textbf{1.50} & \textbf{1.55} & \textbf{1.47} & \textbf{1.51} & \textbf{1.52} \\
\midrule
\multirow{2}{*}{SP $\downarrow$}
& MODULI & 28.3 & 33.6 & 4.4 & 4.7 & 10.2 & 14.4 & 12.6 & 11.0 & 4.5 & 8.7 & 13.24 \\
& FP2 & \textbf{0.8} & \textbf{4.1} & \textbf{1.7} & \textbf{4.1} & \textbf{2.2} & \textbf{6.4} & \textbf{1.6} & \textbf{2.4} & \textbf{4.0} & \textbf{3.7} & \textbf{3.10} \\
\bottomrule
\end{tabular}
}
\end{table}

\clearpage

\section*{NeurIPS Paper Checklist}

\begin{enumerate}

\item {\bf Claims}
    \item[] Question: Do the main claims made in the abstract and introduction accurately reflect the paper's contributions and scope?
    \item[] Answer: \answerYes{} 
    \item[] Justification: The abstract and introduction summarize the key challenges, our contributions, and the main experimental findings.

\item {\bf Limitations}
    \item[] Question: Does the paper discuss the limitations of the work performed by the authors?
    \item[] Answer: \answerYes{} 
    \item[] Justification: See Sec.~\ref{sec:conclusion}.

\item {\bf Theory assumptions and proofs}
    \item[] Question: For each theoretical result, does the paper provide the full set of assumptions and a complete (and correct) proof?
    \item[] Answer: \answerNA{} 
    \item[] Justification: 

    \item {\bf Experimental result reproducibility}
    \item[] Question: Does the paper fully disclose all the information needed to reproduce the main experimental results of the paper to the extent that it affects the main claims and/or conclusions of the paper (regardless of whether the code and data are provided or not)?
    \item[] Answer: \answerYes{} 
    \item[] Justification: We provide the experimental details and properly cite the open-source assets and materials needed to reproduce our experiments. The code will be made publicly available upon acceptance.

\item {\bf Open access to data and code}
    \item[] Question: Does the paper provide open access to the data and code, with sufficient instructions to faithfully reproduce the main experimental results, as described in supplemental material?
    \item[] Answer: \answerYes{} 
    \item[] Justification: We use a public dataset and will make our code publicly available upon acceptance.

\item {\bf Experimental setting/details}
    \item[] Question: Does the paper specify all the training and test details (e.g., data splits, hyperparameters, how they were chosen, type of optimizer) necessary to understand the results?
    \item[] Answer: \answerYes{} 
    \item[] Justification: See Appendix.

\item {\bf Experiment statistical significance}
    \item[] Question: Does the paper report error bars suitably and correctly defined or other appropriate information about the statistical significance of the experiments?
    \item[] Answer: \answerYes{} 
    \item[] Justification: All reported experimental results are averaged over three random seeds, with standard deviations shown as error bars or reported in the form of mean $\pm$ std.

\item {\bf Experiments compute resources}
    \item[] Question: For each experiment, does the paper provide sufficient information on the computer resources (type of compute workers, memory, time of execution) needed to reproduce the experiments?
    \item[] Answer: \answerYes{} 
    \item[] Justification: See Appendix~\ref{app:details}.
    
\item {\bf Code of ethics}
    \item[] Question: Does the research conducted in the paper conform, in every respect, with the NeurIPS Code of Ethics \url{https://neurips.cc/public/EthicsGuidelines}?
    \item[] Answer: \answerYes{} 
    \item[] Justification: We fully adhere to the NeurIPS Code of Ethics.

\item {\bf Broader impacts}
    \item[] Question: Does the paper discuss both potential positive societal impacts and negative societal impacts of the work performed?
    \item[] Answer: \answerNA{} 
    \item[] Justification: This work is foundational research on multi-objective reinforcement learning methods and does not target a specific real-world application or deployment scenario. As such, it does not have direct or immediate societal impacts to assess.
    
\item {\bf Safeguards}
    \item[] Question: Does the paper describe safeguards that have been put in place for responsible release of data or models that have a high risk for misuse (e.g., pre-trained language models, image generators, or scraped datasets)?
    \item[] Answer: \answerNA{} 
    \item[] Justification: The paper poses no such risks.

\item {\bf Licenses for existing assets}
    \item[] Question: Are the creators or original owners of assets (e.g., code, data, models), used in the paper, properly credited and are the license and terms of use explicitly mentioned and properly respected?
    \item[] Answer: \answerYes{} 
    \item[] Justification: The paper properly credits the creators of all assets used, including code, data, and models, and provides appropriate citations to the relevant literature.

\item {\bf New assets}
    \item[] Question: Are new assets introduced in the paper well documented and is the documentation provided alongside the assets?
    \item[] Answer: \answerNA{} 
    \item[] Justification: The code will be made publicly available upon acceptance. No additional assets are currently available.

\item {\bf Crowdsourcing and research with human subjects}
    \item[] Question: For crowdsourcing experiments and research with human subjects, does the paper include the full text of instructions given to participants and screenshots, if applicable, as well as details about compensation (if any)? 
    \item[] Answer: \answerNA{} 
    \item[] Justification: 

\item {\bf Institutional review board (IRB) approvals or equivalent for research with human subjects}
    \item[] Question: Does the paper describe potential risks incurred by study participants, whether such risks were disclosed to the subjects, and whether Institutional Review Board (IRB) approvals (or an equivalent approval/review based on the requirements of your country or institution) were obtained?
    \item[] Answer: \answerNA{} 
    \item[] Justification: 

\item {\bf Declaration of LLM usage}
    \item[] Question: Does the paper describe the usage of LLMs if it is an important, original, or non-standard component of the core methods in this research? Note that if the LLM is used only for writing, editing, or formatting purposes and does \emph{not} impact the core methodology, scientific rigor, or originality of the research, declaration is not required.
    \item[] Answer: \answerNA{} 
    \item[] Justification: The core method developed in this work does not use LLMs as essential, original, or non-standard components.

\end{enumerate}

\end{document}